%% file: main.tex
\documentclass[runningheads]{llncs}

\usepackage[mobile]{eccv}

\usepackage{eccvabbrv}

\usepackage{graphicx}
\graphicspath{{figs/}}
\usepackage{booktabs}
\usepackage{makecell} 

\usepackage{amssymb}
\usepackage{algorithm}
\usepackage{algorithmic}
\usepackage{tikz}
\usetikzlibrary{decorations.pathreplacing,calc}

\usepackage{multirow}

\usepackage[accsupp]{axessibility}  

\usepackage[pagebackref,breaklinks,colorlinks,citecolor=eccvblue]{hyperref}

\usepackage{orcidlink}

\begin{document}

\title{FlowSlider: Training-Free Continuous Image Editing via Fidelity-Steering Decomposition}

\titlerunning{FlowSlider}

\author{Taichi Endo \and
Guoqing Hao \and
Kazuhiko Sumi}

\authorrunning{Endo et al.}

\institute{Aoyama Gakuin University}

\maketitle

\begin{figure}[h]
    \centering
    \includegraphics[width=\linewidth]{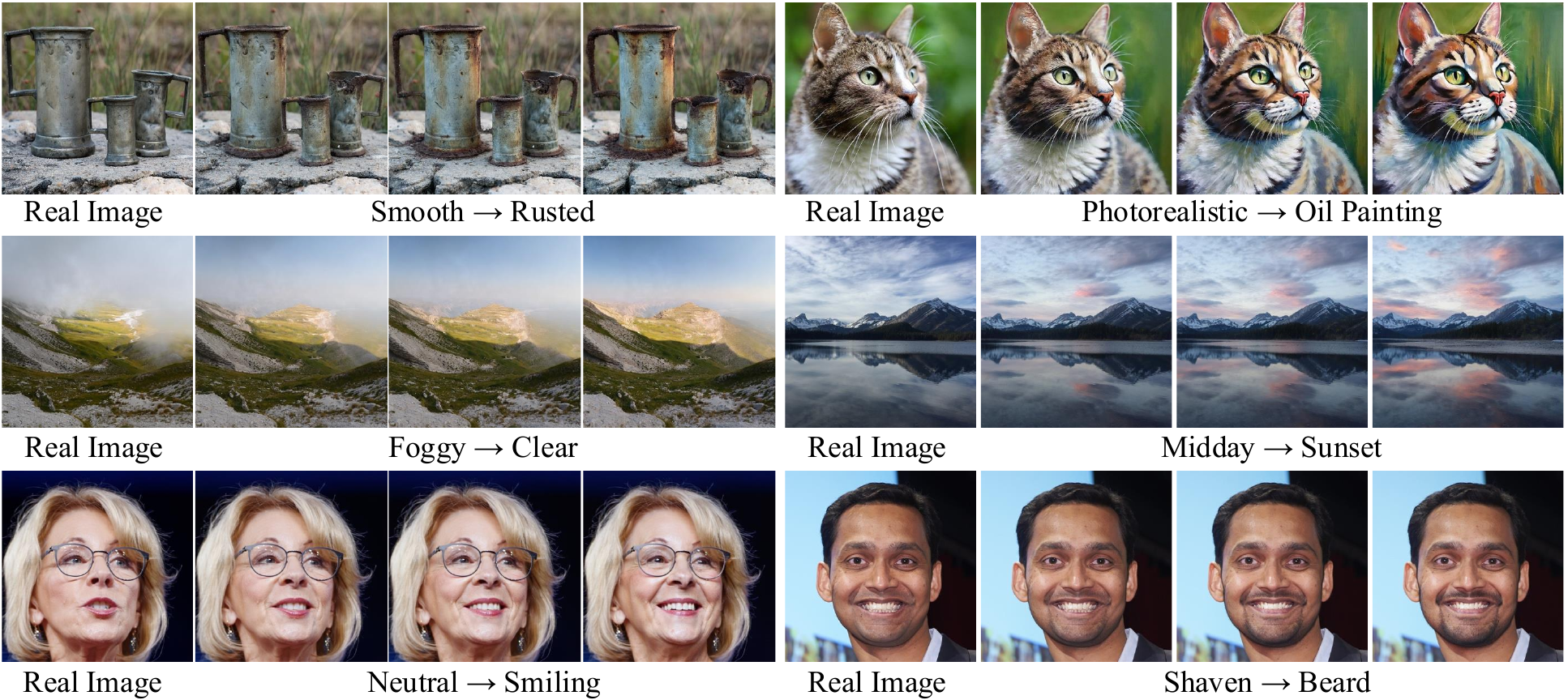}
    \caption{\textbf{FlowSlider for continuous image editing}: A slider-style editing framework that requires no learning and can be applied instantly to any edit. Simply specify a prompt pair to continuously control the editing intensity via the scaling parameter $s$. Each row shows progressively stronger editing results starting from the real image.}
    \label{fig:teaser}
\end{figure}

\begin{abstract}
Continuous image editing aims to provide slider-style
control of edit strength while preserving source-image
fidelity and maintaining a consistent edit direction.
Existing learning-based slider methods typically rely on
auxiliary modules trained with synthetic or proxy
supervision. This introduces additional training overhead
and couples slider behavior to the training distribution,
which can reduce reliability under distribution shifts in edits or
domains. We propose \textit{FlowSlider}, a training-free
method for continuous editing in Rectified Flow that
requires no post-training. \textit{FlowSlider} decomposes
FlowEdit's update into (i) a fidelity term, which acts as a
source-conditioned stabilizer that preserves identity and
structure, and (ii) a steering term that drives semantic
transition toward the target edit. Geometric analysis and
empirical measurements show that these terms are
approximately orthogonal, enabling stable strength control
by scaling only the steering term while keeping the
fidelity term unchanged. As a result, \textit{FlowSlider}
provides smooth and reliable control without post-training,
improving continuous editing quality across diverse tasks.
  
\end{abstract}

\input{01_intro}

\input{02_related}

\input{03_method}

\input{04_exps}

\input{05_conc}

\newpage


%
%
\bibliographystyle{splncs04}
\bibliography{main}

\newpage
\input{appendix_a}

\end{document}

%% file: 01_intro.tex
\section{Introduction}
\label{sec:intro}
Text-guided image editing~\cite{DBLP:conf/iclr/HertzMTAPC23,Kawar_2023_CVPR,Brooks_2023_CVPR,wang2023pix2pixzero,tumanyan2023pnpdiffusion,couairon2023diffedit,cao2023masactrl,bfl2025fluxkontext,wu2025qwenimagetechnicalreport} has made high-quality image manipulation accessible to non-experts: given a natural-language prompt, a model transforms a real input image to realize the intended change. In practice, however, users rarely want only one edited result. They need continuous control over edit strength, i.e., a slider that smoothly modulates edit intensity (\cref{fig:teaser}), to support interactive exploration and to select an appropriate trade-off between semantic change and faithfulness to the source~\cite{parihar2025kontinuouskontext,zarei2025slideredit}. A practically useful slider must preserve source-image fidelity as strength increases. It must also maintain direction consistency, so that semantic changes stay coherent and aligned with the intended edit across the full strength range, rather than drifting or introducing unintended attribute changes.
 
Prior work has explored slider-style control in both image manipulation and text-guided editing. Early approaches, particularly in Generative Adversarial Networks (GAN)-based editing~\cite{Goodfellow2014GAN}, achieved continuous control by learning domain-specific manipulation directions~\cite{harkonen2020ganspace,shen2020interpreting,abdal2021styleflow,shen2020interfacegan,patashnik2021styleclip,gal2021stylegannada}. More recent methods~\cite{parihar2025kontinuouskontext,zarei2025slideredit,DBLP:conf/eccv/GandikotaMZTB24} learn strength control from synthetic or proxy supervision. While effective, they require additional training and couple slider behavior to the training distribution, which can reduce reliability under distribution shifts across edits or domains. Meanwhile, training-free editors based on Rectified Flow~\cite{liu2022rectifiedflow}, such as FlowEdit~\cite{Kulikov_2025_ICCV}, enable high-fidelity real-image editing without additional training or inversion~\cite{rout2025rfinversion,wang2025rfediting}, but they typically lack a principled mechanism for stable strength modulation. As shown in \cref{fig:intro_failcase}, naively scaling the FlowEdit update does not yield a reliable slider: as $s$ increases, it introduces visible artifacts (e.g., ringing and over-sharpening), leading to unrealistic edited images. These observations motivate our goal: reliable slider-style continuous control in a training-free setting while preserving source fidelity and edit-direction consistency.

We propose \textit{FlowSlider}, a training-free slider mechanism for FlowEdit that enables continuous control by decomposing the editing update into a fidelity term and a steering term. The fidelity term acts as a source-conditioned stabilizer, keeping the editing dynamics close to a source-consistent regime and thereby preserving source fidelity as edit strength increases. The steering term drives the intended semantic transition toward the target edit. We control edit strength with a single scalar $s$ by scaling only the steering term while keeping the fidelity term fixed. Given a source image and a source-target prompt pair, FlowSlider generates a continuous family of edits with smooth attenuation and amplification, without post-training.

\newcommand{\failfig}[1]{%
\includegraphics[width=0.18\linewidth]{figs/visual_comps/original/#1} &
\includegraphics[width=0.18\linewidth]{figs/visual_comps/naive/#1_scale_1.jpg} &
\includegraphics[width=0.18\linewidth]{figs/visual_comps/naive/#1_scale_2.jpg} & 
\includegraphics[width=0.18\linewidth]{figs/visual_comps/ours/#1_scale_1.jpg} & 
\includegraphics[width=0.18\linewidth]{figs/visual_comps/ours/#1_scale_2.jpg} %
}

\begin{figure*}[t]
  \centering
  \setlength{\tabcolsep}{1pt}
  \begin{tabular}{c|cc|cc}
   & \multicolumn{2}{c|}{Naive Scaling in FlowEdit}  & \multicolumn{2}{c}{FlowSlider} \\
   \failfig{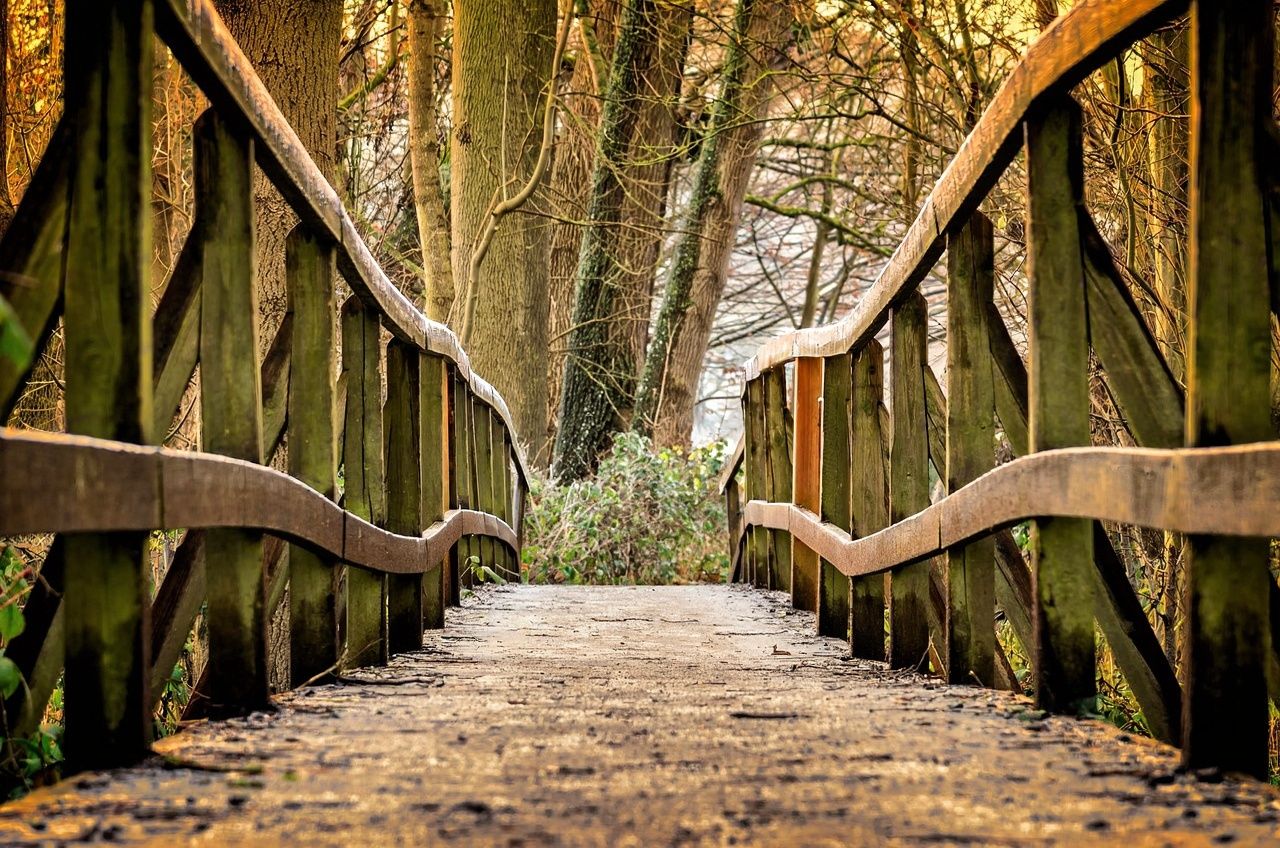} \\
   Original & $s=1$ & $s=2$ & $s=1$ & $s=2$ \\
  \end{tabular}
  \vspace{-3mm}
  \caption{`Autumn $\rightarrow$ Winter' editing example. Naive scaling in FlowEdit~\cite{Kulikov_2025_ICCV} introduces artifacts at $s=2$, whereas FlowSlider maintains stable fidelity. Note that our FlowSlider is identical to FlowEdit when $s=1$. }
  \label{fig:intro_failcase}
\end{figure*}

This formulation offers two key advantages for continuous image editing. First, it is training-free by construction: strength control is implemented as an inference-time modification of the editing dynamics rather than a learned strength mapping. As a result, FlowSlider requires no additional data or training and does not tie slider behavior to a specific training distribution, allowing the same control mechanism to be applied directly to a wide range of prompt pairs and real images. Second, it is stable and requirement-aligned: the fidelity term acts as a source-conditioned stabilizer that mitigates drift and artifacts, thereby preserving source fidelity, while the steering term governs semantic change and provides a principled control axis for edit strength.
Moreover, our geometric analysis, supported by empirical measurements, indicates that these two components are approximately orthogonal. This explains why scaling the steering term modulates edit strength while largely preserving the stabilizing effect of the fidelity term. Together, these properties yield reliable slider behavior that meets the desiderata above.

We evaluate edit quality and slider behavior on our continuous editing benchmark and the general editing benchmark PIE-Bench. Results show that FlowSlider achieves the best slider behavior while maintaining superior edit quality compared with existing methods. We further validate generality on two Rectified Flow backbones, FLUX.1-dev and Stable Diffusion 3 Medium, without any post-training.

%% file: 02_related.tex
\section{Related Work}

\subsection{Text-guided real-image editing}
Text-guided image editing~\cite{Brooks_2023_CVPR,wang2023pix2pixzero,tumanyan2023pnpdiffusion,couairon2023diffedit,cao2023masactrl} has progressed rapidly with large-scale text-to-image generative models~\cite{Rombach2022LDM,bfl2024flux1dev,Saharia2022Imagen,Ramesh2022DALLE2,wu2025qwenimagetechnicalreport,ZImage2025,esser2024scalingrectifiedflowtransformers}, driven by diffusion models~\cite{sohl2015deep,ddpm,song2021denoising,nichol2021improvedddpm,dhariwal2021diffusionbeatgans} and, more recently, flow matching~\cite{Lipman2023FlowMatching}/Rectified Flow~\cite{liu2022rectifiedflow} formulations. 
A useful way to organize this literature is by editing interface.
Prompt-pair editing specifies a source prompt and a target prompt and edits by transporting between the corresponding conditional distributions. Prompt-to-Prompt~\cite{DBLP:conf/iclr/HertzMTAPC23} popularized prompt-pair control for generated images by manipulating cross-attention~\cite{Rombach2022LDM} during generation. Extending this paradigm to real images typically requires inversion or optimization~\cite{dreambooth2023} to preserve input identity and structure, as explored in Imagic~\cite{Kawar_2023_CVPR}, Null-text Inversion~\cite{mokady2022nulltextinversion}, DiffEdit~\cite{couairon2023diffedit}, and Plug-and-Play Diffusion Features~\cite{tumanyan2023pnpdiffusion}.
Instruction-based editing instead uses a natural-language command and produces an edited image directly. A representative direction trains instruction-following editors on curated or synthetic supervision: InstructPix2Pix~\cite{Brooks_2023_CVPR} constructs large-scale synthetic edit pairs using a language model~\cite{Brown2020GPT3} and a text-to-image model~\cite{Rombach2022LDM}, enabling single-pass instruction editing. More recent foundation editors, including FLUX.1 Kontext~\cite{bfl2025fluxkontext}, Z-Image-Edit~\cite{ZImage2025}, and Qwen-Image-Edit~\cite{wu2025qwenimagetechnicalreport}, further scale data and training pipelines, demonstrating broad editing capability across diverse instructions.

These works primarily study what edits can be produced and how faithfully they can be applied to real images. In contrast, our focus is complementary: predictable, slider-style continuous control of edit strength in a training-free real-image editor.

\subsection{Learning-based continuous control and sliders}
Continuous control predates diffusion models. GANs-era editing methods~\cite{harkonen2020ganspace,shen2020interpreting,abdal2021styleflow,shen2020interfacegan,patashnik2021styleclip,gal2021stylegannada} introduced interpretable latent directions that enable smooth attribute and style traversal. In recent diffusion-based frameworks, slider-like control is typically obtained by learning explicit strength-calibration mechanisms, e.g., learning directions, learning mappings from a scalar to a modulation space, or training lightweight control modules. Concept Sliders~\cite{DBLP:conf/eccv/GandikotaMZTB24}, for example, train Low-Rank Adaptation (LoRA)~\cite{Hu2022LoRA} parameter directions for targeted concepts and expose the scale as a continuous control.

Most relevant to modern instruction-driven editors, Kontinuous Kontext~\cite{parihar2025kontinuouskontext} introduces an explicit scalar strength by training a lightweight projector on diffusion transformers~\cite{Peebles2023DiT} that maps a strength value to modulation offsets, using synthesized supervision with strength labels~\cite{parihar2025kontinuouskontext}. They use Qwen-VL~\cite{Bai2023QwenVL} and FLUX.1 Kontext~\cite{bfl2025fluxkontext} to generate editing triplets (source image, instruction, edited image), and then use FreeMorph~\cite{Cao_2025_ICCV} to create intermediate edited images. SliderEdit~\cite{zarei2025slideredit} targets multi-instruction editing by training per-instruction sliders (e.g., adapter/LoRA-style updates) to enable independent strength adjustment for each instruction. Collectively, these methods demonstrate that slider control is achievable and often effective; however, this controllability is realized through additional supervision and training to calibrate the relationship between a scalar and the resulting edit effect.

Our goal is a different operating point: slider-style control without any post-training, derived directly from the editing dynamics of a pre-trained Rectified Flow editor.

\subsection{Training-free editing and heuristic strength control}
A common post-hoc heuristic approach to strength control in generative editing is to vary inference-time parameters that trade off source fidelity and semantic change. For example, SDEdit~\cite{meng2022sdedit} exposes natural control axes by changing the injected noise level and denoising duration. In text-guided editing, practitioners often adjust heuristic knobs such as classifier-free guidance (CFG)~\cite{Ho2022CFG}, guidance schedules, or the effective editing window (e.g., the start timestep), often in combination with fast ODE samplers~\cite{lu2022dpmsolver}. While these controls can change the magnitude of edits, they frequently entangle edit strength with edit direction and source fidelity, leading to non-monotonic behavior, semantic drift, or degraded preservation, particularly when extrapolating beyond normal settings.

This issue is especially salient for training-free editing with Rectified Flow models. FlowEdit~\cite{Kulikov_2025_ICCV} enables inversion-free, optimization-free real-image editing by constructing an ODE that maps between source- and target-prompt distributions. However, FlowEdit itself does not provide a principled slider: naively scaling or interpolating its velocity update, or tuning CFG-like parameters, can destabilize the ODE path, amplify residual components from approximate differencing, and thereby sacrifice fidelity. \textit{FlowSlider} is designed precisely to address this gap. We decompose FlowEdit's editing velocity update into a fidelity component and a steering component, and modulate strength by scaling only the steering component while preserving the fidelity term, yielding stable and predictable continuous control.

%% file: 03_method.tex
\section{Method}
\label{sec:method}

\subsection{Slider-style continuous image editing}
\label{sec:method_setup}
We study continuous text-guided real-image editing under a prompt-pair interface.
Given a real source image $x_{\mathrm{src}}$, a noise sample $\epsilon$, and a source--target prompt pair
$(c_{\mathrm{src}}, c_{\mathrm{tar}})$, we aim to generate a continuous family of edited images
indexed by a scalar strength $s \in \mathbb{R}$.

A practically useful slider should satisfy three desiderata:
(i) monotonic strength control: increasing $s$ should increase edit magnitude in a smooth manner;
(ii) source fidelity: identity and structure of $x_{\mathrm{src}}$ should be preserved as strength increases;
and (iii) direction consistency: the semantic change should remain aligned with the intended edit across $s$,
without drift or unintended attribute changes.
Our goal is to realize this slider behavior in a training-free manner, i.e., without post-training or curated data, by building on FlowEdit~\cite{Kulikov_2025_ICCV}.

\subsection{Preliminaries: Rectified Flow and FlowEdit}
\label{sec:preliminaries}

\subsubsection{Rectified Flow and classifier-free guidance}
\label{sec:rf_cfg}
Rectified Flow (RF) models learn an ordinary differential equation (ODE) that transports a source distribution $p_0$
(e.g., noise) to a target distribution $p_1$ (e.g., images). Given $z_1\sim p_1$ and $z_0\sim p_0$,
RF defines a linear probability path
\begin{equation}
  z_t = (1-t)z_1 + t z_0, \quad t\in[0,1],
  \label{eq:rf_interp}
\end{equation}
and trains a neural network to predict the velocity field $V_\theta(z_t,t)$ corresponding to $\frac{d z_t}{dt} = z_0 - z_1$.
At inference time, sampling is performed by numerically solving the learned ODE backward from $t=1$ to $t=0$~\cite{liu2022rectifiedflow}.

To achieve text-to-image generation, classifier-free guidance (CFG)~\cite{Ho2022CFG} is commonly applied. Let $\emptyset$ denote the null prompt.
The guided velocity can be written as
\begin{equation}
  V_{\mathrm{cfg}}(z_t,t,c) = V_\theta(z_t,t,\emptyset) + \omega\big(V_\theta(z_t,t,c)-V_\theta(z_t,t,\emptyset)\big),
  \label{eq:cfg}
\end{equation}
where $\omega\ge 0$ is the guidance scale and $c$ is a text prompt.
For simplicity, we write $V(\cdot)$ for the guided velocity in the remainder.

\subsubsection{FlowEdit for prompt-pair editing}
\label{sec:flowedit_update}

FlowEdit~\cite{Kulikov_2025_ICCV} is an inference algorithm for prompt-pair editing that avoids ODE inversion.
Given a source image $x_{\mathrm{src}}$, noise $\epsilon$, and prompts $(c_{\mathrm{src}},c_{\mathrm{tar}})$,
FlowEdit constructs the noisy source state at any time $t$:
\begin{equation}
  z_t^{\mathrm{src}} = (1-t)x_{\mathrm{src}} + t\epsilon.
  \label{eq:flowedit_src}
\end{equation}
It maintains an editing path $z_t^{\mathrm{edit}}$, initialized at the start of the integration window as the corresponding
noisy source state $z_t^{\mathrm{src}}$. At each step, it forms an anchored target state
\begin{equation}
  z_t^{\mathrm{tar}} = z_t^{\mathrm{edit}} + z_t^{\mathrm{src}} - x_{\mathrm{src}},
  \label{eq:flowedit_tar}
\end{equation}
and computes a velocity difference
\begin{equation}
  V^\Delta(t) = V(z_t^{\mathrm{tar}},t,c_{\mathrm{tar}}) - V(z_t^{\mathrm{src}},t,c_{\mathrm{src}}),
  \label{eq:vdelta}
\end{equation}
The edit path is subsequently updated using this difference: $\frac{d z_t^{\text{edit}}}{dt} = V^\Delta(t)$.

\subsection{FlowSlider}
\label{sec:flowslider}

\subsubsection{Motivation: why naive scaling fails}
\label{sec:motivation}
The difference update in \cref{eq:vdelta} explains how FlowEdit preserves fidelity while following the edit direction. The term $V(z_t^{\mathrm{tar}},t,c_{\mathrm{tar}})$ provides target-directed semantic drive, while the subtracted term $V(z_t^{\mathrm{src}},t,c_{\mathrm{src}})$ acts as a source-conditioned reference. Because $z_t^{\mathrm{tar}}$ and $z_t^{\mathrm{src}}$ share the same noise realization $\epsilon$ through \cref{eq:flowedit_tar}, their predicted velocities contain correlated noise components, and subtraction yields approximate noise cancellation.
This keeps the trajectory near a source-consistent regime and preserves fidelity while moving toward target semantics, consistent with FlowEdit's lower transport-cost observation.

A natural attempt at continuous control is to introduce a scalar strength $s$ and scale the update as $s\,V^\Delta(t)$.
However, this naive scaling amplifies both parts of the coupled mechanism above: semantic drive and source-conditioned stabilization. Because the cancellation in \cref{eq:vdelta} is approximate, scaling by $s>1$ also amplifies residual non-cancelled components. 
These residuals accumulate during integration, perturb $z_t^{\mathrm{edit}}$ and thus $z_t^{\mathrm{tar}}$ via \cref{eq:flowedit_tar}, and weaken the $z_t^{\mathrm{src}}$--$z_t^{\mathrm{tar}}$ coupling as the trajectory leaves the source-consistent regime. 
Consequently, noise suppression degrades and drift increases. This often leads to unrealistic artifacts in edited images, as shown in \cref{fig:intro_failcase}, motivating a mechanism that increases semantic strength while preserving source-stabilizing dynamics.

\subsubsection{Fidelity--steering decomposition}
\label{sec:decomp}
\begin{figure*}[t]
\centering
\begin{minipage}[t]{0.48\linewidth}
\centering
\includegraphics[width=\linewidth]{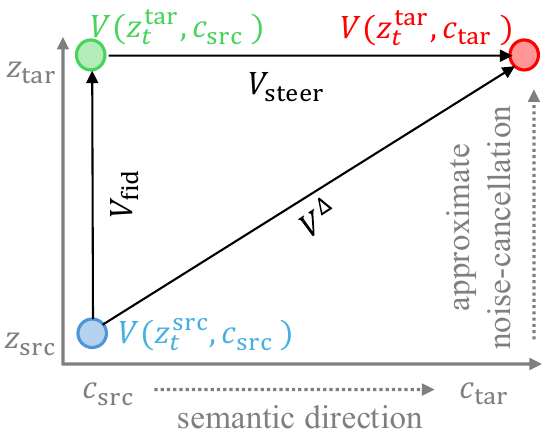}
\\[0.5mm]
\footnotesize (a) Fidelity--steering decomposition ($V^{\Delta}=V_{\mathrm{fid}}+V_{\mathrm{steer}}$)
\end{minipage}
\hfill
\begin{minipage}[t]{0.50\linewidth}
\centering
\includegraphics[width=\linewidth]{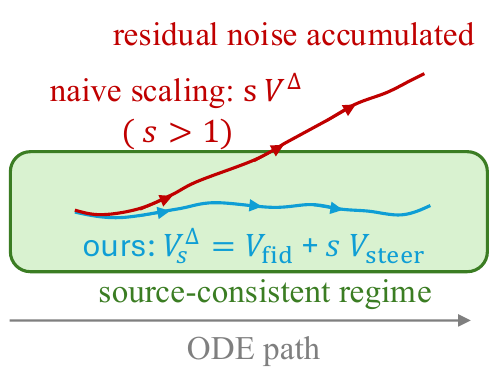}
\\[0.5mm]
\footnotesize (b) Naive scaling amplifies both terms; decomposed scaling steers only
\end{minipage}
\caption{\textbf{Fidelity--steering decomposition and its implication for strength control.}
\textbf{(a)} Exact decomposition of the FlowEdit update into a same-state, cross-prompt steering term and a same-prompt, cross-state fidelity term:
$V^{\Delta}=V_{\mathrm{fid}}+V_{\mathrm{steer}}$.
\textbf{(b)} Naive scaling ($sV^{\Delta}$) magnifies both components and residual non-cancelled noise, which can drive the trajectory out of the source-consistent regime; decomposed scaling
$V^{\Delta}_{s}=V_{\mathrm{fid}}+sV_{\mathrm{steer}}$ increases semantic strength while keeping source-conditioned stabilization unchanged.}
\label{fig:decomp_schematic}
\end{figure*}

Motivated by the failure mode above, we isolate the two coupled effects in FlowEdit: prompt-driven semantic steering and source-conditioned stabilization. We introduce an intermediate velocity at the anchored target state $z_t^{\mathrm{tar}}$ with the source prompt $c_{\mathrm{src}}$, denoted as $V(z_t^{\mathrm{tar}},t,c_{\mathrm{src}})$. Adding and subtracting this term in \cref{eq:vdelta} yields the exact decomposition:
\begin{align}
  V^\Delta(t)
  &= \underbrace{\Big(V(z_t^{\mathrm{tar}},t,c_{\mathrm{tar}})-V(z_t^{\mathrm{tar}},t,c_{\mathrm{src}})\Big)}_{V_{\mathrm{steer}}(t)}
   + \underbrace{\Big(V(z_t^{\mathrm{tar}},t,c_{\mathrm{src}})-V(z_t^{\mathrm{src}},t,c_{\mathrm{src}})\Big)}_{V_{\mathrm{fid}}(t)}.
  \label{eq:decomp}
\end{align}
We denote $V_{\mathrm{steer}}(t)$ as the steering term and $V_{\mathrm{fid}}(t)$ as the fidelity term; this is an exact algebraic identity and introduces no additional approximation.
A schematic illustration is shown in \cref{fig:decomp_schematic}. In \cref{fig:decomp_schematic}(a), the horizontal axis denotes the prompt transition $c_{\mathrm{src}}\!\rightarrow\!c_{\mathrm{tar}}$, and the vertical axis denotes the shared-noise coupled state pair $(z_t^{\mathrm{src}},z_t^{\mathrm{tar}})$.
Accordingly,
$V_{\mathrm{steer}}(t)=V(z_t^{\mathrm{tar}},t,c_{\mathrm{tar}})-V(z_t^{\mathrm{tar}},t,c_{\mathrm{src}})$ is a same-state, cross-prompt difference,
$V_{\mathrm{fid}}(t)=V(z_t^{\mathrm{tar}},t,c_{\mathrm{src}})-V(z_t^{\mathrm{src}},t,c_{\mathrm{src}})$ is a same-prompt, cross-state difference; and
$V^{\Delta}(t)=V_{\mathrm{fid}}(t)+V_{\mathrm{steer}}(t)$ is their sum.
This view is consistent with the motivation above: naive scaling of the full $V^{\Delta}$ amplifies residual non-cancelled components, while separating the terms enables later strength modulation of semantic steering without changing source-regime stabilization (\cref{fig:decomp_schematic}(b)).

\subsubsection{Interpretation: semantic steering vs. fidelity stabilization}
This decomposition makes the role of each component explicit: one term governs prompt-induced semantic movement, while the other provides source-conditioned trajectory stabilization.
\begin{itemize}
    \item \textbf{Steering term $V_{\mathrm{steer}}$: semantic control axis.} 
    $V_{\mathrm{steer}}(t)=V(z_t^{\mathrm{tar}},t,c_{\mathrm{tar}})-V(z_t^{\mathrm{tar}},t,c_{\mathrm{src}})$
    is a same-state, cross-prompt difference.
    Because both evaluations share the same state $z_t^{\mathrm{tar}}$ and differ only by prompt condition, this term isolates the prompt-induced directional change at that state.
    Scaling $V_{\mathrm{steer}}$ therefore primarily changes edit strength along the semantic transition $c_{\mathrm{src}}\!\rightarrow\!c_{\mathrm{tar}}$, making it a natural control handle for a continuous slider.

    \item \textbf{Fidelity term $V_{\mathrm{fid}}$: source-consistency stabilizer.}   
    $V_{\mathrm{fid}}(t)=V(z_t^{\mathrm{tar}},t,c_{\mathrm{src}})-V(z_t^{\mathrm{src}},t,c_{\mathrm{src}})$
    is a same-prompt, cross-state difference.
    It measures how the anchored target state deviates from the source-coupled reference under the same conditioning, and acts as a corrective component that keeps the trajectory in a source-consistent regime.
    Maintaining this term unscaled preserves the shared-noise coupling between $z_t^{\mathrm{src}}$ and $z_t^{\mathrm{tar}}$, which helps limit accumulation of residual non-cancelled components and better retain identity/structure.
\end{itemize}

\subsubsection{Decomposed strength modulation}
\label{sec:strength_mod}
With the editing update cleanly separated, FlowSlider achieves continuous control by introducing a scalar strength $s$ that scales only the steering term:
\begin{equation}
  V^\Delta_s(t) = V_{\mathrm{fid}}(t) + s\,V_{\mathrm{steer}}(t).
  \label{eq:flowslider_vdelta}
\end{equation}
The ODE path is then integrated using $\frac{d z_t^{\text{edit}}}{dt} = V^\Delta_s(t)$. This formulation yields two critical properties: (i) setting $s=1$ exactly recovers the original FlowEdit update, and (ii) varying $s$ modulates the semantic intensity of the edit while keeping the source-conditioned stabilizer $V_{\mathrm{fid}}$ intact.

\subsubsection{Why is decomposed scaling stable?}
\label{sec:ortho}
The key point is geometric: if $V_{\mathrm{fid}}$ and $V_{\mathrm{steer}}$ are close to orthogonal, they are weakly coupled.
For
$V_s^\Delta(t)=V_{\mathrm{fid}}(t)+sV_{\mathrm{steer}}(t)$,
the update magnitude satisfies
\begin{equation}
\|V_s^\Delta(t)\|^2
=\|V_{\mathrm{fid}}(t)\|^2+s^2\|V_{\mathrm{steer}}(t)\|^2
+2s\,\|V_{\mathrm{fid}}(t)\|\,\|V_{\mathrm{steer}}(t)\|\cos\theta(t),
\end{equation}
where $\theta(t)$ is the angle between $V_{\mathrm{fid}}(t)$ and $V_{\mathrm{steer}}(t)$.
The interaction is entirely in the cross term.
Thus, when $\theta(t)\approx 90^\circ$ (i.e., $\cos\theta(t)\approx 0$), changing $s$ mainly increases the steering component and has limited effect on fidelity dynamics.
This can be seen directly by projecting onto the fidelity direction
$\hat{u}_{\mathrm{fid}}=V_{\mathrm{fid}}/\|V_{\mathrm{fid}}\|$:
\begin{equation}
p_{\mathrm{fid}}(s,t)=\langle V_s^\Delta(t),\hat{u}_{\mathrm{fid}}\rangle
=\|V_{\mathrm{fid}}(t)\|+s\,\|V_{\mathrm{steer}}(t)\|\cos\theta(t),
\end{equation}
so
$\partial p_{\mathrm{fid}}/\partial s=\|V_{\mathrm{steer}}(t)\|\cos\theta(t)\approx 0$
under near-orthogonality.
In contrast, the steering projection satisfies
$\partial p_{\mathrm{steer}}/\partial s=\|V_{\mathrm{steer}}(t)\|$,
so steering strength grows linearly with $s$.

\begin{figure}[t]
    \centering
    \includegraphics[width=0.95\linewidth]{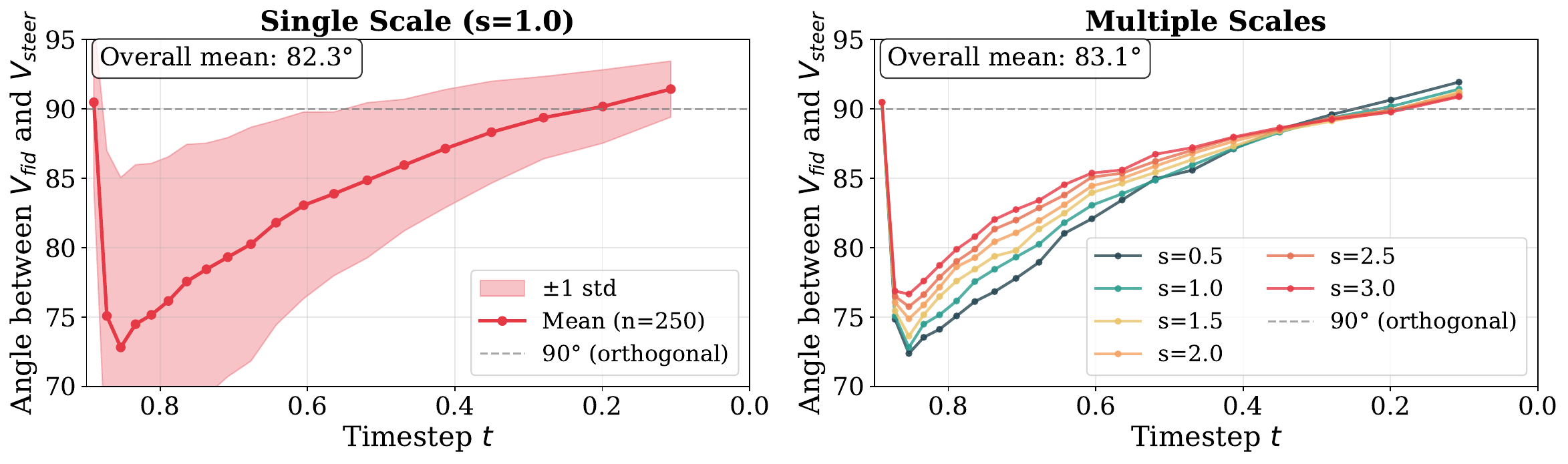}
    \caption{\textbf{Orthogonality between fidelity and steering components on our benchmark data.}
    We plot the angle  $\theta(t) = \arccos(
\frac{\langle V_{\mathrm{fid}}(t),V_{\mathrm{steer}}(t)\rangle}
{\|V_{\mathrm{fid}}(t)\|\,\|V_{\mathrm{steer}}(t)\|})$ across timesteps and strength settings.
    Angles remain concentrated near $90^\circ$, indicating weak coupling between source-conditioned stabilization and semantic steering.}
    \label{fig:orthogonality}
\end{figure}

Empirically, \cref{fig:orthogonality} shows that $\theta(t)$ is concentrated near $90^\circ$ across timesteps and strength settings, supporting this weak-coupling regime.
In contrast, naive scaling applies
$sV^\Delta=sV_{\mathrm{fid}}+sV_{\mathrm{steer}}$,
so both components are amplified regardless of their geometric coupling.
As shown in our ablations (\cref{tab:ablation}, \cref{fig:ablt}), it also magnifies residual non-cancelled components in $V^\Delta$; accumulated error can push trajectories away from the source-consistent regime and induce drift or ringing/over-sharpening artifacts at large $s$.

\subsubsection{Algorithm}
\label{sec:algo}
Algorithm~\ref{alg:flowslider} summarizes the FlowSlider inference procedure.
Compared with FlowEdit, FlowSlider requires one additional velocity evaluation per step,
$V(z_t^{\mathrm{tar}},t,c_{\mathrm{src}})$, to realize the decomposition in Eq.~\eqref{eq:decomp}.
Accordingly, each step evaluates three guided velocity fields:
$V(z_t^{\mathrm{tar}},t,c_{\mathrm{tar}})$, $V(z_t^{\mathrm{tar}},t,c_{\mathrm{src}})$, and $V(z_t^{\mathrm{src}},t,c_{\mathrm{src}})$.

\begin{algorithm}[tb]
   \caption{FlowSlider algorithm}
   \label{alg:flowslider}
\begin{algorithmic}
   \STATE {\bfseries Input:}
   source image $x_{\mathrm{src}}$, prompts $(c_{\mathrm{src}}, c_{\mathrm{tar}})$, noise $\epsilon$,
   strength $s$, time grid $\{t_i\}_{i=0}^{T}$, edit-start index $n_{\max}$
   \STATE {\bfseries Output:} edited image $x^{\mathrm{edit}}_{0}$
   \STATE {\bfseries Init:} $z^{\mathrm{edit}}_{t_{n_{\max}}} \leftarrow z^{\mathrm{src}}_{t_{n_{\max}}}$, where
   $z^{\mathrm{src}}_{t}=(1-t)x_{\mathrm{src}}+t\epsilon$

    \FOR{$i=n_{\max}$ {\bfseries to} $1$}
   \STATE $z^{\mathrm{src}}_{t_i} \leftarrow (1-t_i)x_{\mathrm{src}} + t_i\epsilon$
   \STATE $z^{\mathrm{tar}}_{t_i} \leftarrow z^{\mathrm{edit}}_{t_i} + z^{\mathrm{src}}_{t_i} - x_{\mathrm{src}}$
   \STATE $V_{\mathrm{steer}} \leftarrow V(z^{\mathrm{tar}}_{t_i}, t_i, c_{\mathrm{tar}}) - V(z^{\mathrm{tar}}_{t_i}, t_i, c_{\mathrm{src}})$
   \STATE $V_{\mathrm{fid}} \leftarrow V(z^{\mathrm{tar}}_{t_i}, t_i, c_{\mathrm{src}}) - V(z^{\mathrm{src}}_{t_i}, t_i, c_{\mathrm{src}})$
   \STATE $V^{\Delta}_{s} \leftarrow V_{\mathrm{fid}} + s\,V_{\mathrm{steer}}$
   \STATE $z^{\mathrm{edit}}_{t_{i-1}} \leftarrow z^{\mathrm{edit}}_{t_i} + (t_{i-1}-t_i)V^{\Delta}_{s}$
   \ENDFOR
   \STATE $x^{\mathrm{edit}}_{0} \leftarrow z^{\mathrm{edit}}_{0}$
   \STATE {\bfseries Return:} $x^{\mathrm{edit}}_{0}$
\end{algorithmic}
\end{algorithm}

%% file: 04_exps.tex
\section{Experiments}
\label{sec:exp}

\subsection{Experimental setup}
We evaluate FlowSlider using FLUX.1-dev~\cite{bfl2024flux1dev} and Stable Diffusion 3 Medium~\cite{esser2024scalingrectifiedflowtransformers} as backbones. For both backbones, we use $T{=}28$ steps, an edit-start step of $n_{\max}{=}20$, and a CFG scale of $\omega{=}3.5$ for both source and target prompts. We evaluate continuous editing using a fixed set of strength values, $s\in\{1,2,3,4,5\}$, and reuse the same noise seed across different $s$ for each example to isolate the effect of strength modulation.

\subsection{Benchmark: continuous image editing}
To evaluate slider-style continuous control in a source--target prompt-pair setting, we construct a 250-sample benchmark from two sources: PIE-Bench~\cite{ju2023direct}, from which we select 31 samples suitable for continuous image editing, and a set of Pixabay~\cite{pixabay} images annotated with editing instructions following the PIE-Bench format.
The tasks include season change, style change, time decay, weather change, expression change, color change, hair color, makeup, aging/de-aging, facial hair, and hairstyle changes.
Each sample consists of a real source image, an editing instruction, and a source--target prompt pair $(c_{\mathrm{src}},c_{\mathrm{tar}})$.
We report results averaged across all samples unless otherwise specified, and also provide category-wise qualitative examples in the appendix.

In addition to this continuous-editing benchmark, we conduct a general-editing evaluation on PIE-Bench~\cite{ju2023direct} with the FLUX.1 backbone, following the protocol of Kontinuous Kontext~\cite{parihar2025kontinuouskontext}.
Importantly, PIE-Bench is not designed for slider-style continuous control: even after excluding \emph{Add} and \emph{Remove}, it still contains edits such as `transform dog to cat' that are not naturally slider-continuous.
Therefore, we use PIE-Bench only to assess generalization.

\subsection{Baselines}
\label{sec:baselines}
We compare FlowSlider against (i) learning-based slider methods that learn explicit strength calibration and (ii) training-free, inference-time strength controls based on post-hoc heuristic parameter tuning.

\paragraph{Learning-based slider methods.}
We compare against representative learning-based continuous-control approaches, including Kontinuous Kontext~\cite{parihar2025kontinuouskontext} and SliderEdit~\cite{zarei2025slideredit}, using their official inference pipelines and recommended settings.
These methods realize slider behavior by learning strength-dependent modulation from synthetic or proxy supervision. Since they are instruction-based editing approaches, we use editing instructions for comparison rather than the source--target prompt pairs used in our method.

\paragraph{Training-free heuristic controls.}
We first compare with CFG-scaling results for FLUX-Kontext~\cite{bfl2025fluxkontext} under its official inference pipeline.
We also evaluate two common inference-time strategies for modulating edit strength in FlowEdit~\cite{Kulikov_2025_ICCV}: (a) CFG scaling, which varies the target classifier-free guidance scale while keeping the source fixed; we use CFG
$\omega_{\mathrm{tar}}\in\{1.5, 3.5, 5.5, 7.5, 9.5\}$.
(b) edit-window $n_{\max}$ tuning, which varies the edit-start step to change the effective editing path; we use
$n_{\max}\in\{20, 22, 24, 26, 28\}$ for comparison.
In addition, we include naive strength scaling of the original FlowEdit update (scaling $V^\Delta$ directly) as an ablation, as it is the most direct but typically unstable way to introduce a strength parameter.

\newcommand{\vcfig}[2]{%
\includegraphics[width=0.155\linewidth]{figs/visual_comps/original/#2} &
\includegraphics[width=0.155\linewidth]{figs/visual_comps/#1/#2_scale_1.jpg} &
\includegraphics[width=0.155\linewidth]{figs/visual_comps/#1/#2_scale_2.jpg} &
\includegraphics[width=0.155\linewidth]{figs/visual_comps/#1/#2_scale_3.jpg} &
\includegraphics[width=0.155\linewidth]{figs/visual_comps/#1/#2_scale_4.jpg} &
\includegraphics[width=0.155\linewidth]{figs/visual_comps/#1/#2_scale_5.jpg}%
}

\begin{figure*}[t]
  \centering
  \setlength{\tabcolsep}{1pt}
  \begin{tabular}{cc|ccccc}
  & \multicolumn{1}{c|}{Original} & \multicolumn{5}{c}{Clean $\rightarrow$ Decay} \\
  \raisebox{3 mm}{\rotatebox{90}{Ours}} & \vcfig{ours}{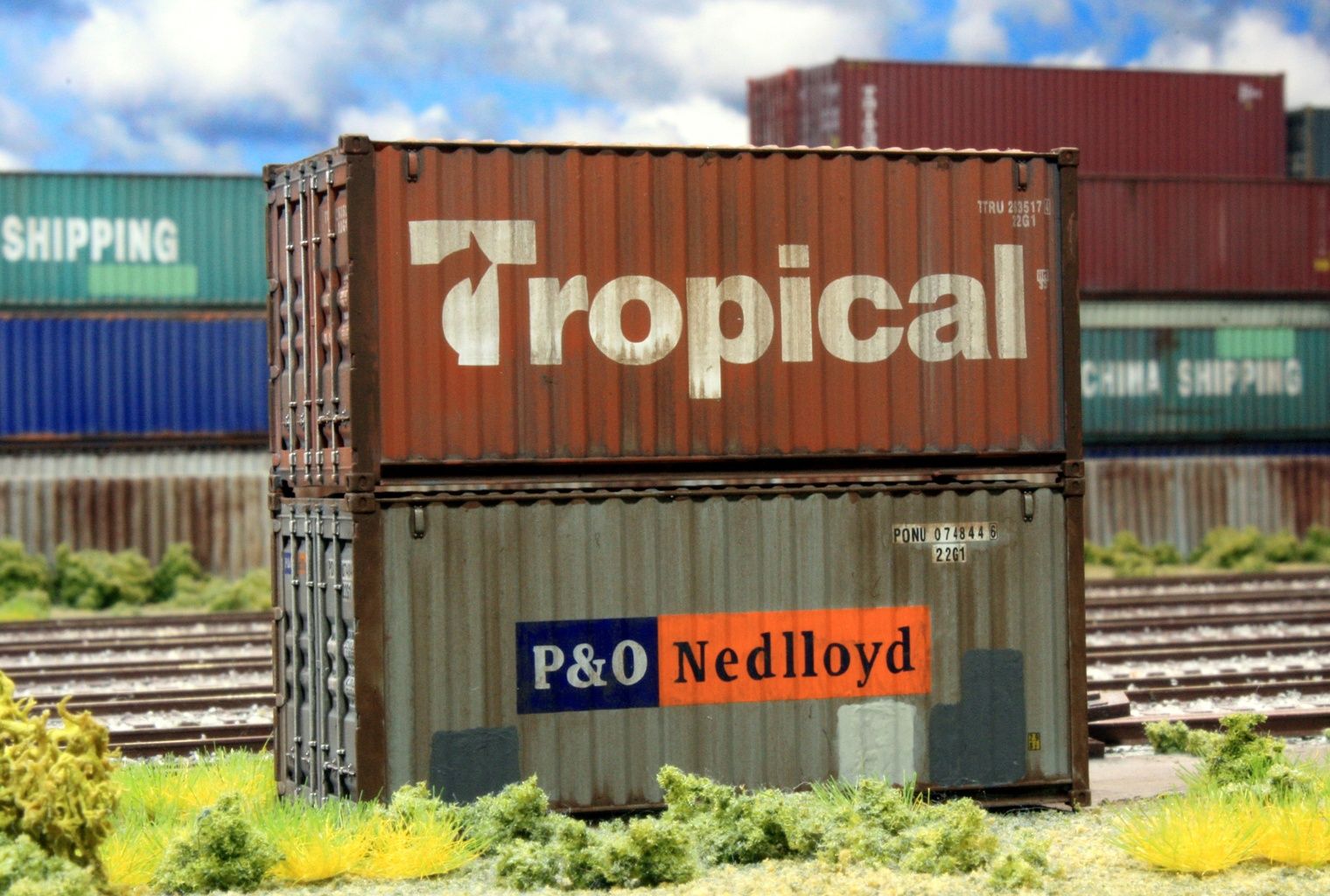} \\
  \raisebox{1 mm}{\rotatebox{90}{KK~\cite{parihar2025kontinuouskontext}}} & \vcfig{kk}{decay_011} \\
  \raisebox{1 mm}{\rotatebox{90}{SE~\cite{zarei2025slideredit}}} & \vcfig{se}{decay_011} \\
  \midrule
  & \multicolumn{1}{c|}{Original} & \multicolumn{5}{c}{Neutral $\rightarrow$ Smiling} \\
  \raisebox{5 mm}{\rotatebox{90}{Ours}} & \vcfig{ours}{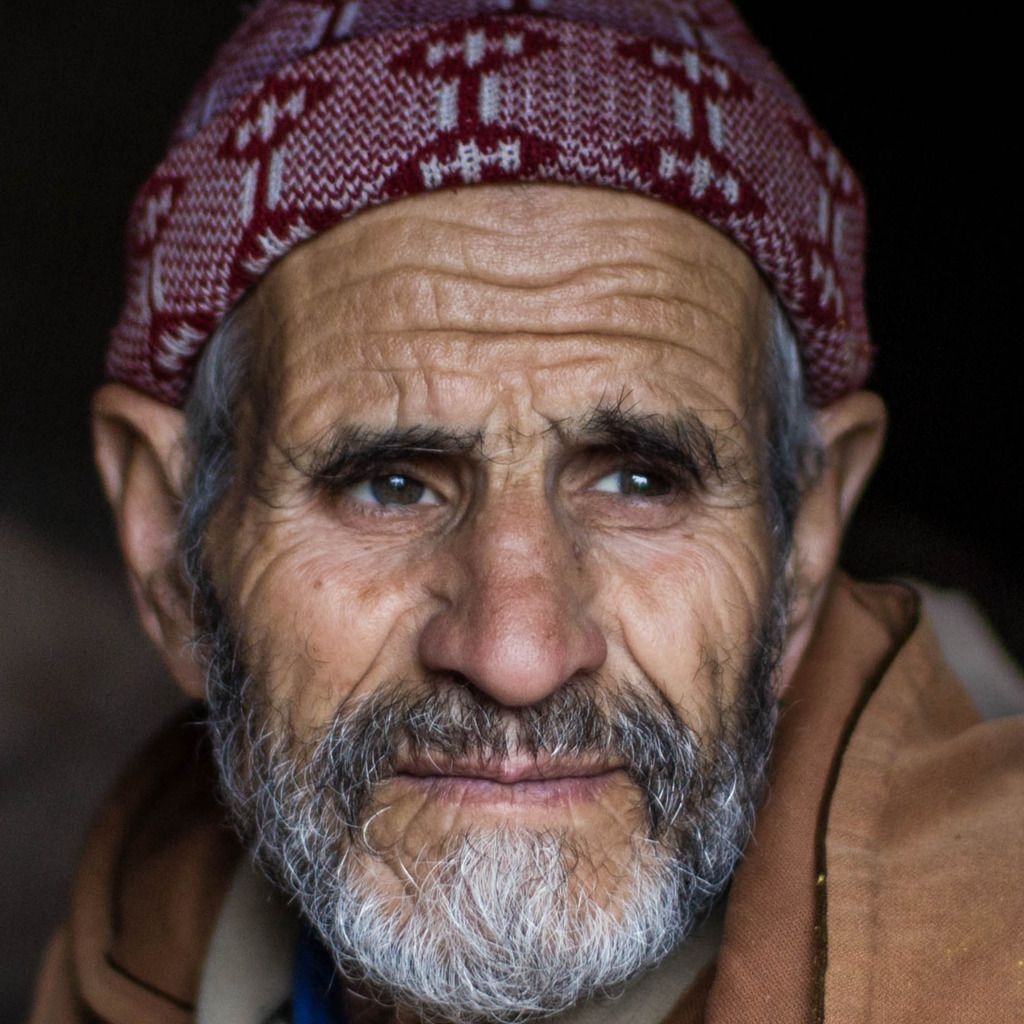} \\
  \raisebox{4 mm}{\rotatebox{90}{KK~\cite{parihar2025kontinuouskontext}}} & \vcfig{kk}{00387} \\
  \raisebox{4 mm}{\rotatebox{90}{SE~\cite{zarei2025slideredit}}} & \vcfig{se}{00387} \\
  \end{tabular}
  \vspace{-4mm}
  \caption{\textbf{Qualitative slider results.} We compare our method against Kontinuous Kontext~\cite{parihar2025kontinuouskontext} and SliderEdit~\cite{zarei2025slideredit}. Each row shows the original (leftmost) and edited outputs at increasing slider strengths (left to right).}
  \label{fig:visual_comp}
\end{figure*}

\subsection{Evaluation metrics}
\label{sec:metrics}
We evaluate methods along two axes: edit quality (does the edit achieve the intended semantic change?) and slider behavior (does a single scalar strength produce predictable, well-behaved trajectories?). All metrics are computed over a fixed grid of increasing strength values $s$ and averaged across the benchmark. Full metric definitions are provided in the appendix.

\paragraph{Edit quality.}
We measure prompt-aligned semantic change using CLIP-dir~\cite{radford2021clip,Brooks_2023_CVPR}, which evaluates whether the direction of image change from the source aligns with the direction induced by the source-to-target prompt change in CLIP embedding space. To quantify source preservation, we report the DreamSim distance~\cite{fu2023dreamsim} between edited and source images, a perceptual similarity measure designed to reflect high-level structure and identity preservation. Ideally, edited results should preserve identity and structure while aligning with the intended edit direction, reflected by a lower DreamSim score and a higher CLIP-dir score.

\paragraph{Slider behavior.}
We evaluate whether strength modulation behaves like a true slider. Monotonicity measures how consistently the edit effect increases with $s$ without reversals, capturing predictable strength ordering. Smoothness~\cite{parihar2025kontinuouskontext} measures whether intermediate strengths form a coherent gradual transition, penalizing abrupt changes between neighboring $s$ values. 

\subsection{Qualitative slider results}
\cref{fig:visual_comp} shows qualitative slider results for time and expression changes as strength $s$ increases.
Kontinuous Kontext produces minimal changes for out-of-distribution samples, as shown in the upper example. It also loses source fidelity at early strengths and shows non-smooth transitions throughout the lower example.
SliderEdit either fails to preserve source-image identity (lower example) or tends to produce synthetic-looking outputs instead of photorealistic ones (upper example). In contrast, FlowSlider achieves stable continuous control while preserving source identity.

\begin{table*}[t]
    \centering
    \caption{\textbf{Quantitative evaluation results on our continuous-editing benchmark.} We compare our method with other slider approaches along two dimensions: edit quality and slider behavior. An ideal slider method should achieve strong slider behavior while maintaining high edit quality.}
    \label{tab:quantitative}
\begin{tabular}{lcccc}
\toprule
\multicolumn{1}{c}{\multirow{2}{*}{Method}} & \multicolumn{2}{c}{Edit quality} & \multicolumn{2}{c}{Slider behavior} \\ \cline{2-5} 
\multicolumn{1}{c}{} & CLIP-dir $\uparrow$ & DreamSim $\downarrow$ & Mono $\uparrow$ & Smooth $\downarrow$ \\ \midrule
FlowEdit (CFG) & 0.113 & \underline{0.128} & 0.214 & 0.61 \\
FlowEdit ($n_{\max}$) & 0.151 & 0.266 & 0.530 & 0.61 \\
FLUX-Kontext & \textbf{0.470} & 0.269 & 0.324 & 4.02 \\
Kontinuous Kontext & 0.358 & 0.179 & 0.588 & 0.85 \\ 
SliderEdit & \underline{0.457} & 0.337 & 0.391 & 0.27 \\
\midrule
FlowSlider (SD3) & 0.436 & 0.137 & \underline{0.739} & \underline{0.04} \\
FlowSlider (FLUX.1) & 0.400 & \textbf{0.090} & \textbf{0.833} & \textbf{0.01} \\ \bottomrule
\end{tabular}
\end{table*}

\begin{table}[t]
    \centering
    \caption{\textbf{General editing evaluation on PIE-Bench~\cite{ju2023direct}.} This evaluation is included to demonstrate general performance beyond our continuous-editing benchmark. We report both edit quality and slider-behavior metrics, and FlowSlider shows trends consistent with those on our continuous-editing benchmark.}
    \label{tab:piebench_general}
    \begin{tabular}{lcccc}
    \toprule
    \multicolumn{1}{c}{\multirow{2}{*}{Method}} & \multicolumn{2}{c}{Edit quality} & \multicolumn{2}{c}{Slider behavior} \\ \cline{2-5}
    \multicolumn{1}{c}{} & CLIP-dir $\uparrow$ & DreamSim $\downarrow$ & Mono $\uparrow$ & Smooth $\downarrow$ \\ \midrule
    FlowEdit (CFG) & 0.063 & \textbf{0.045} & 0.350 & 1.64 \\
    FlowEdit ($n_{\max}$) & 0.114 & 0.161 & \underline{0.593} & 0.60 \\
    Kontinuous Kontext & 0.245 & 0.188 & 0.545 & \underline{0.27} \\ 
    SliderEdit & \textbf{0.371} & 0.290 & 0.433 & 0.30 \\
    \midrule
    FlowSlider & \underline{0.286} & \underline{0.093} & \textbf{0.672} & \textbf{0.11} \\
    \bottomrule
    \end{tabular}
\end{table}

\subsection{Quantitative evaluation}
\cref{tab:quantitative} compares edit quality and slider behavior.
FlowSlider achieves the best slider behavior and source fidelity simultaneously: it attains the highest monotonicity and the lowest smoothness error, while also yielding the lowest DreamSim and a relatively high CLIP-dir score.
Although FLUX-Kontext and SliderEdit obtain higher CLIP-dir, they do so with substantially worse source preservation and less stable slider trajectories.
In contrast, FlowEdit heuristic parameter tuning fails to provide reliable slider control: CFG scaling is weakly monotonic, and edit-window tuning $n_{\max}$ improves Mono but degrades fidelity.
These trends are consistent with our design: scaling only the steering term modulates semantic strength while keeping the fidelity correction intact.
As a general evaluation, \cref{tab:piebench_general} reports PIE-Bench results (excluding \emph{Add}/\emph{Remove}) across both edit quality and slider behavior.
Although PIE-Bench is not a true continuous-strength benchmark, FlowSlider still shows strong performance consistent with trends on our continuous-editing benchmark.

\subsection{Ablation study}
\label{sec:ablation}

\cref{fig:ablt} compares FlowSlider against two training-free alternatives for modulating edit strength within the FLUX.1-dev backbone.
Naive scaling multiplies the entire FlowEdit update by $s$: $V^{\Delta}_{\mathrm{naive}}=sV^{\Delta}=sV_{\mathrm{fid}}+sV_{\mathrm{steer}}$.
Thus, for $s>1$, it amplifies not only semantic steering but also the source-conditioned stabilizing component and residual non-cancelled components contained in $V^{\Delta}$.
These amplified residuals can accumulate during integration, push the trajectory away from the source-consistent regime where shared-noise differencing is effective, and produce visible artifacts (e.g., ringing/over-sharpening and degraded structural coherence), rather than a clean increase in edit strength.
The linear interpolation baseline combines source- and target-conditioned velocities,
$V_{\mathrm{interp}}(s)=(1-s)V_{\mathrm{src}}+sV_{\mathrm{tar}}$,
where $V_{\mathrm{src}}=V(z_t^{\mathrm{src}},t,c_{\mathrm{src}})$ and $V_{\mathrm{tar}}=V(z_t^{\mathrm{tar}},t,c_{\mathrm{tar}})$.
We extrapolate this form to $s>1$ to mimic stronger edits; however, it often causes appearance drift and inconsistent target realization.

The quantitative comparison is summarized in \cref{tab:ablation}.
Naive scaling achieves higher CLIP-dir than interpolation but severely degrades source fidelity and slider behavior, consistent with destabilized integration at larger $s$.
Linear interpolation improves smoothness but under-edits and shows limited extrapolation ability, yielding the lowest CLIP-dir and only moderate monotonicity.
In contrast, FlowSlider uses the decomposed update $V^{\Delta}_{s}=V_{\mathrm{fid}}+sV_{\mathrm{steer}}$, scaling only semantic steering while keeping the source-conditioned stabilizer fixed.
This yields the best overall ablation performance, with the strongest prompt alignment, highest source fidelity, and most reliable slider behavior.

\begin{table}[t]
    \centering
    \caption{\textbf{Quantitative ablation results.} We compare two training-free alternatives---naive scaling of the FlowEdit update ($s\,V^{\Delta}$) and linear velocity interpolation---against FlowSlider. The evaluation is conducted under the same setting as \cref{tab:quantitative}.}
    \label{tab:ablation}
    \begin{tabular}{rcccc}
    \toprule
    \multicolumn{1}{c}{\multirow{2}{*}{Method}} & \multicolumn{2}{c}{Edit quality} & \multicolumn{2}{c}{Slider behavior} \\ \cline{2-5} 
    \multicolumn{1}{c}{} & CLIP-dir $\uparrow$ & DreamSim $\downarrow$ & Mono $\uparrow$ & Smooth $\downarrow$ \\ \midrule
    Naive scaling & 0.255 & 0.423 & 0.371 & 0.31 \\
    Linear Interpolation & 0.119 & 0.342 & 0.417 & 0.04 \\ \midrule
    FlowSlider (Ours) & \textbf{0.400} & \textbf{0.090} & \textbf{0.833} & \textbf{0.01} \\ \bottomrule
    \end{tabular}
\end{table}

\newcommand{\abfig}[2]{%
\includegraphics[width=0.153\linewidth]{figs/visual_comps/original/#2} &
\includegraphics[width=0.153\linewidth]{figs/visual_comps/#1/#2_scale_1.jpg} &
\includegraphics[width=0.153\linewidth]{figs/visual_comps/#1/#2_scale_2.jpg} %
}

\begin{figure*}[t]
  \centering
  \setlength{\tabcolsep}{1pt}
  \begin{tabular}{cccc|ccc}
  &  \multicolumn{3}{c|}{Green $\rightarrow$ Snowy} & \multicolumn{3}{c}{Calm $\rightarrow$ Stormy} \\
  \raisebox{2 mm}{\rotatebox{90}{Ours}} & \abfig{ours}{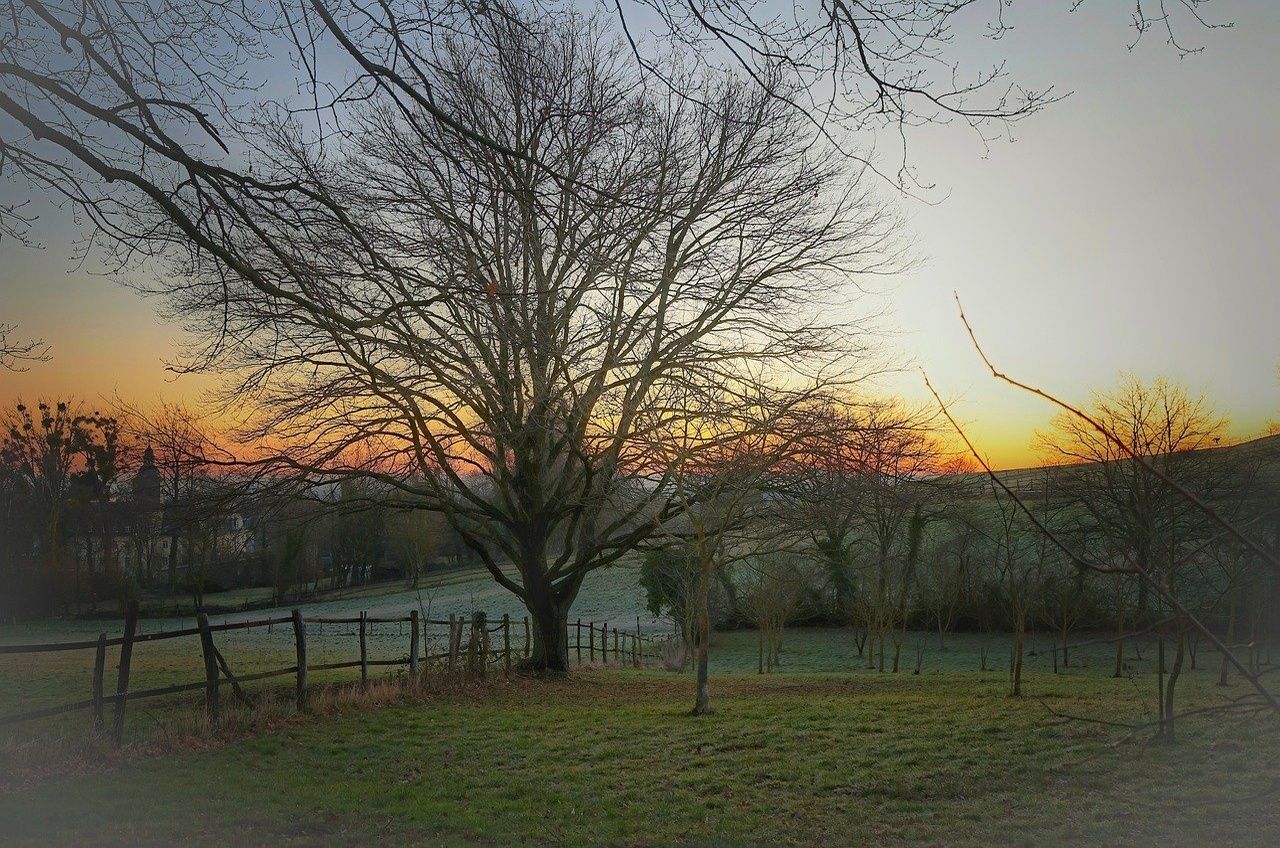} & \abfig{ours}{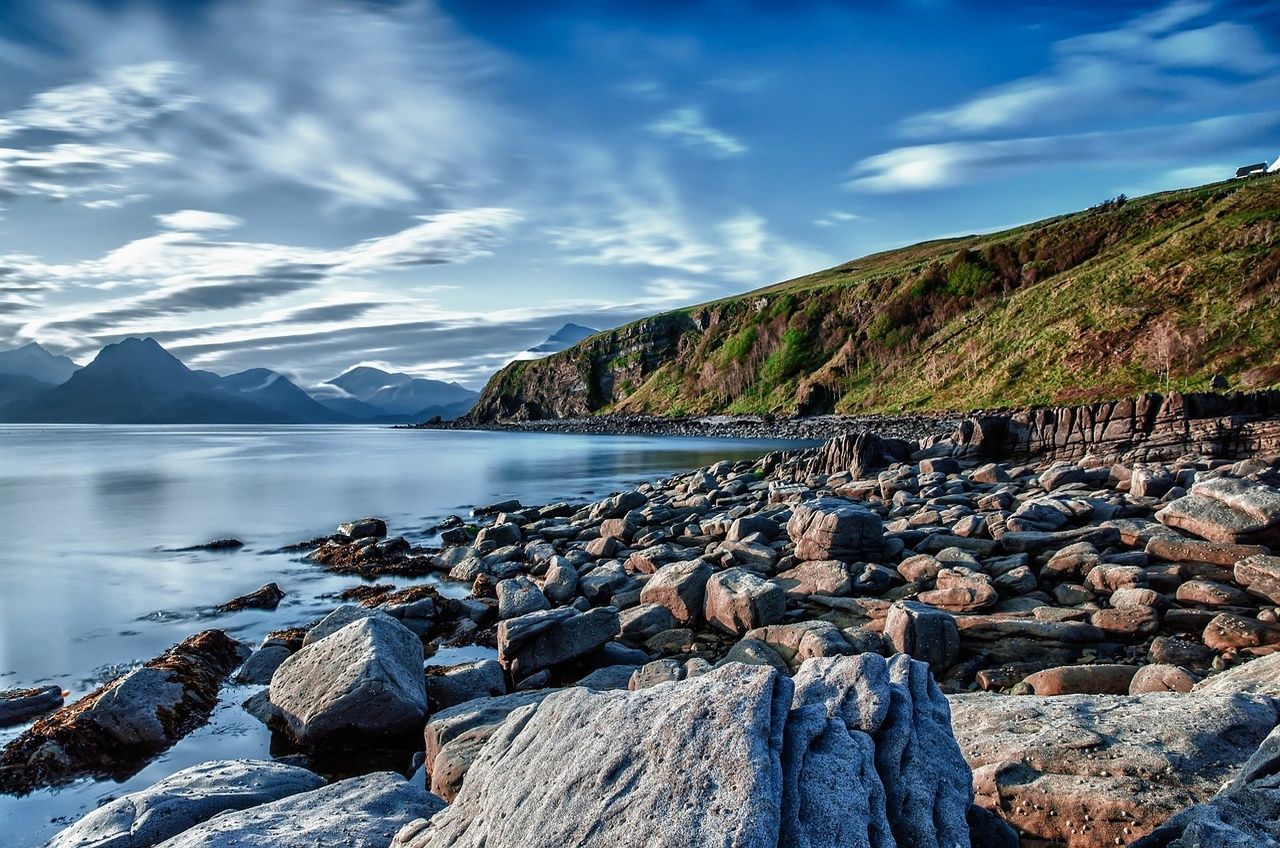} \\
  \raisebox{1 mm}{\rotatebox{90}{Scaling}} & \abfig{naive}{9448275} & \abfig{naive}{192979} \\
  \raisebox{1 mm}{\rotatebox{90}{Interp.}} & \abfig{linear}{9448275} & \abfig{linear}{192979} \\
   & Original & $s = 1$ & $s = 2$ & Original & $s = 1$ & $s = 2$ \\
  \end{tabular}
  \vspace{-3mm}
  \caption{\textbf{Ablation on strength modulation.} Naive scaling of the FlowEdit update and linear velocity interpolation lead to visible artifacts and degraded structure for $s>1$. FlowSlider remains stable across strengths by keeping the fidelity term fixed and scaling only the steering term.}
  \label{fig:ablt}
\end{figure*}

%% file: 05_conc.tex
\section{Conclusion}
\label{sec:conc}

We introduced \textit{FlowSlider}, a training-free method for continuous image editing in Rectified Flow. By decomposing FlowEdit's update into a fidelity term and a steering term, FlowSlider provides stable slider-style control of edit strength while preserving source identity and structure. Our geometric analysis and empirical evidence show that these components are weakly coupled in practice, explaining why scaling only the steering term yields predictable strength changes and avoids the drift often seen with naive global scaling. Across FLUX.1-dev and SD3 backbones, FlowSlider consistently achieves strong prompt alignment, high fidelity, and improved monotonic and smooth slider behavior without post-training or curated data. 

%% file: appendix_a.tex
\appendix
\renewcommand{\theHsection}{A\arabic{section}}

\section{Evaluation Metric Definitions}
\label{sec:app_metrics}

We provide formal definitions of all metrics used in the main paper.
Let $x_{\mathrm{src}}$ denote the source image, $x^{\mathrm{edit}}_s$ the edited image at strength~$s$,
and $(c_{\mathrm{src}}, c_{\mathrm{tar}})$ the source--target prompt pair.
We evaluate $K$ discrete strength values $s_1 < s_2 < \cdots < s_K$.

\subsection{Edit quality metrics}

\paragraph{CLIP-dir (directional CLIP similarity).}
We measure whether the image-space editing direction aligns with the text-space prompt change~\cite{radford2021clip,Brooks_2023_CVPR}.
Let $f_I$ and $f_T$ denote the $\ell_2$-normalized CLIP ViT-L/14 image and text encoders, respectively.
We define the image-space edit direction $\Delta_{\mathrm{img}} = f_I(x^{\mathrm{edit}}_s) - f_I(x_{\mathrm{src}})$
and the text-space direction $\Delta_{\mathrm{txt}} = f_T(c_{\mathrm{tar}}) - f_T(c_{\mathrm{src}})$.
CLIP-dir is their cosine similarity:
\begin{equation}
  \text{CLIP-dir} = \frac{\langle \Delta_{\mathrm{img}},\, \Delta_{\mathrm{txt}} \rangle}
  {\|\Delta_{\mathrm{img}}\|\;\|\Delta_{\mathrm{txt}}\|}.
  \label{eq:clip_dir}
\end{equation}
Higher values indicate better alignment between the edit effect and the intended prompt change.

\paragraph{DreamSim (perceptual distance).}
DreamSim~\cite{fu2023dreamsim} evaluates perceptual similarity using distances in a learned feature space. 
Given two images $x$ and $\tilde{x}$ and a feature extractor $f_\theta$, their perceptual distance is
\begin{equation}
  D(x,\tilde{x};f_\theta) = 1 - \cos\!\bigl(f_\theta(x),\, f_\theta(\tilde{x})\bigr).
\end{equation}
Accordingly, we define the DreamSim score between the source image $x_{\mathrm{src}}$ and the edited image $x^{\mathrm{edit}}_s$ as
\begin{equation}
  \text{DreamSim} = d_{\mathrm{DS}}(x_{\mathrm{src}},\, x^{\mathrm{edit}}_s)
  = D(x_{\mathrm{src}},\, x^{\mathrm{edit}}_s; f_\theta).
  \label{eq:dreamsim}
\end{equation}
Lower values indicate that the edited image remains more perceptually faithful to the source.

\subsection{Slider behavior metrics}

\paragraph{Monotonicity (Mono).}
A well-behaved slider should exhibit a non-decreasing edit effect and image change as $s$ increases.
We therefore measure the fraction of consecutive strength pairs for which both CLIP-T (text--image similarity between $x^{\mathrm{edit}}_s$ and $c_{\mathrm{tar}}$)
and DreamSim are non-decreasing:
\begin{equation}
  \mathrm{Mono}
  =
  \frac{1}{K-1}
  \sum_{k=1}^{K-1}
  \mathbf{1}\!\left[
    \mathrm{CLIP\mbox{-}T}(s_{k+1}) \ge \mathrm{CLIP\mbox{-}T}(s_k)
    \;\wedge\;
    d_{\mathrm{DS}}(s_{k+1}) \ge d_{\mathrm{DS}}(s_k)
  \right].
\end{equation}
where
$\text{CLIP-T}(s_k) = \langle f_I(x^{\mathrm{edit}}_{s_k}),\, f_T(c_{\mathrm{tar}}) \rangle$
and
$d_{\mathrm{DS}}(s_k) = d_{\mathrm{DS}}(x_{\mathrm{src}},\, x^{\mathrm{edit}}_{s_k})$.
A higher Mono score indicates that larger slider values tend to produce edits that are both more aligned with the target prompt and farther from the source image.

\paragraph{Smoothness (Smooth).}
Following Kontinuous Kontext~\cite{parihar2025kontinuouskontext}, we evaluate trajectory smoothness via the normalized triangle deficit.
For three consecutive edited images $I_k = x^{\mathrm{edit}}_{s_k}$, $I_{k+1} = x^{\mathrm{edit}}_{s_{k+1}}$, $I_{k+2} = x^{\mathrm{edit}}_{s_{k+2}}$,
the normalized deficit is:
\begin{equation}
  \tilde{\Delta}_k
  = \frac{d(I_k, I_{k+1}) + d(I_{k+1}, I_{k+2}) - d(I_k, I_{k+2})}
         {d(I_k, I_{k+2})},
  \label{eq:smooth_triplet}
\end{equation}
where $d$ denotes the LPIPS distance~\cite{zhang2018lpips}.
Smoothness is the worst-case deficit across all triplets:
\begin{equation}
  \delta_{\mathrm{smooth}} = \max_{k}\;\tilde{\Delta}_k.
  \label{eq:smooth}
\end{equation}
A lower value indicates smoother transitions; $\tilde{\Delta}_k = 0$ corresponds to a geodesic (perfectly smooth) trajectory in perceptual space.

\section{Benchmark Specification}
\label{sec:app_benchmark}

Our continuous editing benchmark consists of 250 samples from two sources: 31 samples from PIE-Bench~\cite{ju2023direct} and 219 Pixabay~\cite{pixabay} images annotated with source–target prompt pairs as well as editing instructions.
\cref{tab:benchmark_spec} shows the category breakdown. Representative samples from each category are shown in \cref{fig:benchmark_samples}.

\begin{table}[h]
  \centering
  \caption{\textbf{Benchmark specification.}
  The dataset contains 250 samples spanning 5 categories and 13 editing tasks.}
  \label{tab:benchmark_spec}
  \small
  \setlength{\tabcolsep}{4pt}
  \begin{tabular}{@{}llr@{}}
    \toprule
    Category & Editing tasks & Samples \\
    \midrule
    Scene/weather/time
      & season (49), weather (21), time of day (35) & 105 \\
    Style
      & style change (44) & 44 \\
    Degradation
      & time decay (38) & 38 \\
    Color
      & color change (18) & 18 \\
    Portrait Attributes
      & expression, hair color, makeup, aging, etc. & 45 \\
    \midrule
    Total & & 250 \\
    \bottomrule
  \end{tabular}
\end{table}

\begin{figure*}[t]
  \centering
  \setlength{\tabcolsep}{2pt}
  \small
  \begin{tabular}{ccccc}
    \textbf{Scene/weather} & \textbf{Style} & \textbf{Degradation} & \textbf{Color} & \textbf{Portrait} \\[2pt]
    \includegraphics[width=0.185\linewidth]{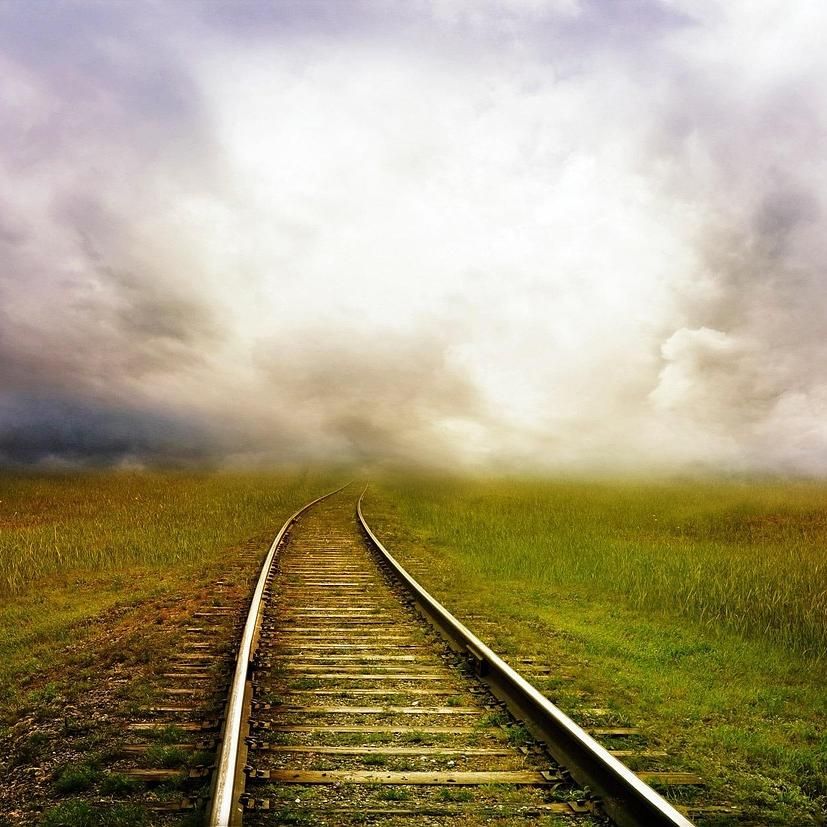} &
    \includegraphics[width=0.185\linewidth]{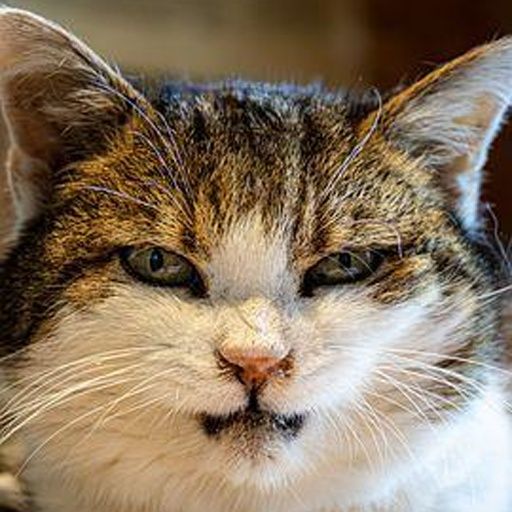} &
    \includegraphics[width=0.185\linewidth]{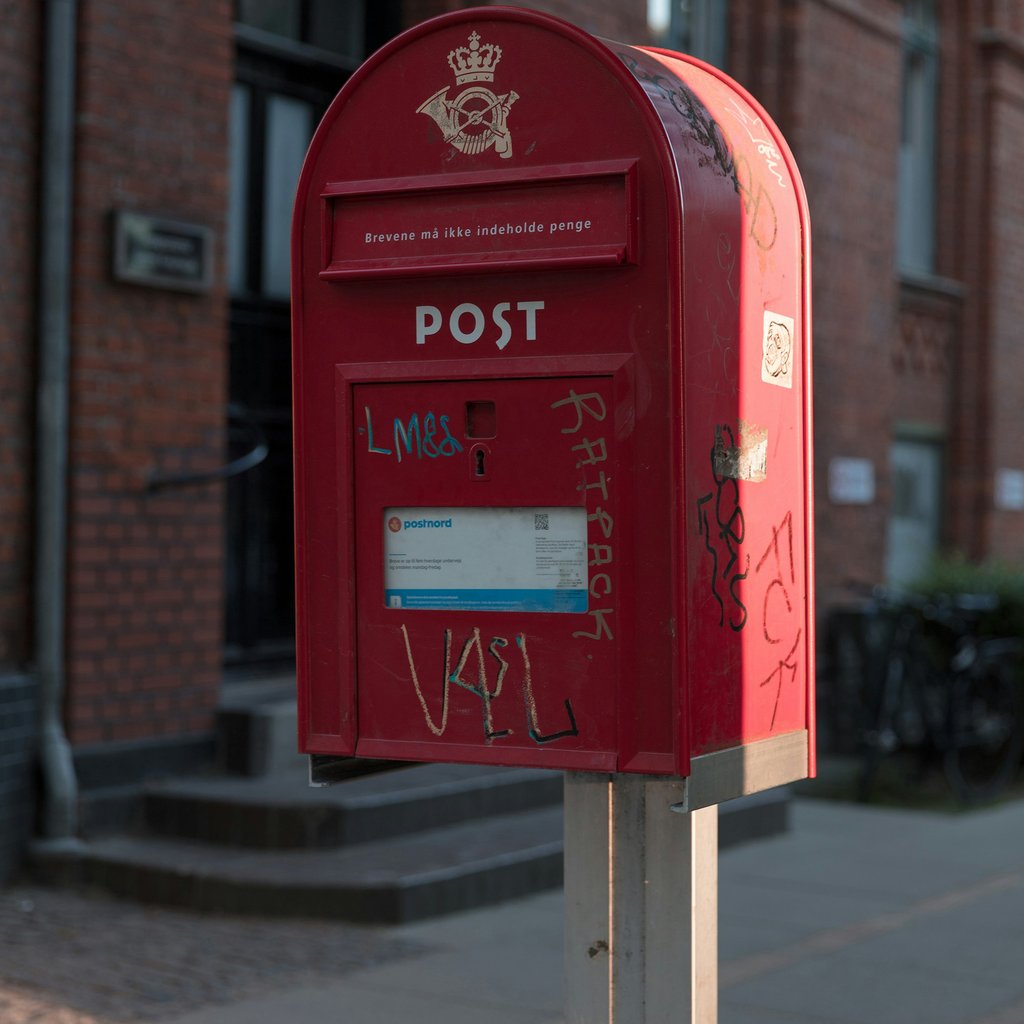} &
    \includegraphics[width=0.185\linewidth]{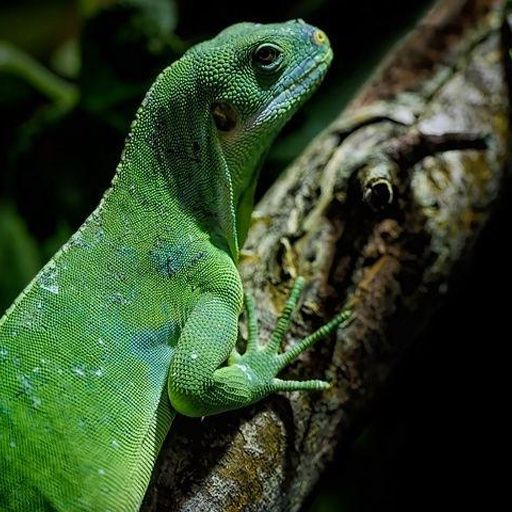} &
    \includegraphics[width=0.185\linewidth]{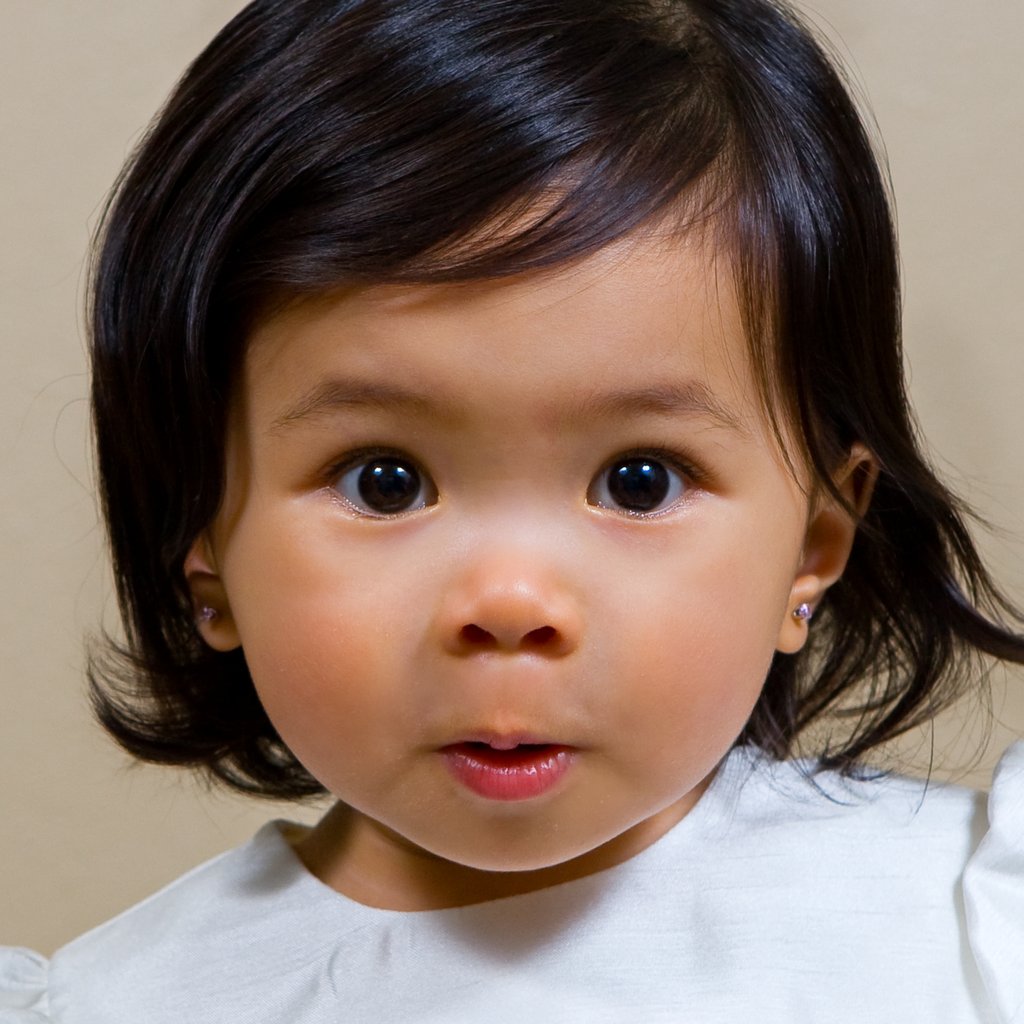} \\[2pt]
    \parbox{0.185\linewidth}{\scriptsize\centering
      \textbf{Src:} ``\dots cloudy sky with fog''\\
      \textbf{Tar:} ``\dots clear sunny blue sky''} &
    \parbox{0.185\linewidth}{\scriptsize\centering
      \textbf{Src:} ``portrait of a tabby cat\dots''\\
      \textbf{Tar:} ``pencil sketch of a tabby cat\dots''} &
    \parbox{0.185\linewidth}{\scriptsize\centering
      \textbf{Src:} ``\dots bright red mailbox\dots''\\
      \textbf{Tar:} ``\dots heavily rusted mailbox\dots''} &
    \parbox{0.185\linewidth}{\scriptsize\centering
      \textbf{Src:} ``a bright green iguana\dots''\\
      \textbf{Tar:} ``a brown iguana\dots''} &
    \parbox{0.185\linewidth}{\scriptsize\centering
      \textbf{Src:} ``\dots neutral expression\dots''\\
      \textbf{Tar:} ``\dots happy smile\dots''} \\[20pt]
    \includegraphics[width=0.185\linewidth]{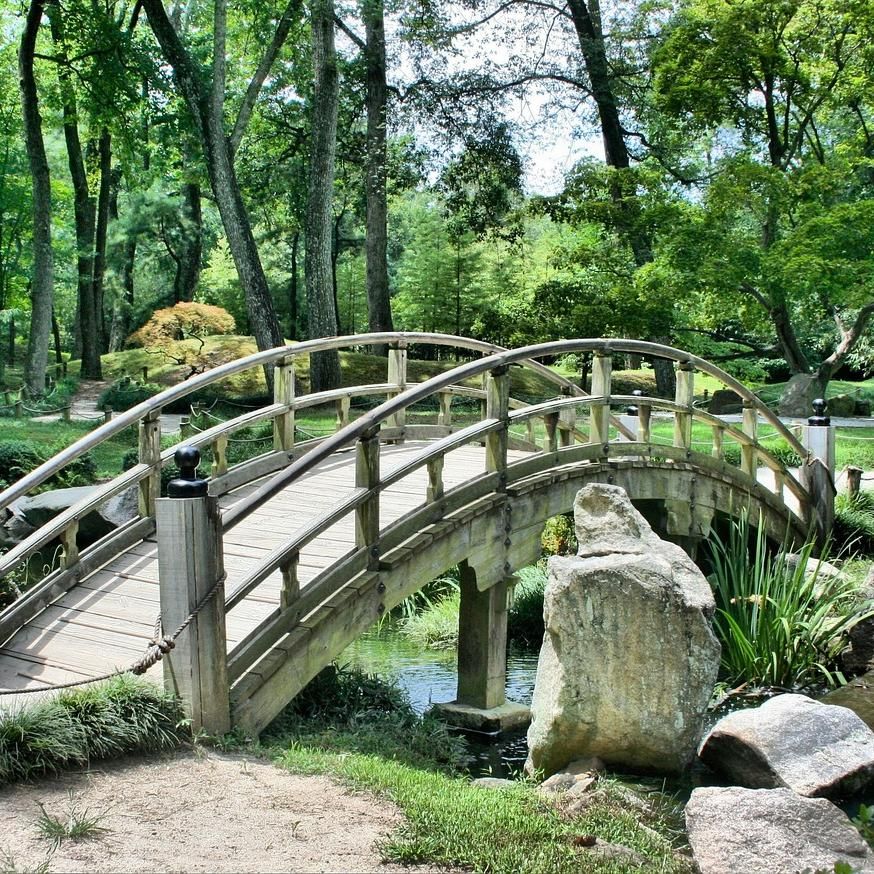} &
    \includegraphics[width=0.185\linewidth]{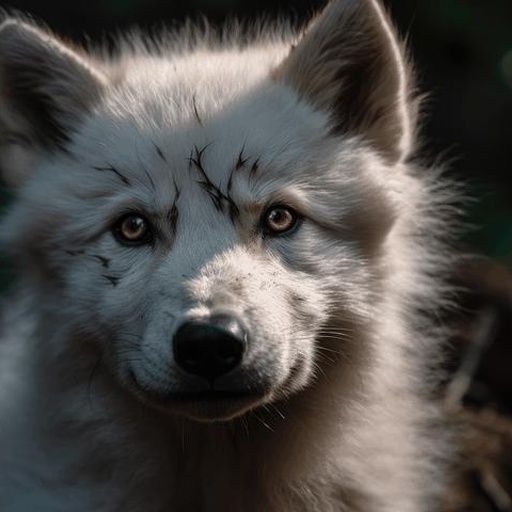} &
    \includegraphics[width=0.185\linewidth]{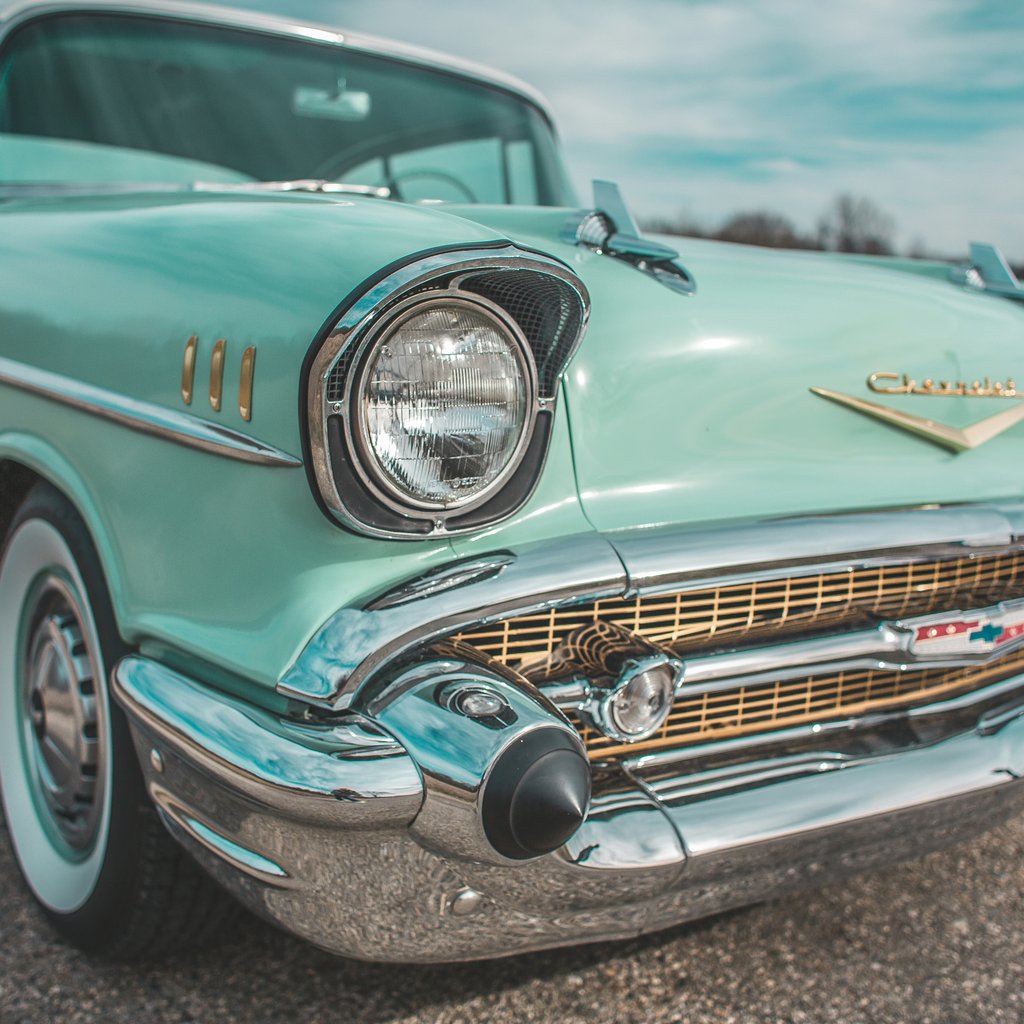} &
    \includegraphics[width=0.185\linewidth]{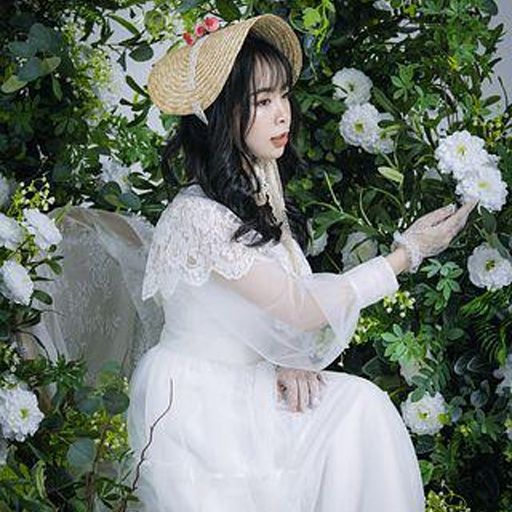} &
    \includegraphics[width=0.185\linewidth]{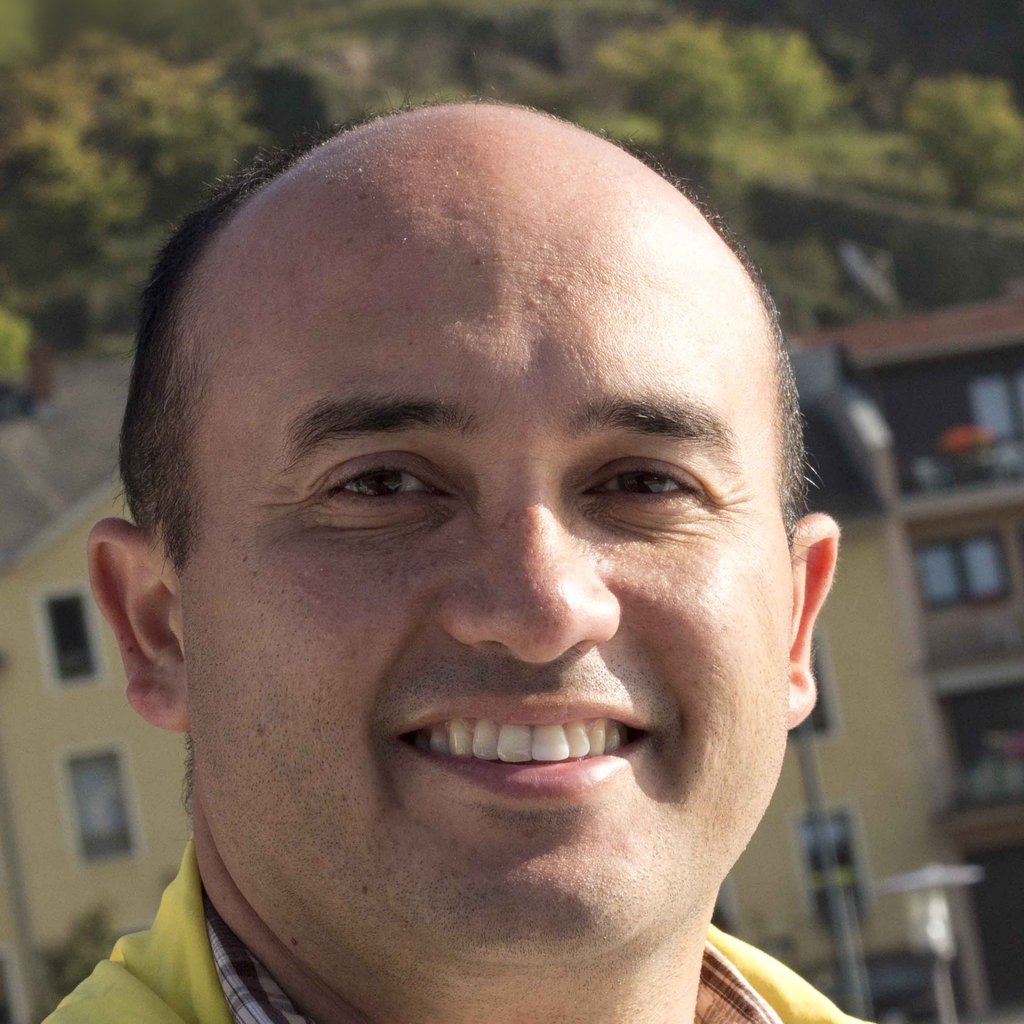} \\[2pt]
    \parbox{0.185\linewidth}{\scriptsize\centering
      \textbf{Src:} ``traditional Japanese garden\dots''\\
      \textbf{Tar:} ``\dots vibrant autumn foliage\dots''} &
    \parbox{0.185\linewidth}{\scriptsize\centering
      \textbf{Src:} ``portrait of a wolf\dots''\\
      \textbf{Tar:} ``pen and ink sketch of a wolf\dots''} &
    \parbox{0.185\linewidth}{\scriptsize\centering
      \textbf{Src:} ``\dots shiny body, clean chrome\dots''\\
      \textbf{Tar:} ``\dots rusted body, corroded\dots''} &
    \parbox{0.185\linewidth}{\scriptsize\centering
      \textbf{Src:} ``\dots wearing a white lace dress\dots''\\
      \textbf{Tar:} ``\dots wearing a red lace dress\dots''} &
    \parbox{0.185\linewidth}{\scriptsize\centering
      \textbf{Src:} ``a middle-aged bald man\dots''\\
      \textbf{Tar:} ``a young bald man\dots''}
  \end{tabular}
  \caption{\textbf{Representative benchmark samples.}
  Two examples from each of the five categories with abbreviated source--target prompt pairs.
  The benchmark contains 250 samples spanning 13 editing tasks (see \cref{tab:benchmark_spec}).}
  \label{fig:benchmark_samples}
\end{figure*}

\section{Additional Qualitative Comparisons}
\label{sec:app_qualitative}

We provide additional qualitative comparisons between FlowSlider, Kontinuous Kontext~\cite{parihar2025kontinuouskontext}, and SliderEdit~\cite{zarei2025slideredit}, supplementing qualitative comparisons of the main paper.
\cref{fig:app_qual_1,fig:app_qual_2,fig:app_qual_3,fig:app_qual_4} show further examples across all five editing categories.
Each row shows the original image (leftmost) followed by edited results at increasing slider strengths (left to right).

\newcommand{\appqc}[2]{%
\includegraphics[width=0.136\linewidth]{figs/appendix/qual_comp/original/#2} &
\includegraphics[width=0.136\linewidth]{figs/appendix/qual_comp/#1/#2_scale_1.jpg} &
\includegraphics[width=0.136\linewidth]{figs/appendix/qual_comp/#1/#2_scale_2.jpg} &
\includegraphics[width=0.136\linewidth]{figs/appendix/qual_comp/#1/#2_scale_3.jpg} &
\includegraphics[width=0.136\linewidth]{figs/appendix/qual_comp/#1/#2_scale_4.jpg} &
\includegraphics[width=0.136\linewidth]{figs/appendix/qual_comp/#1/#2_scale_5.jpg}%
}

\begin{figure*}[t]
  \centering
  \setlength{\tabcolsep}{1pt}
  \begin{tabular}{cc|ccccc}
  & \multicolumn{1}{c|}{Original} & \multicolumn{5}{c}{Autumn $\rightarrow$ Spring} \\
  \raisebox{4mm}{\rotatebox{90}{Ours}} & \appqc{ours}{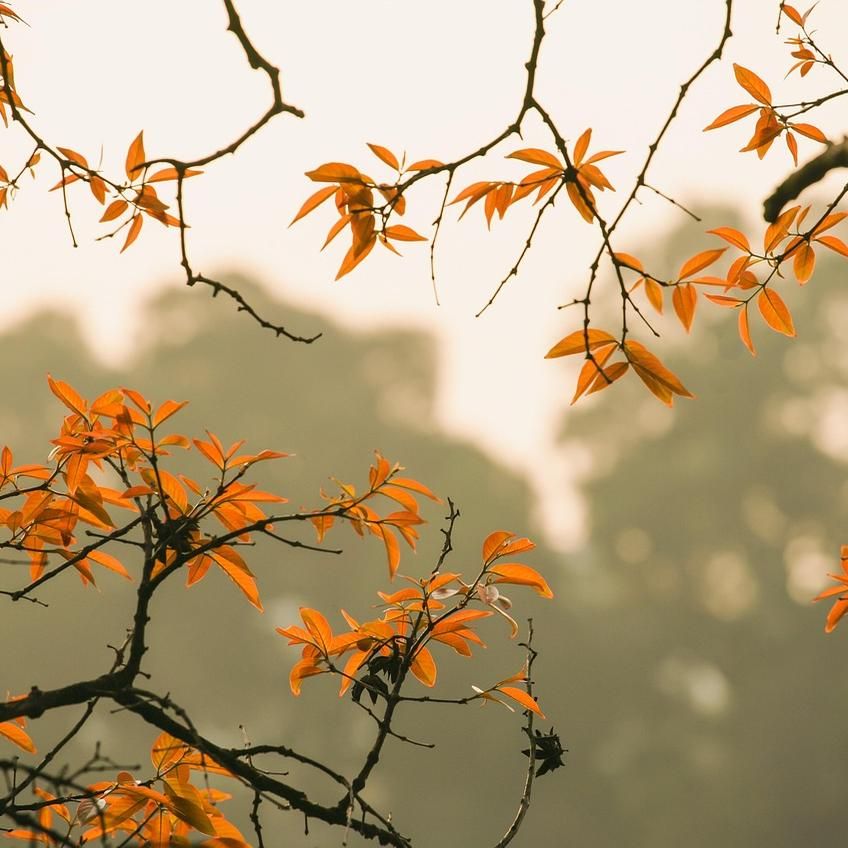} \\
  \raisebox{3mm}{\rotatebox{90}{KK~\cite{parihar2025kontinuouskontext}}} & \appqc{kk}{pixabay_images_scenery_6631518_01} \\
  \raisebox{3mm}{\rotatebox{90}{SE~\cite{zarei2025slideredit}}} & \appqc{se}{pixabay_images_scenery_6631518_01} \\
  \midrule
  & \multicolumn{1}{c|}{Original} & \multicolumn{5}{c}{Clean $\rightarrow$ Decayed} \\
  \raisebox{4mm}{\rotatebox{90}{Ours}} & \appqc{ours}{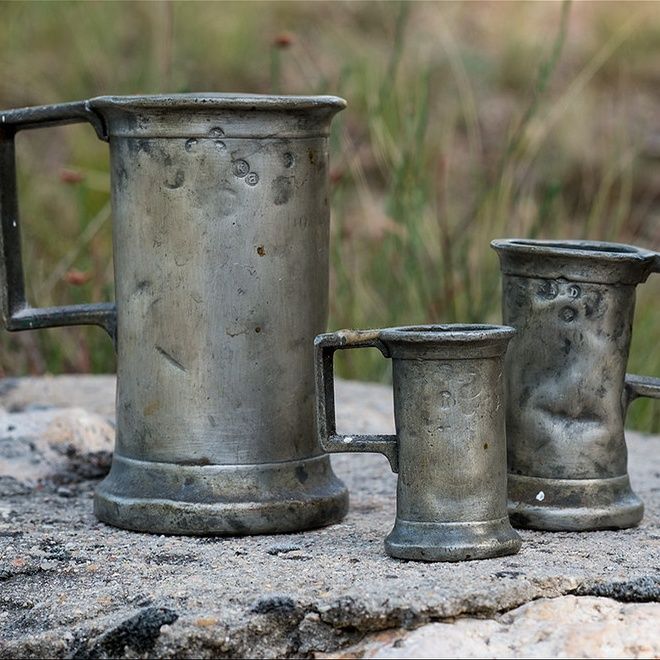} \\
  \raisebox{3mm}{\rotatebox{90}{KK~\cite{parihar2025kontinuouskontext}}} & \appqc{kk}{decay_decay_028} \\
  \raisebox{4mm}{\rotatebox{90}{SE~\cite{zarei2025slideredit}}} & \appqc{se}{decay_decay_028} \\
  \midrule
  & \multicolumn{1}{c|}{Original} & \multicolumn{5}{c}{Color $\rightarrow$ Monochrome} \\
  \raisebox{4mm}{\rotatebox{90}{Ours}} & \appqc{ours}{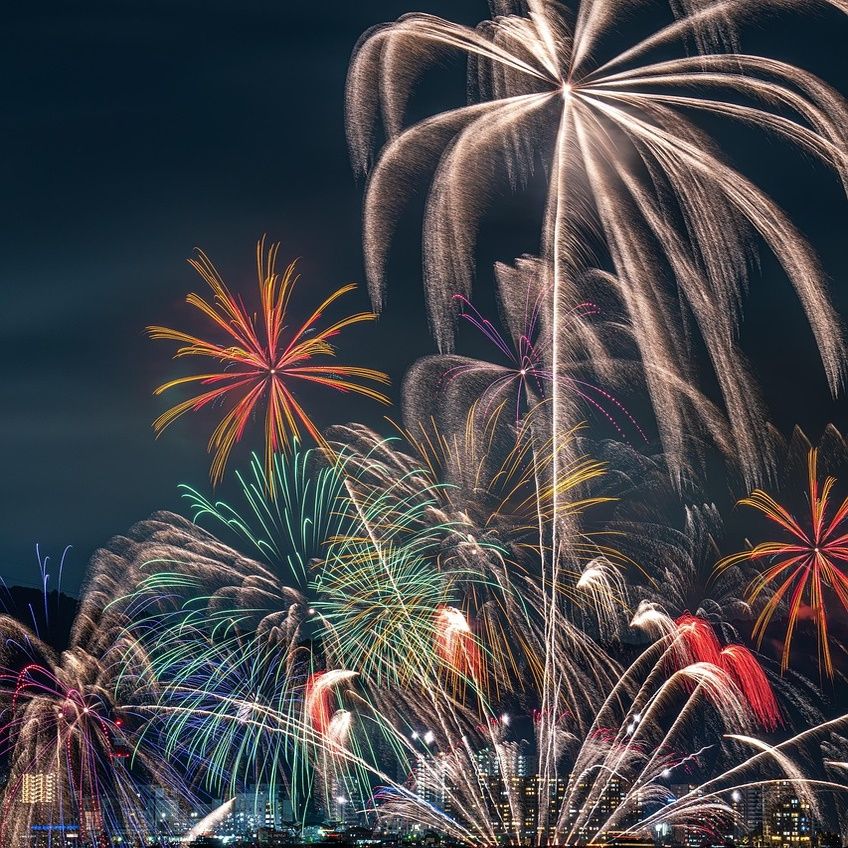} \\
  \raisebox{3mm}{\rotatebox{90}{KK~\cite{parihar2025kontinuouskontext}}} & \appqc{kk}{pixabay_images_scenery_8182800_01} \\
  \raisebox{4mm}{\rotatebox{90}{SE~\cite{zarei2025slideredit}}} & \appqc{se}{pixabay_images_scenery_8182800_01} \\
  \end{tabular}
  \vspace{-2mm}
  \caption{\textbf{Additional qualitative comparisons (1/4).}
  Each row shows the original (leftmost) and edited outputs at increasing slider strengths (left to right).
  Images are center-cropped to square for display.}
  \label{fig:app_qual_1}
\end{figure*}

\begin{figure*}[t]
  \centering
  \setlength{\tabcolsep}{1pt}
  \begin{tabular}{cc|ccccc}
  & \multicolumn{1}{c|}{Original} & \multicolumn{5}{c}{Pouty $\rightarrow$ Smiling} \\
  \raisebox{4mm}{\rotatebox{90}{Ours}} & \appqc{ours}{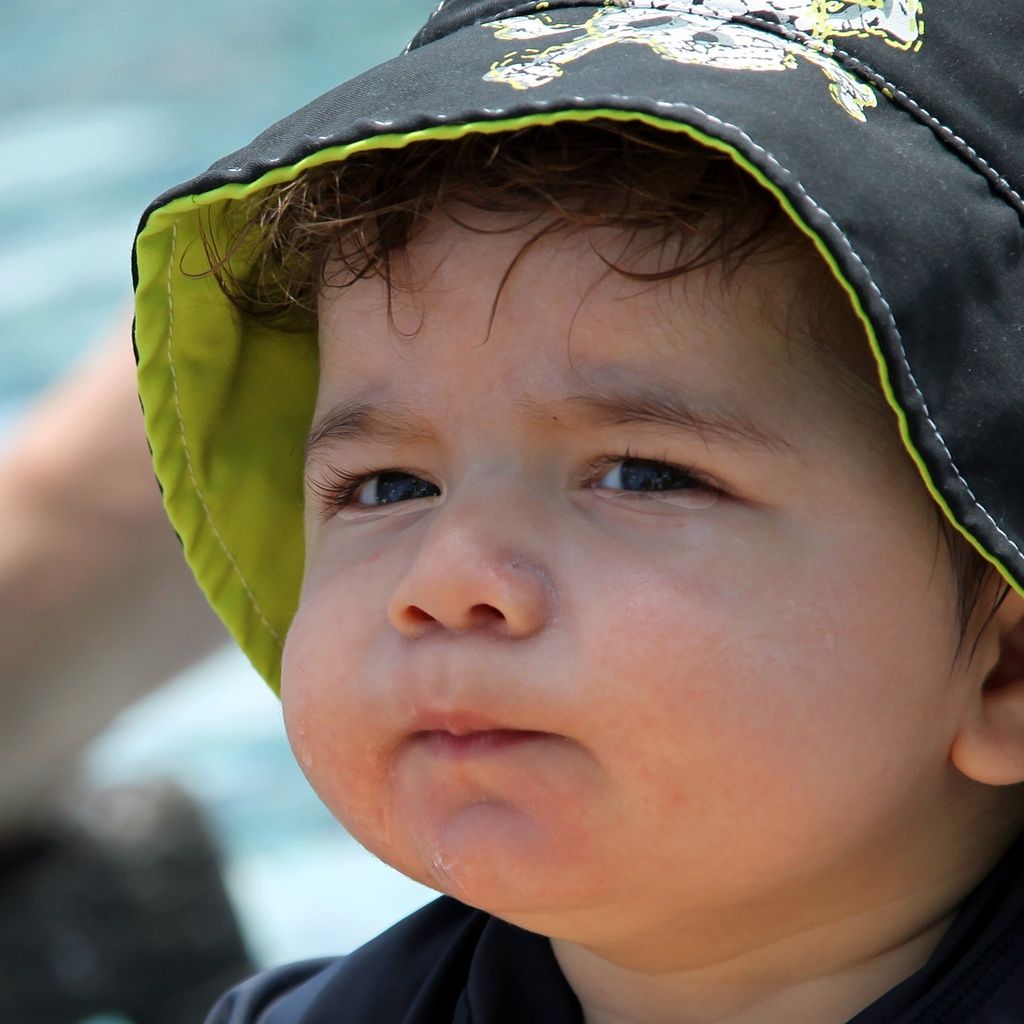} \\
  \raisebox{3mm}{\rotatebox{90}{KK~\cite{parihar2025kontinuouskontext}}} & \appqc{kk}{portrait_00070_01} \\
  \raisebox{4mm}{\rotatebox{90}{SE~\cite{zarei2025slideredit}}} & \appqc{se}{portrait_00070_01} \\
  \midrule
  & \multicolumn{1}{c|}{Original} & \multicolumn{5}{c}{Cloudy $\rightarrow$ Sunny} \\
  \raisebox{4mm}{\rotatebox{90}{Ours}} & \appqc{ours}{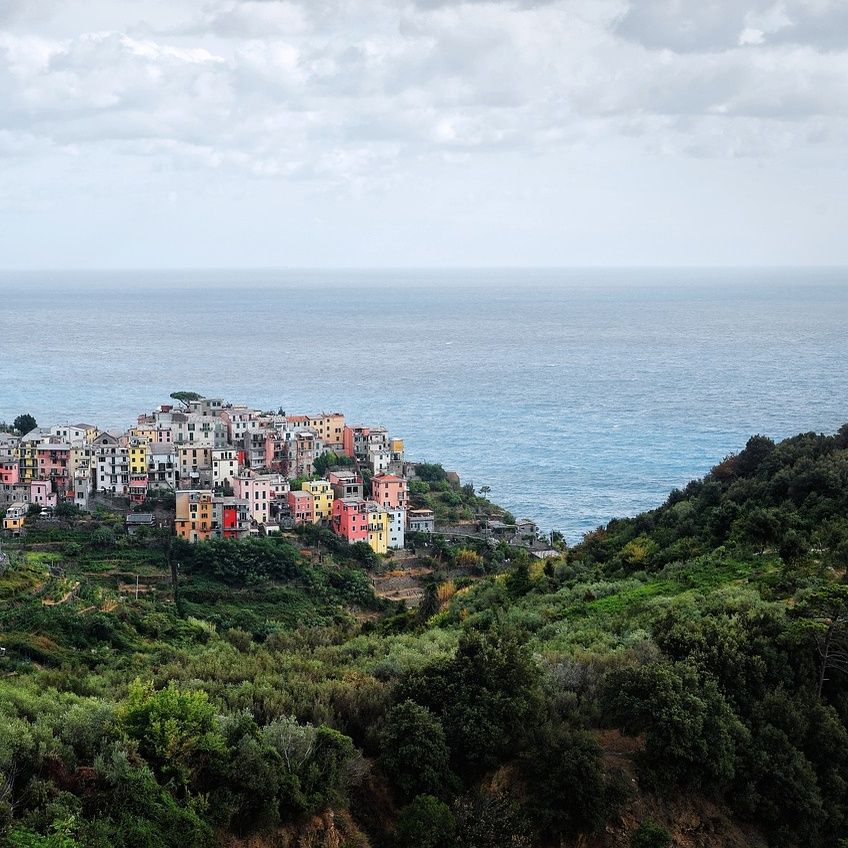} \\
  \raisebox{3mm}{\rotatebox{90}{KK~\cite{parihar2025kontinuouskontext}}} & \appqc{kk}{pixabay_images_scenery_4237666_01} \\
  \raisebox{4mm}{\rotatebox{90}{SE~\cite{zarei2025slideredit}}} & \appqc{se}{pixabay_images_scenery_4237666_01} \\
  \midrule
  & \multicolumn{1}{c|}{Original} & \multicolumn{5}{c}{Clean $\rightarrow$ Decayed} \\
  \raisebox{4mm}{\rotatebox{90}{Ours}} & \appqc{ours}{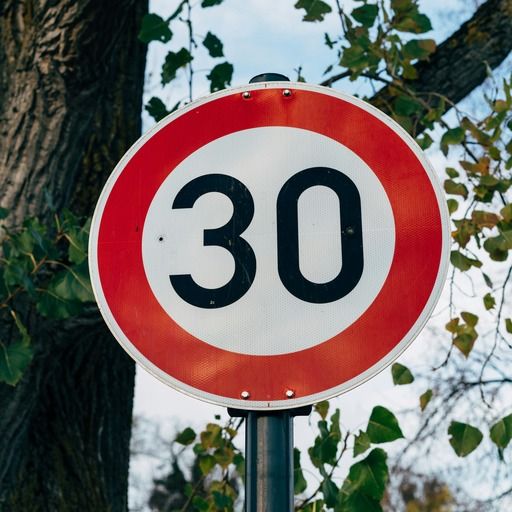} \\
  \raisebox{3mm}{\rotatebox{90}{KK~\cite{parihar2025kontinuouskontext}}} & \appqc{kk}{decay_decay_007} \\
  \raisebox{3mm}{\rotatebox{90}{SE~\cite{zarei2025slideredit}}} & \appqc{se}{decay_decay_007} \\
  \end{tabular}
  \vspace{-2mm}
  \caption{\textbf{Additional qualitative comparisons (2/4).}
  Same layout as \cref{fig:app_qual_1}.}
  \label{fig:app_qual_2}
\end{figure*}

\begin{figure*}[t]
  \centering
  \setlength{\tabcolsep}{1pt}
  \begin{tabular}{cc|ccccc}
  & \multicolumn{1}{c|}{Original} & \multicolumn{5}{c}{Black Chair $\rightarrow$ Blue Chair} \\
  \raisebox{4mm}{\rotatebox{90}{Ours}} & \appqc{ours}{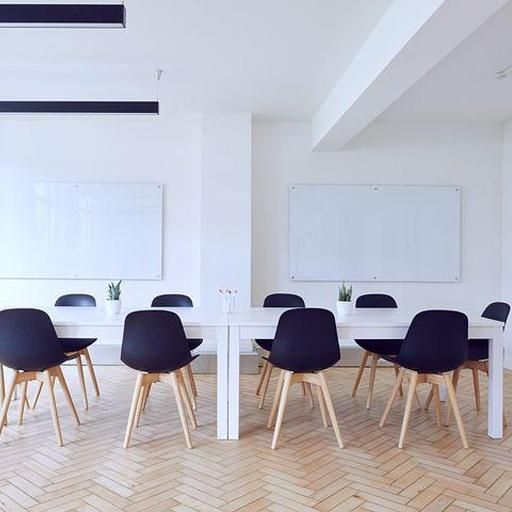} \\
  \raisebox{3mm}{\rotatebox{90}{KK~\cite{parihar2025kontinuouskontext}}} & \appqc{kk}{pie_000000000049} \\
  \raisebox{3mm}{\rotatebox{90}{SE~\cite{zarei2025slideredit}}} & \appqc{se}{pie_000000000049} \\
  \midrule
  & \multicolumn{1}{c|}{Original} & \multicolumn{5}{c}{Summer $\rightarrow$ Autumn} \\
  \raisebox{4mm}{\rotatebox{90}{Ours}} & \appqc{ours}{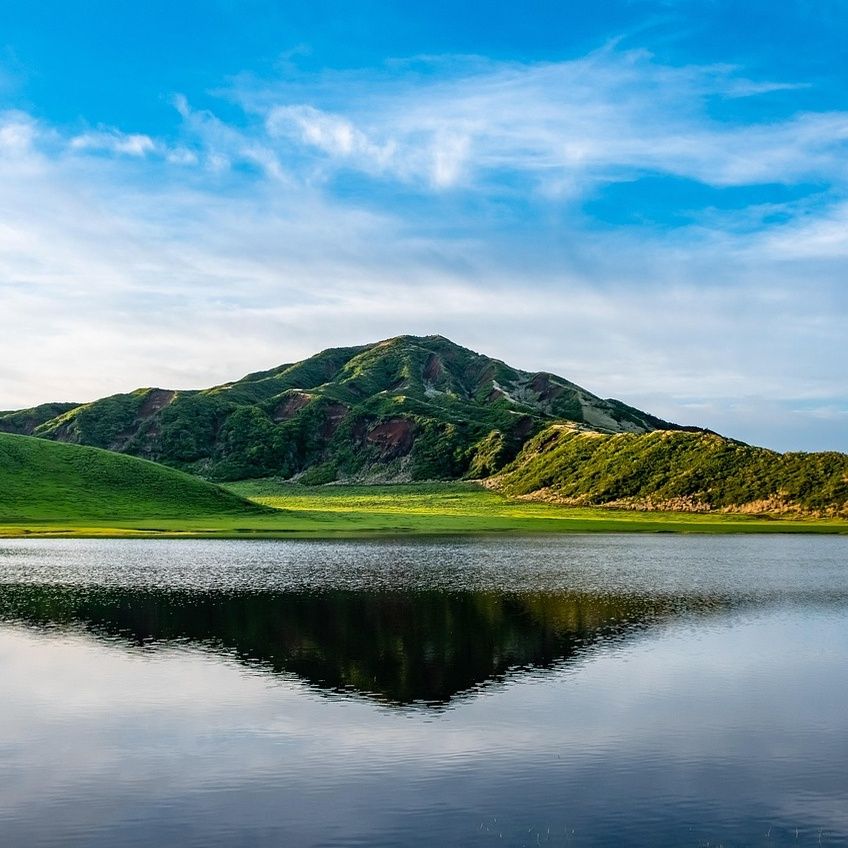} \\
  \raisebox{3mm}{\rotatebox{90}{KK~\cite{parihar2025kontinuouskontext}}} & \appqc{kk}{pixabay_images_scenery_6310371_01} \\
  \raisebox{3mm}{\rotatebox{90}{SE~\cite{zarei2025slideredit}}} & \appqc{se}{pixabay_images_scenery_6310371_01} \\
  \midrule
  & \multicolumn{1}{c|}{Original} & \multicolumn{5}{c}{Smiling $\rightarrow$ Angry} \\
  \raisebox{4mm}{\rotatebox{90}{Ours}} & \appqc{ours}{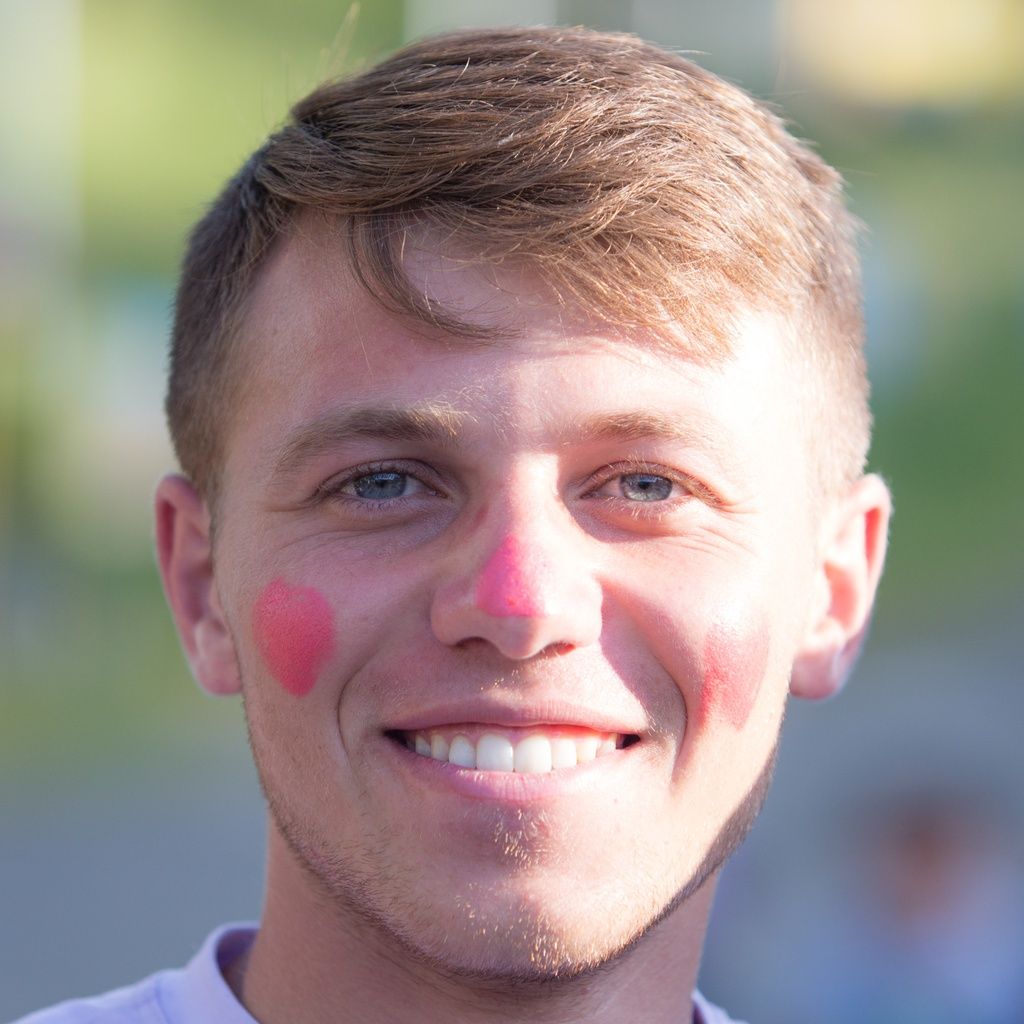} \\
  \raisebox{3mm}{\rotatebox{90}{KK~\cite{parihar2025kontinuouskontext}}} & \appqc{kk}{portrait_00056_01} \\
  \raisebox{3mm}{\rotatebox{90}{SE~\cite{zarei2025slideredit}}} & \appqc{se}{portrait_00056_01} \\
  \end{tabular}
  \vspace{-2mm}
  \caption{\textbf{Additional qualitative comparisons (3/4).}
  Same layout as \cref{fig:app_qual_1}.}
  \label{fig:app_qual_3}
\end{figure*}

\begin{figure*}[t]
  \centering
  \setlength{\tabcolsep}{1pt}
  \begin{tabular}{cc|ccccc}
  & \multicolumn{1}{c|}{Original} & \multicolumn{5}{c}{Photorealistic $\rightarrow$ Oil Painting} \\
  \raisebox{4mm}{\rotatebox{90}{Ours}} & \appqc{ours}{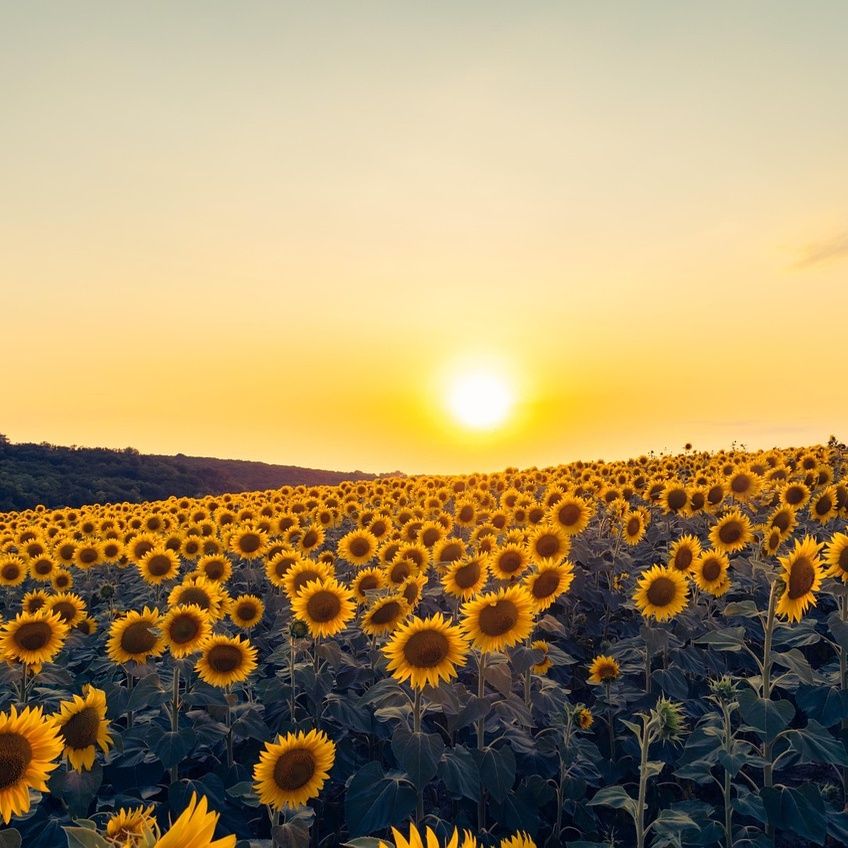} \\
  \raisebox{3mm}{\rotatebox{90}{KK~\cite{parihar2025kontinuouskontext}}} & \appqc{kk}{pixabay_images_scenery_8980921_02} \\
  \raisebox{3mm}{\rotatebox{90}{SE~\cite{zarei2025slideredit}}} & \appqc{se}{pixabay_images_scenery_8980921_02} \\
  \midrule
  & \multicolumn{1}{c|}{Original} & \multicolumn{5}{c}{Clean $\rightarrow$ Decayed} \\
  \raisebox{4mm}{\rotatebox{90}{Ours}} & \appqc{ours}{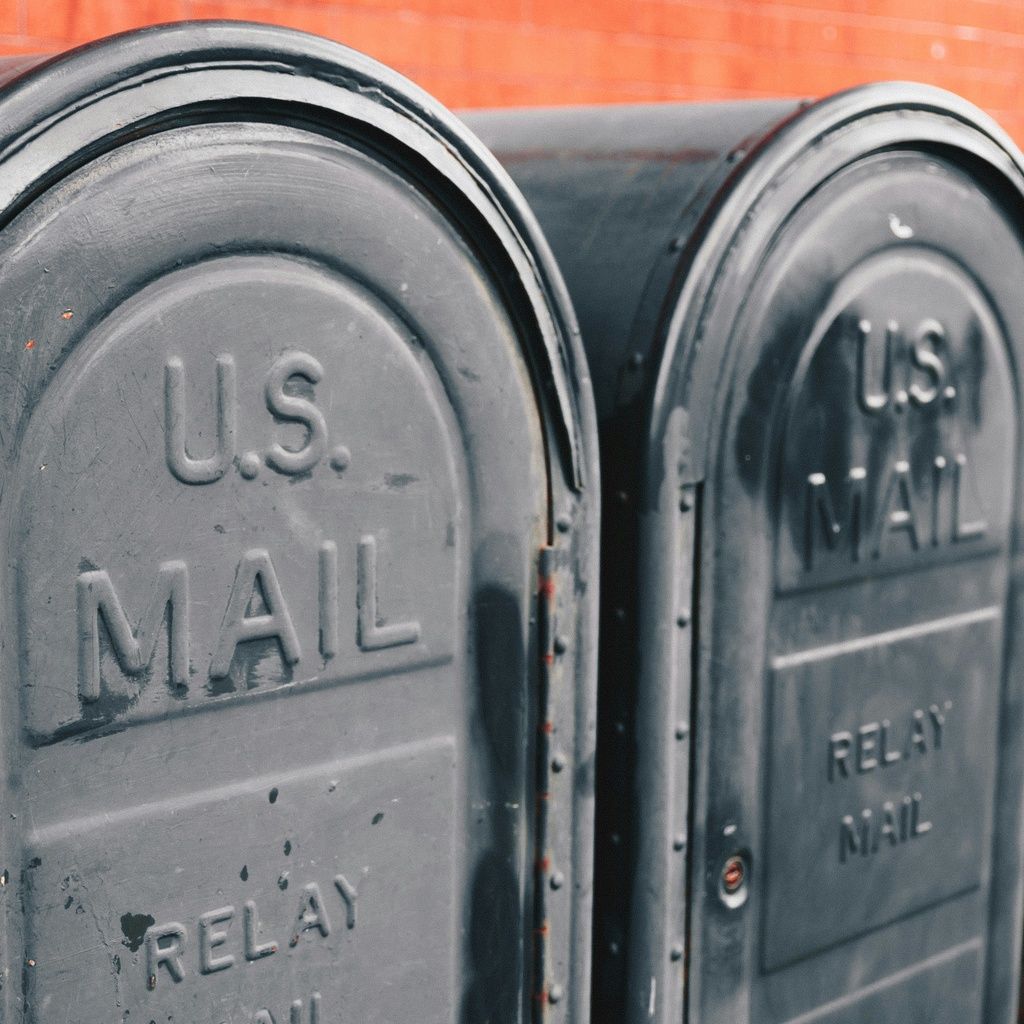} \\
  \raisebox{3mm}{\rotatebox{90}{KK~\cite{parihar2025kontinuouskontext}}} & \appqc{kk}{decay_decay_041} \\
  \raisebox{3mm}{\rotatebox{90}{SE~\cite{zarei2025slideredit}}} & \appqc{se}{decay_decay_041} \\
  \midrule
  & \multicolumn{1}{c|}{Original} & \multicolumn{5}{c}{White Hair $\rightarrow$ Black} \\
  \raisebox{4mm}{\rotatebox{90}{Ours}} & \appqc{ours}{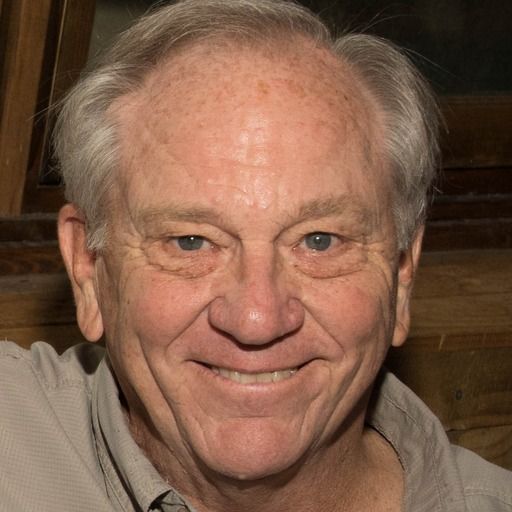} \\
  \raisebox{3mm}{\rotatebox{90}{KK~\cite{parihar2025kontinuouskontext}}} & \appqc{kk}{portrait_00144_02} \\
  \raisebox{3mm}{\rotatebox{90}{SE~\cite{zarei2025slideredit}}} & \appqc{se}{portrait_00144_02} \\
  \end{tabular}
  \vspace{-2mm}
  \caption{\textbf{Additional qualitative comparisons (4/4).}
  Same layout as \cref{fig:app_qual_1}.}
  \label{fig:app_qual_4}
\end{figure*}

\section{Fidelity--Edit Strength Trade-off}
\label{sec:app_tradeoff}

\begin{figure}[htbp]
  \centering
  \includegraphics[width=\linewidth]{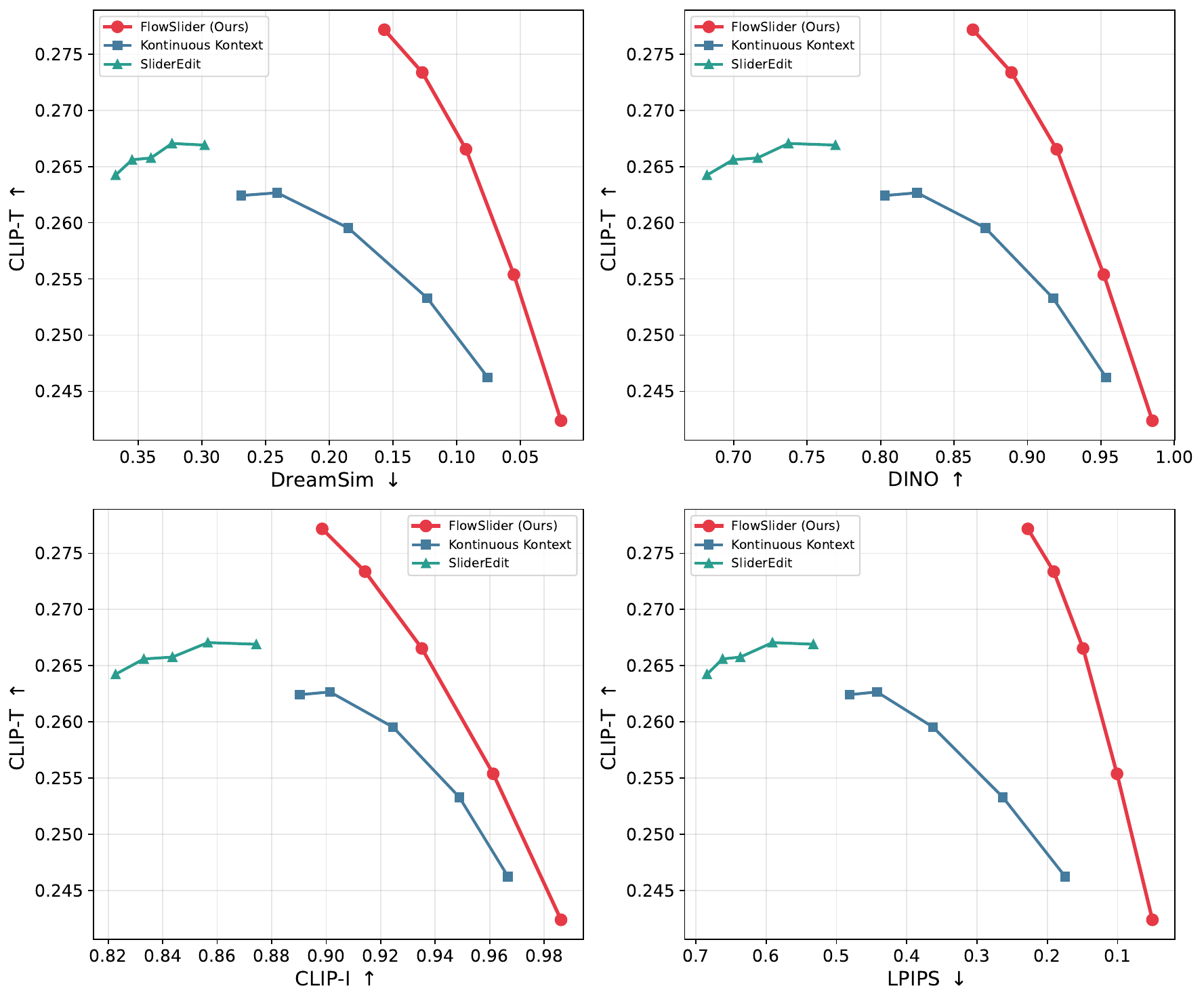}
  \caption{\textbf{Fidelity--edit strength trade-off.}
  CLIP-T (edit effect) vs.\ four preservation metrics across slider strengths.
  FlowSlider achieves higher edit effect while better preserving source consistency compared to Kontinuous Kontext~\cite{parihar2025kontinuouskontext} and SliderEdit~\cite{zarei2025slideredit}, occupying the upper-left (better) region in all four panels.}
  \label{fig:tradeoff_curve}
\end{figure}

\cref{fig:tradeoff_curve} plots the trade-off between edit effect (CLIP-T) and four source-preservation metrics as slider strength increases. Each point corresponds to a fixed strength value, averaged across the benchmark.
An ideal method occupies the upper-left region: high edit effect while maintaining strong source consistency.

As shown in \cref{fig:tradeoff_curve}, FlowSlider consistently occupies the upper-left region across all four metrics, achieving higher edit effect (CLIP-T) while better preserving source identity and structure.
Notably, even at the strongest setting ($s{=}5$), FlowSlider preserves source fidelity better than Kontinuous Kontext and SliderEdit at their weakest settings, indicating a fundamentally more favorable trade-off.
Furthermore, FlowSlider exhibits a gentler slope: as CLIP-T increases, the degradation in preservation metrics is markedly slower compared to the baselines.
This reflects the benefit of scaling only the steering term while keeping the fidelity term fixed, which limits fidelity loss per unit of semantic change.
The smooth, evenly spaced progression of FlowSlider's curve further confirms stable and predictable strength modulation.

\section{Failure Cases}
\label{sec:app_failure}

We identify two failure modes of FlowSlider.

\paragraph{Structural collapse at extreme strengths.}
A limitation of our method appears at large editing strengths. Since the fidelity term and the steering term are not perfectly orthogonal, excessively increasing $s$ allows the scaled steering term to interfere with the fixed fidelity term, resulting in artifacts such as structural collapse and over-saturation. As illustrated in \cref{fig:failure_extreme}, for the gradual editing example (``shiny compass $\to$ rusted compass''), the output remains plausible up to about $s \approx 4$, but degrades progressively for larger values, with severe artifacts at $s = 10$. We therefore recommend using a moderate range, $s \in [0,5]$, in practice. We believe this behavior reflects a limitation of the backbone in disentangling semantically opposite yet closely related prompts; using a larger backbone may improve robustness at higher editing strengths.

\paragraph{Early saturation for discrete concept changes.}
FlowSlider is designed for \emph{gradual} attribute editing (e.g., aging, season, degradation) where intermediate strengths are semantically meaningful.
For discrete concept changes such as ``rabbit $\to$ cat,'' the target concept is already fully realized at $s{=}1$, leaving no meaningful direction for further scaling.
In this regime, additional steering no longer drives a semantically coherent change; instead, the excess signal manifests as color shifts and over-sharpening artifacts, as shown in \cref{fig:failure_discrete}.

These two failure modes suggest that the effective range of $s$ is task-dependent: gradual attribute changes (e.g., aging, weathering) tolerate larger $s$, whereas discrete concept replacements saturate early.
In practice, selecting an appropriate $s$ range for a given editing task is important for obtaining reliable results.

\newcommand{\failimg}[1]{\includegraphics[width=0.155\linewidth]{#1}}

\begin{figure*}[t]
  \centering
  \setlength{\tabcolsep}{1pt}
  \small
  \begin{tabular}{@{}cccccc@{}}
    Original & $s{=}2$ & $s{=}4$ & $s{=}6$ & $s{=}8$ & $s{=}10$ \\[2pt]
    \failimg{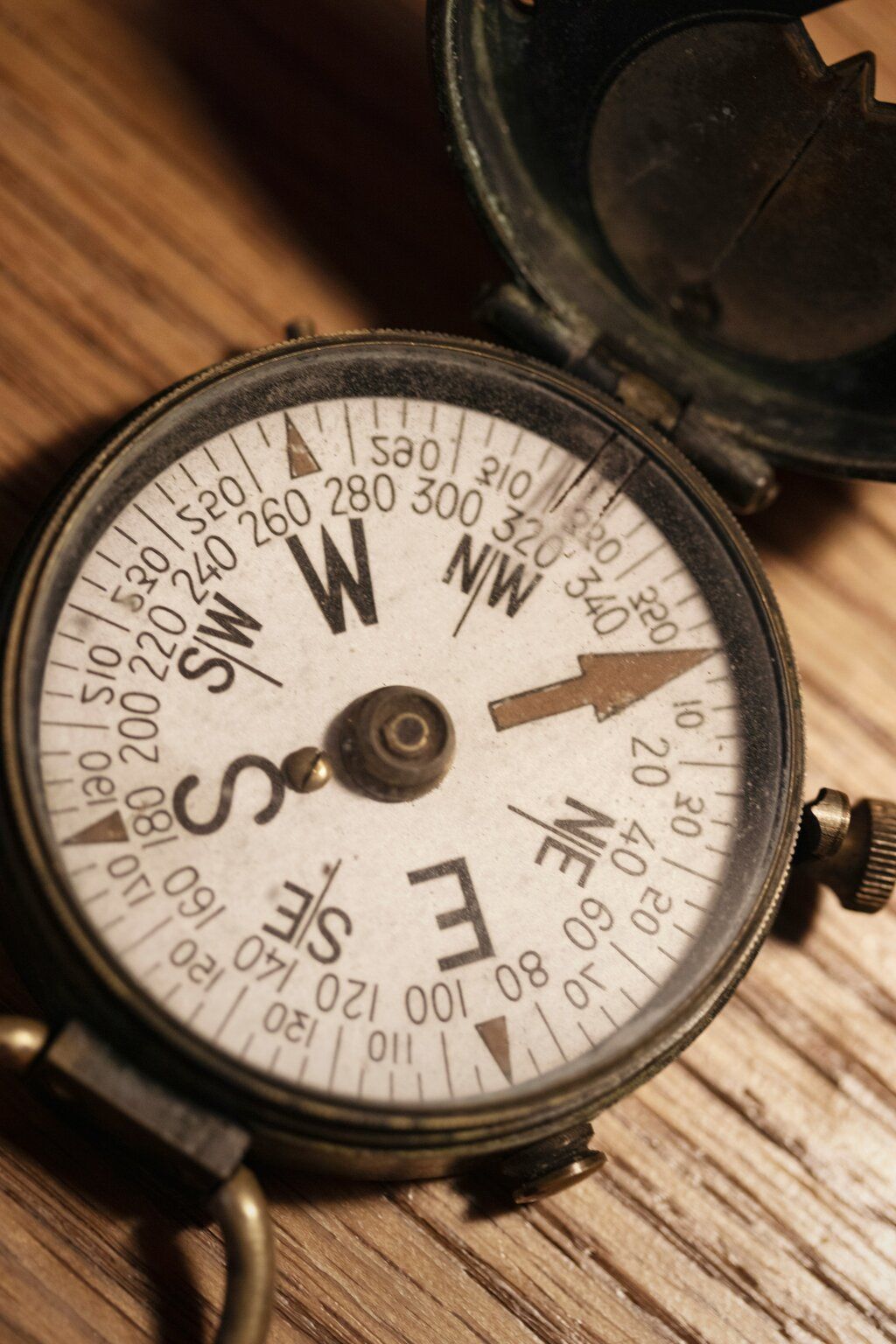} &
    \failimg{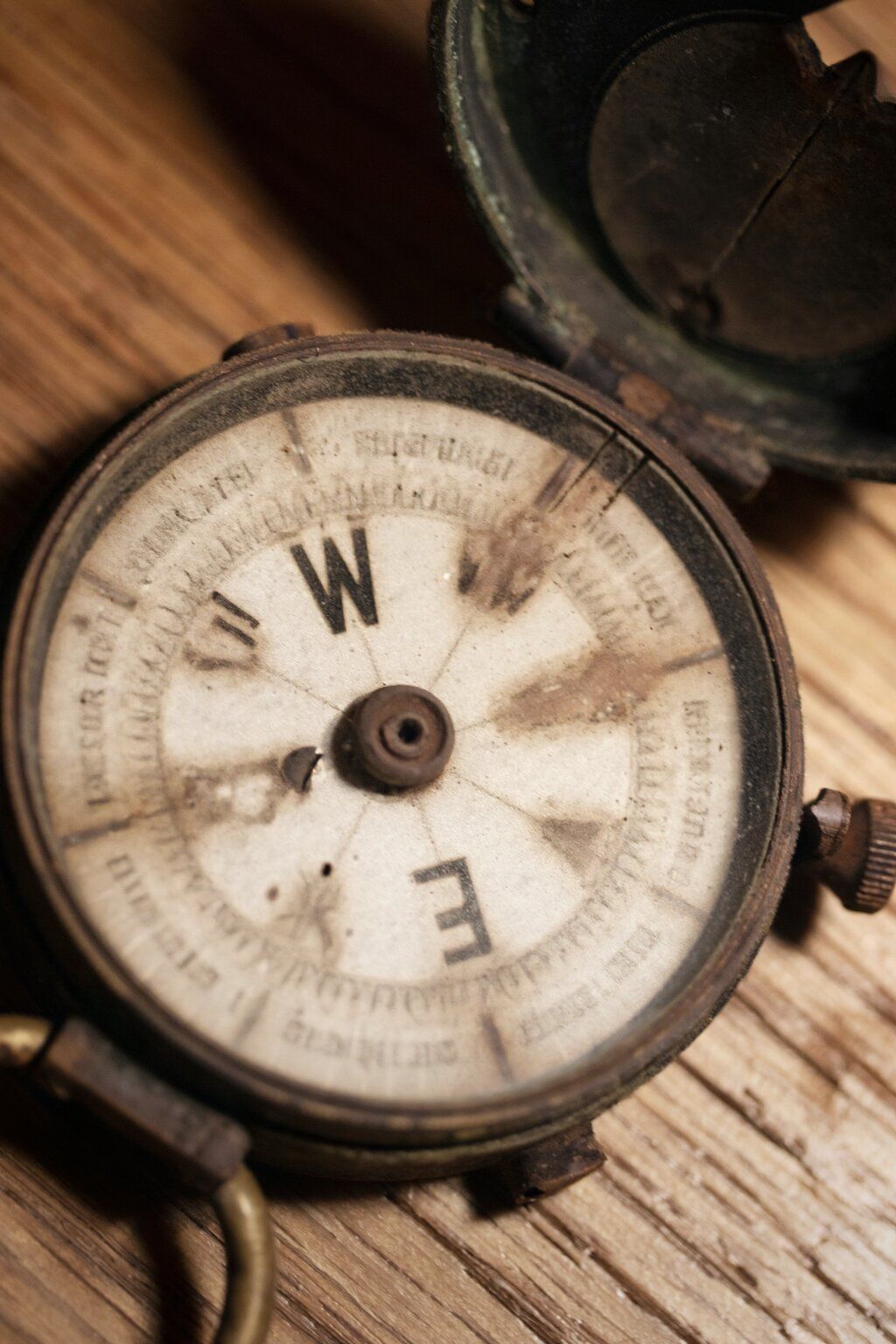} &
    \failimg{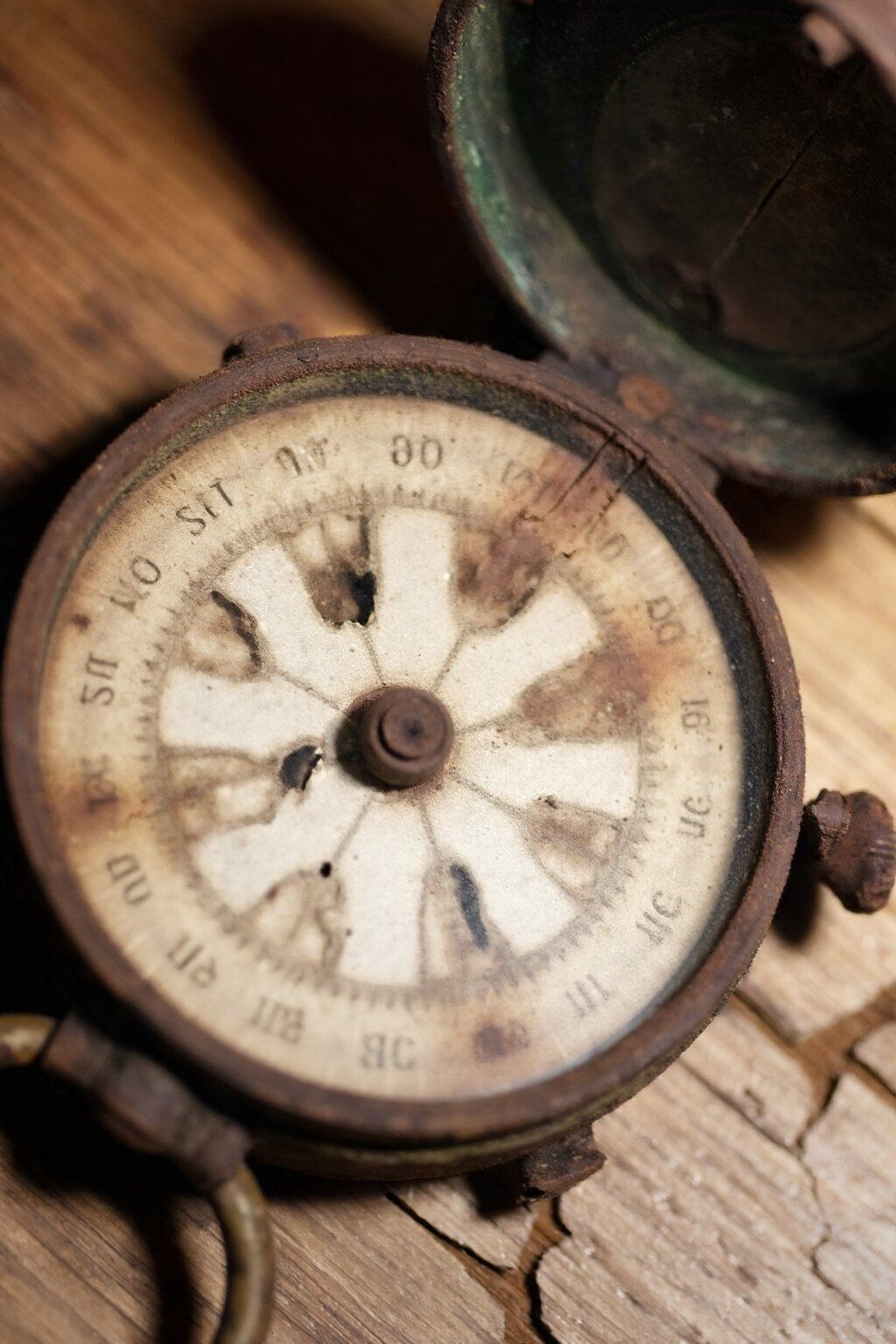} &
    \failimg{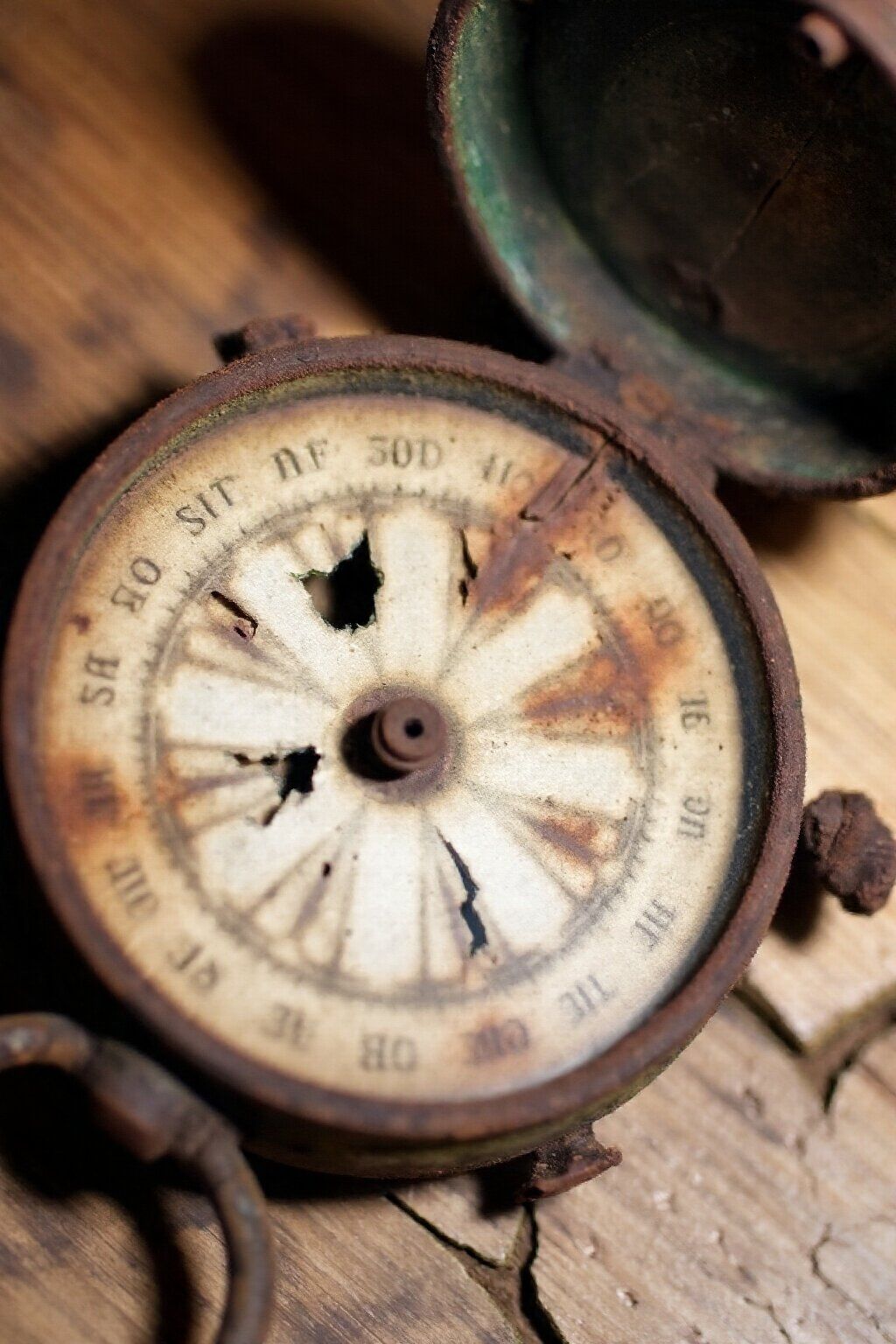} &
    \failimg{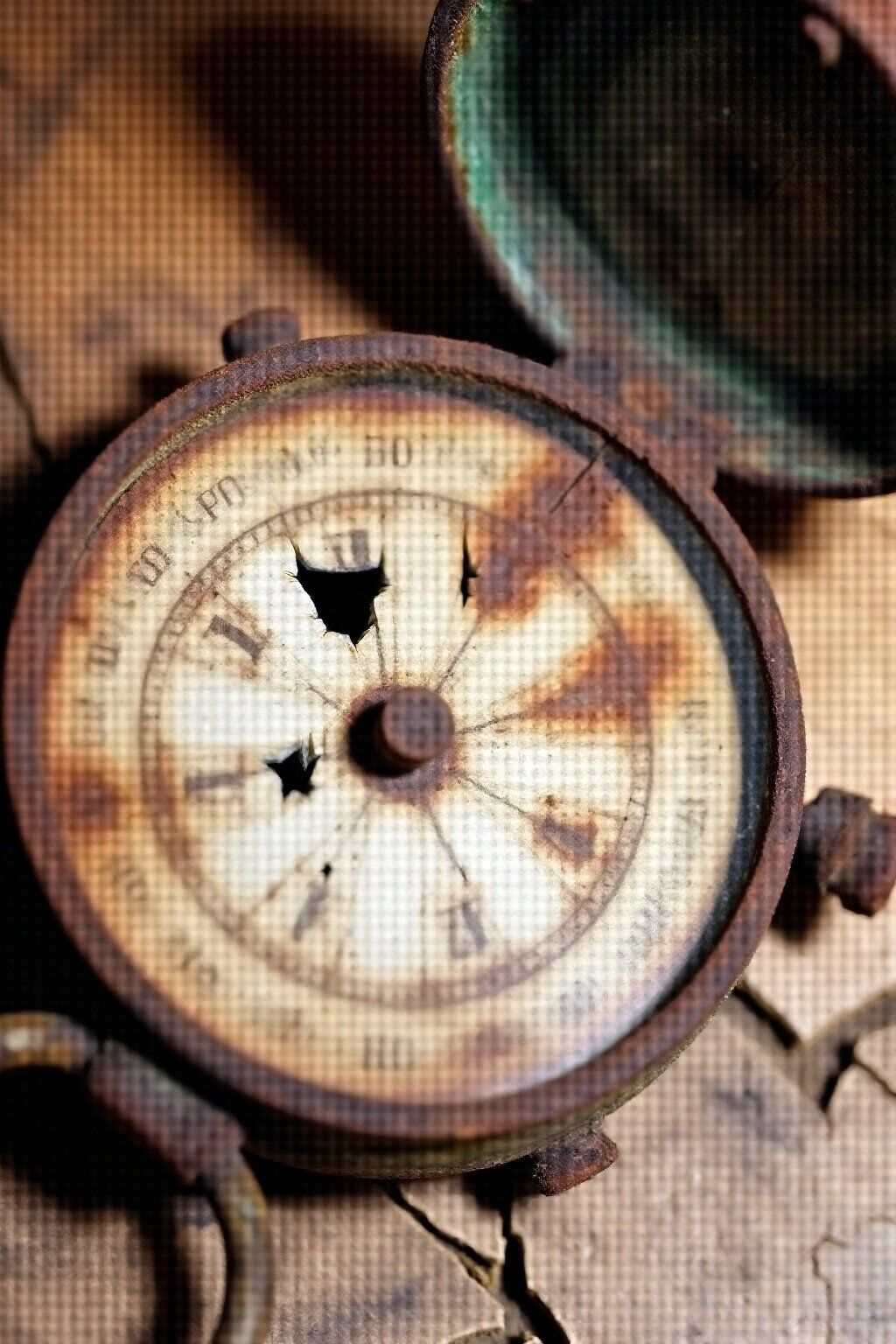} &
    \failimg{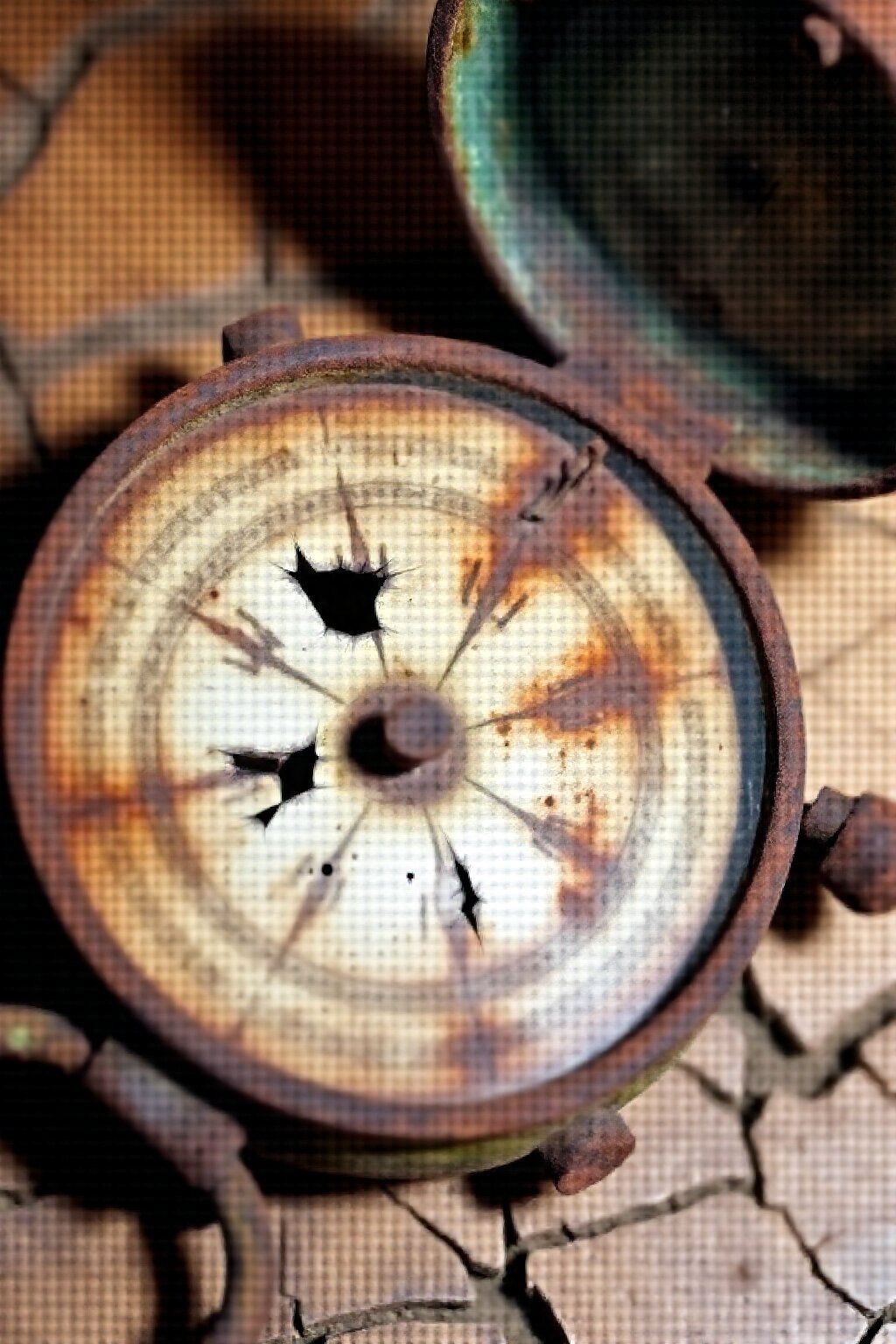} \\
    \multicolumn{6}{c}{\textbf{Src:} ``Shiny brass compass\dots'' ~$\rightarrow$~ \textbf{Tar:} ``Heavily rusted brass compass\dots''} \\
  \end{tabular}
  \vspace{-2mm}
  \caption{\textbf{Failure at extreme strengths.}
  A gradual-editing task (decay) at increasing $s$.
  Results remain plausible up to $s=4$, but structural collapse and over-saturation emerge at larger values.}
  \label{fig:failure_extreme}
\end{figure*}

\begin{figure*}[t]
  \centering
  \setlength{\tabcolsep}{1pt}
  \small
  \begin{tabular}{@{}cccccc@{}}
    Original & $s{=}1$ & $s{=}2$ & $s{=}3$ & $s{=}4$ & $s{=}5$ \\[2pt]
    \failimg{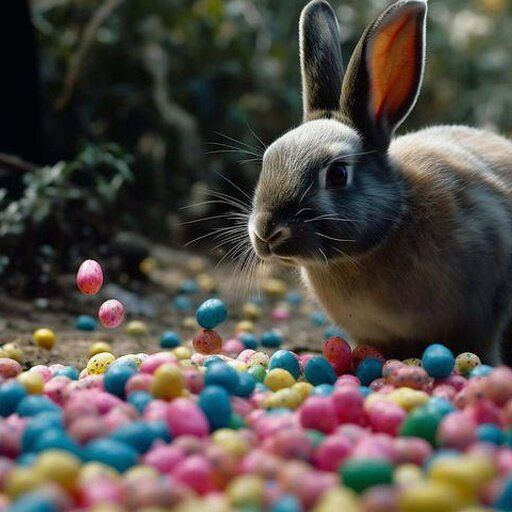} &
    \failimg{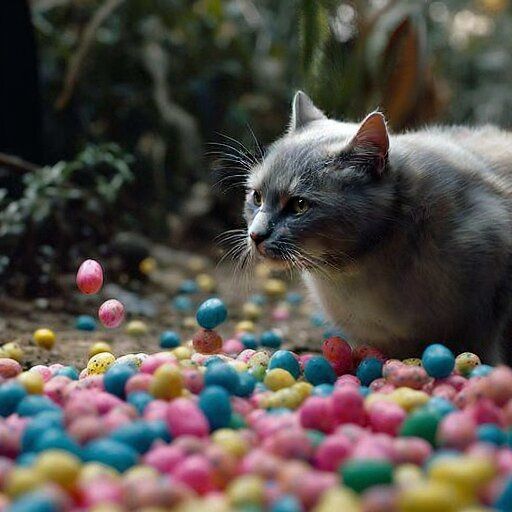} &
    \failimg{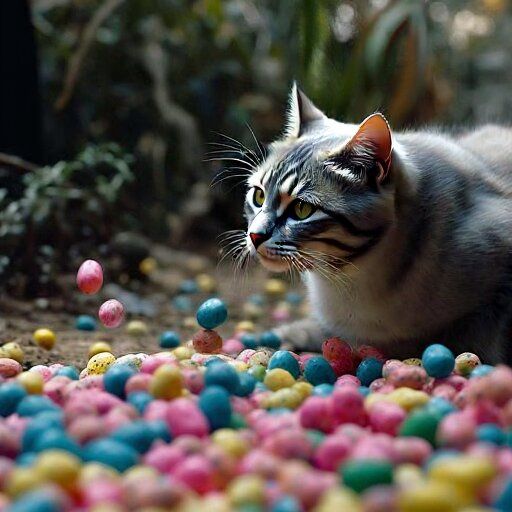} &
    \failimg{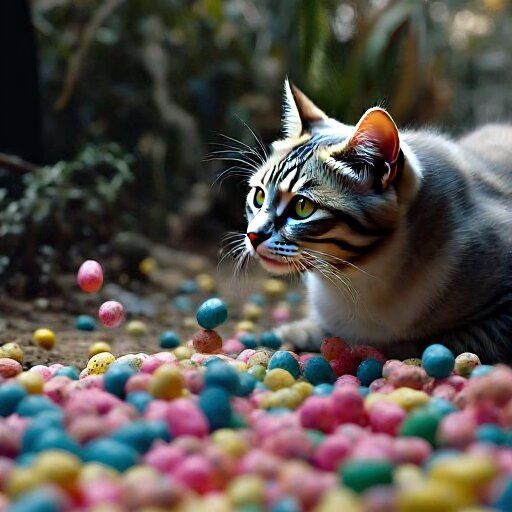} &
    \failimg{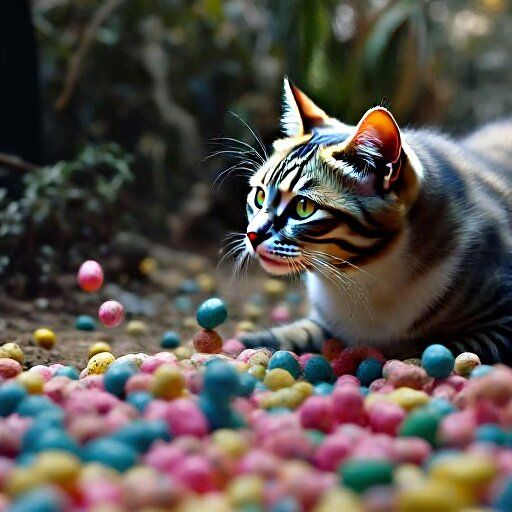} &
    \failimg{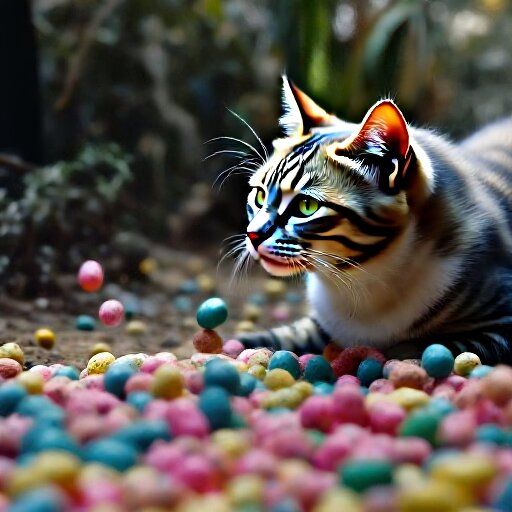} \\
    \multicolumn{6}{c}{\textbf{Src:} ``a rabbit\dots'' ~$\rightarrow$~ \textbf{Tar:} ``a cat\dots''} \\
  \end{tabular}
  \vspace{-2mm}
  \caption{\textbf{Early saturation for discrete concept change.}
  The concept ``rabbit $\to$ cat'' is fully realized at $s{=}1$. Further scaling introduces over-saturation and color artifacts rather than meaningful editing.}
  \label{fig:failure_discrete}
\end{figure*}

\section{Reverse Editing with Negative Strength}
\label{sec:app_reverse}

An additional property of FlowSlider is that setting $s < 0$ reverses the editing direction: the steering term drives semantics opposite to the $c_{\mathrm{src}} \to c_{\mathrm{tar}}$ transition.
This allows, for example, aging to become de-aging or winter-to-summer to become summer-to-winter, all from the same prompt pair and without retraining.

\cref{fig:reverse_editing} demonstrates this across four diverse tasks.
Negative $s$ yields semantically coherent reverse edits: hair color deepens instead of lightening (row~1), colors become more vivid instead of desaturating (row~2), the surface appears even cleaner and newer (row~3), and a smile reverts to a more serious expression (row~4).
This bidirectional control emerges from a single prompt pair, effectively providing a continuous slider from $-s$ to $+s$ without any additional supervision.

This reverse behavior is a byproduct of the semantic structure implicitly learned by the pre-trained generative model: the steering vector defined by the prompt pair approximately aligns with a meaningful semantic axis in latent space, and negating it traverses the opposite direction along this axis.
However, since the reverse direction is not explicitly specified by the prompt pair, the quality of reverse editing is task-dependent.
In practice, we observe that reverse editing tends to be less stable than forward editing, particularly for prompt pairs where the implicit semantic opposite is ambiguous or poorly defined in the model's latent space.

\newcommand{\revimg}[1]{\includegraphics[width=0.185\linewidth]{#1}}

\begin{figure*}[t]
  \centering
  \setlength{\tabcolsep}{1pt}
  \small
  \begin{tabular}{@{}ccccc@{}}
    $s{=}{-3}$ & $s{=}{-1}$ & Original & $s{=}1$ & $s{=}3$ \\[2pt]
    \revimg{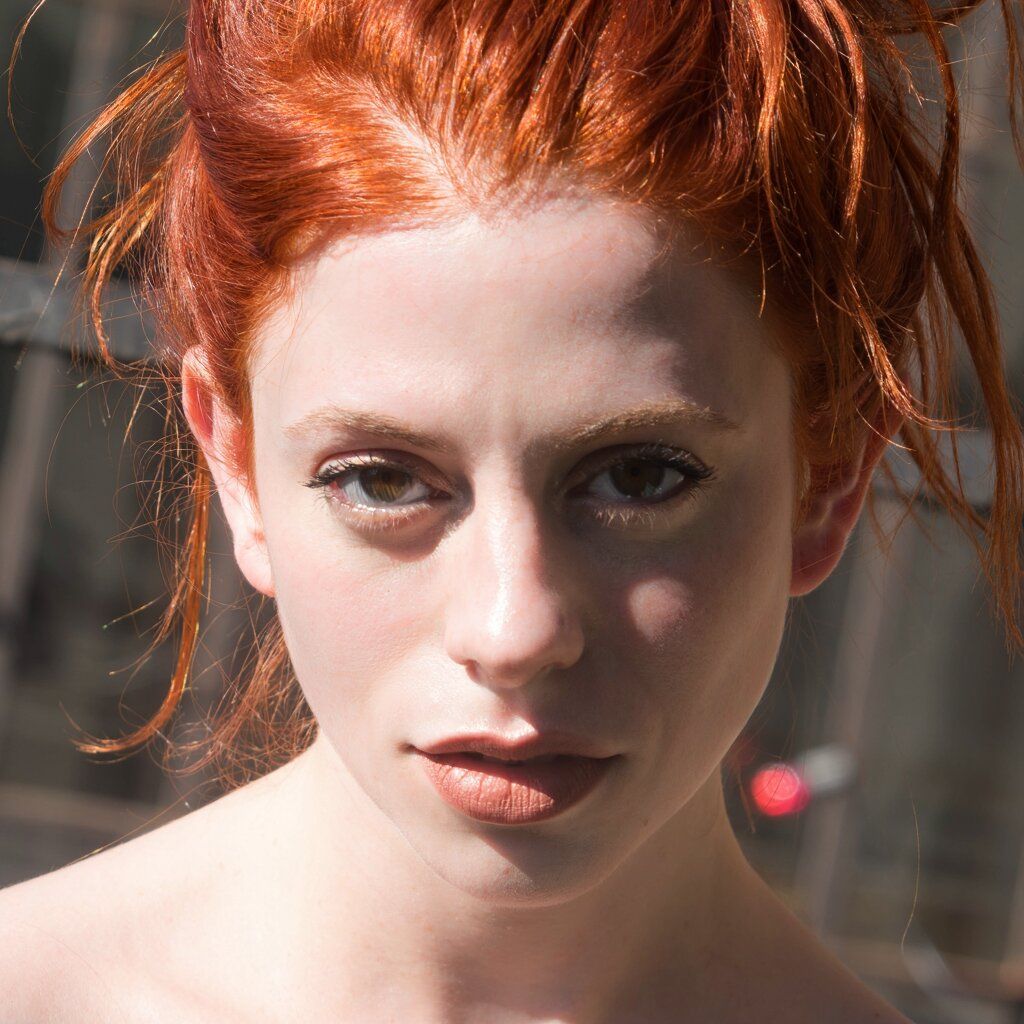} &
    \revimg{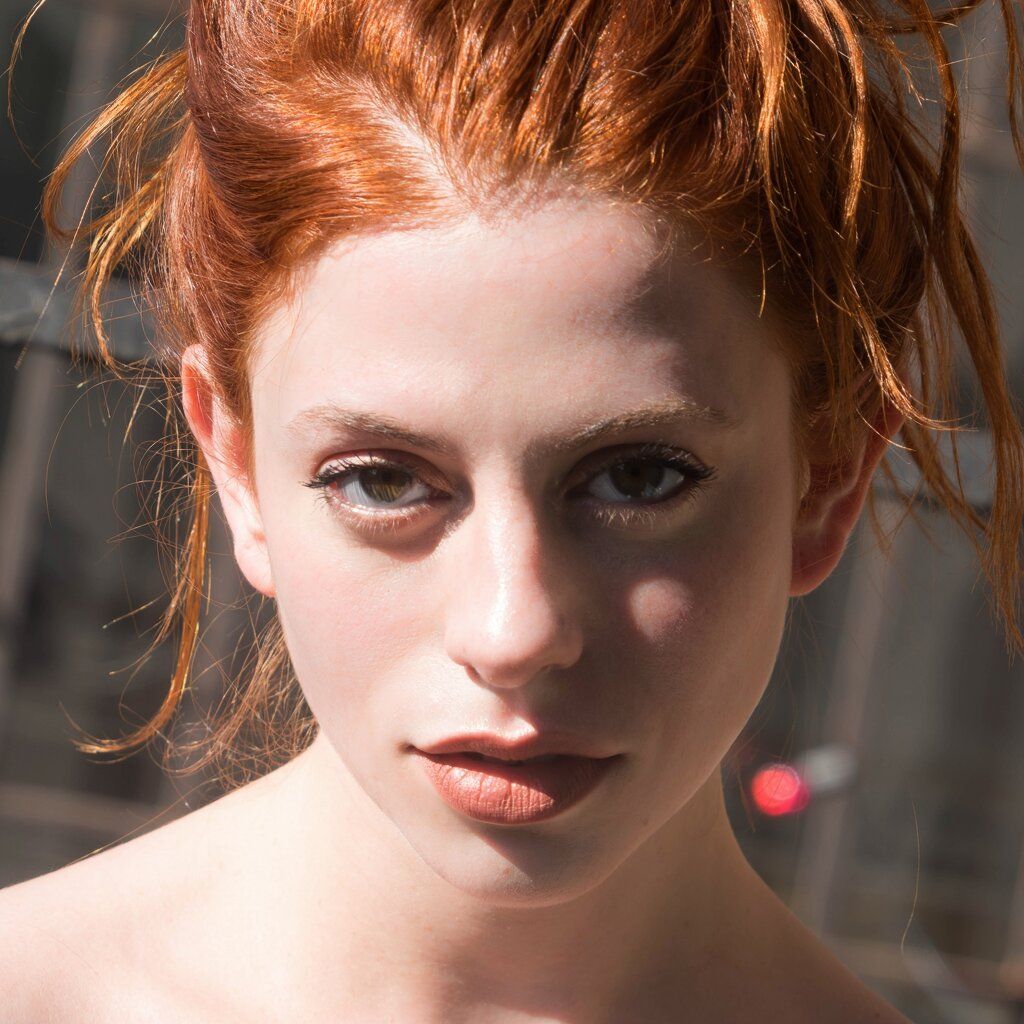} &
    \revimg{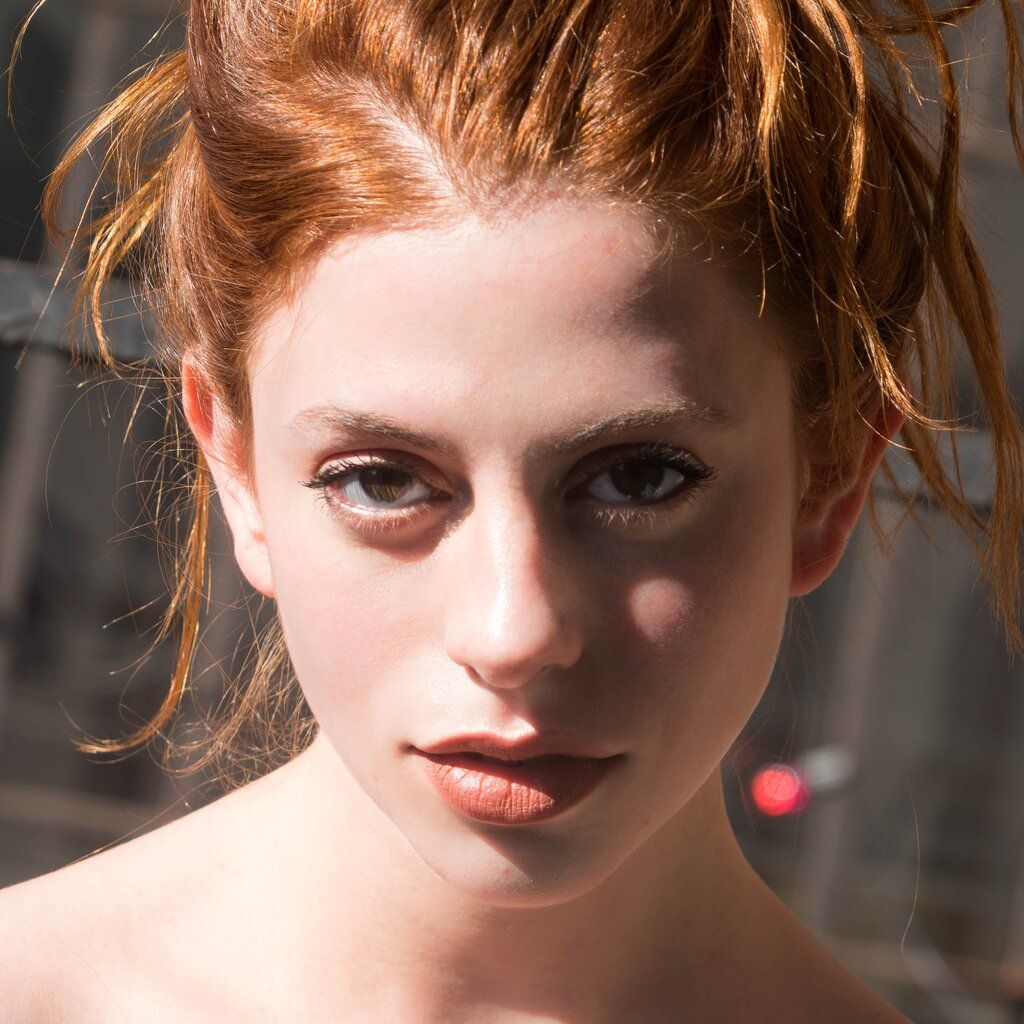} &
    \revimg{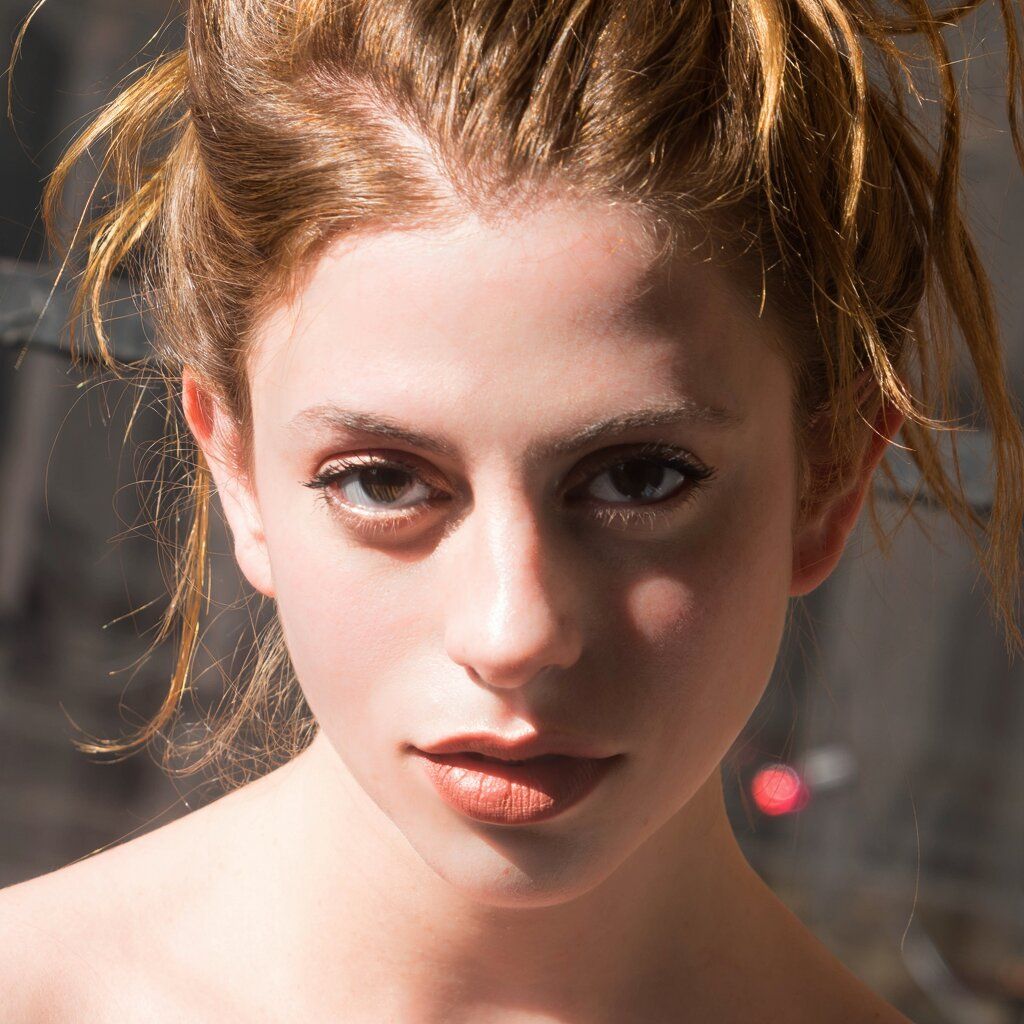} &
    \revimg{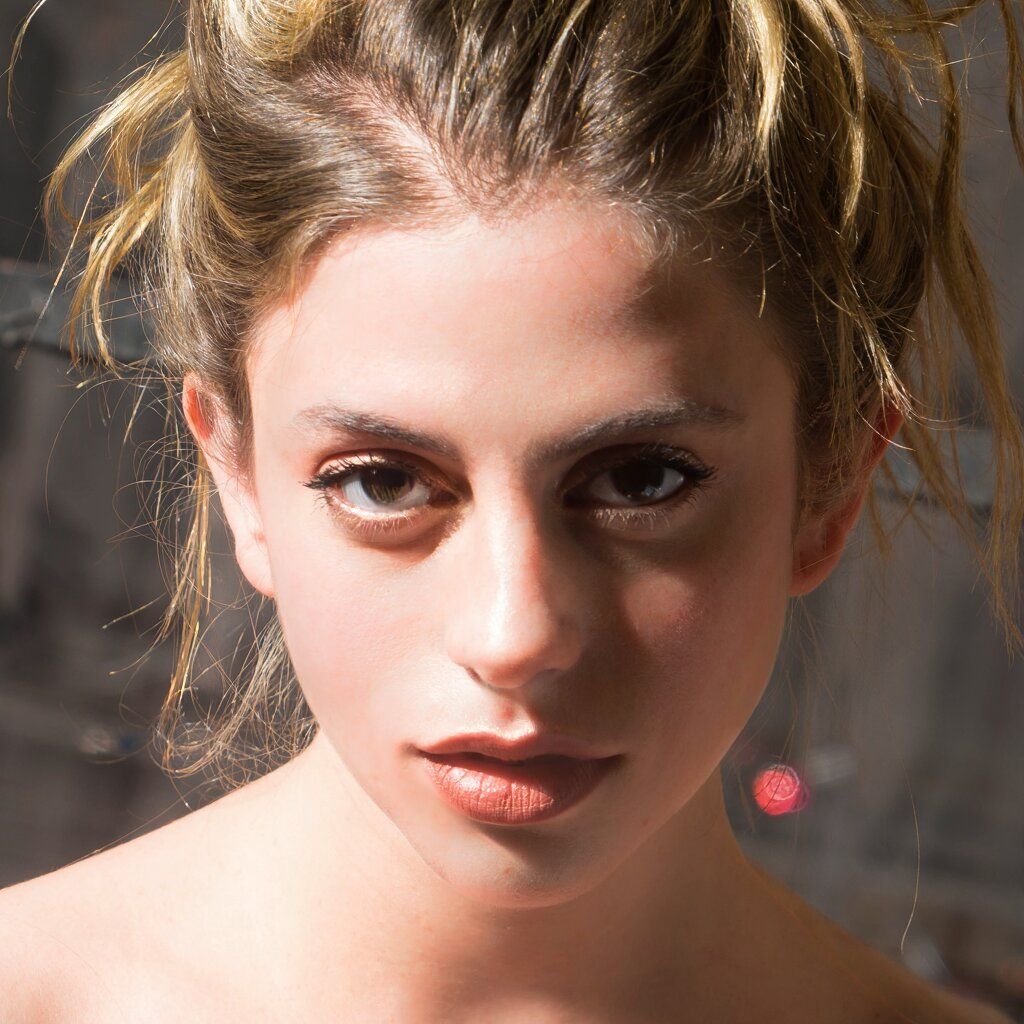} \\
    \multicolumn{5}{c}{Red Hair $\rightarrow$ Blonde Hair (negative: deeper red)} \\[4pt]
    \revimg{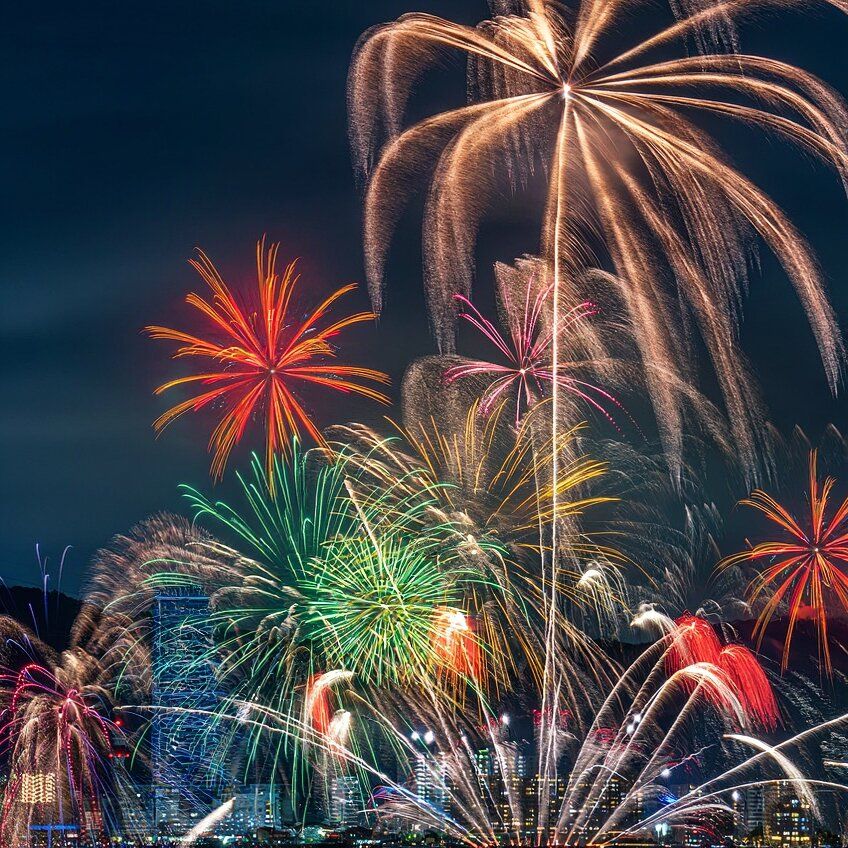} &
    \revimg{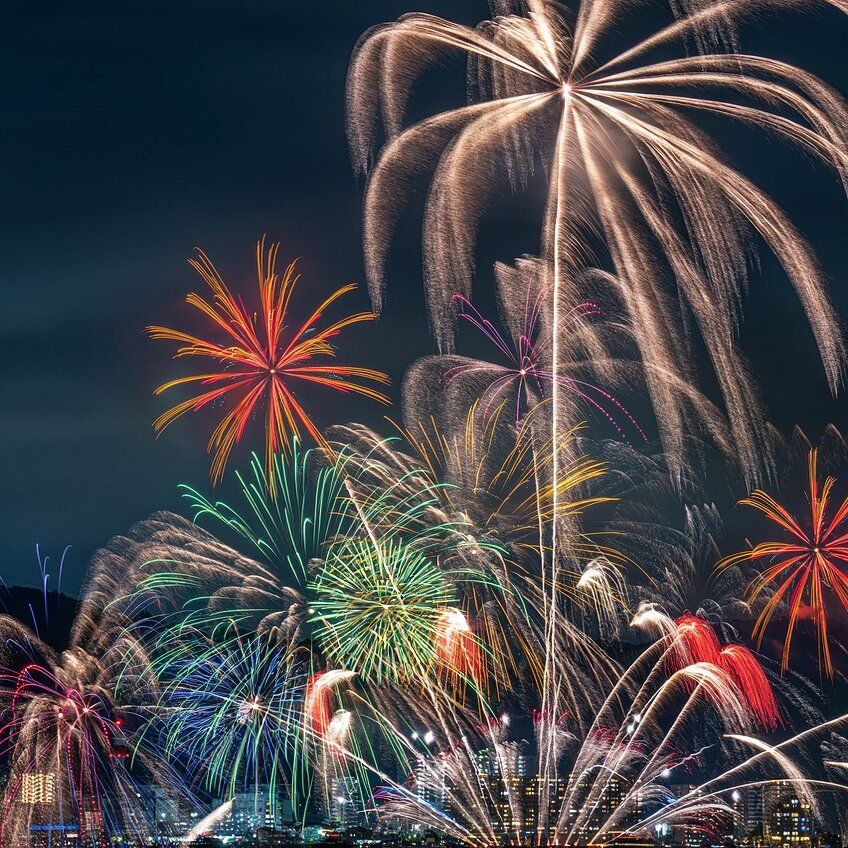} &
    \revimg{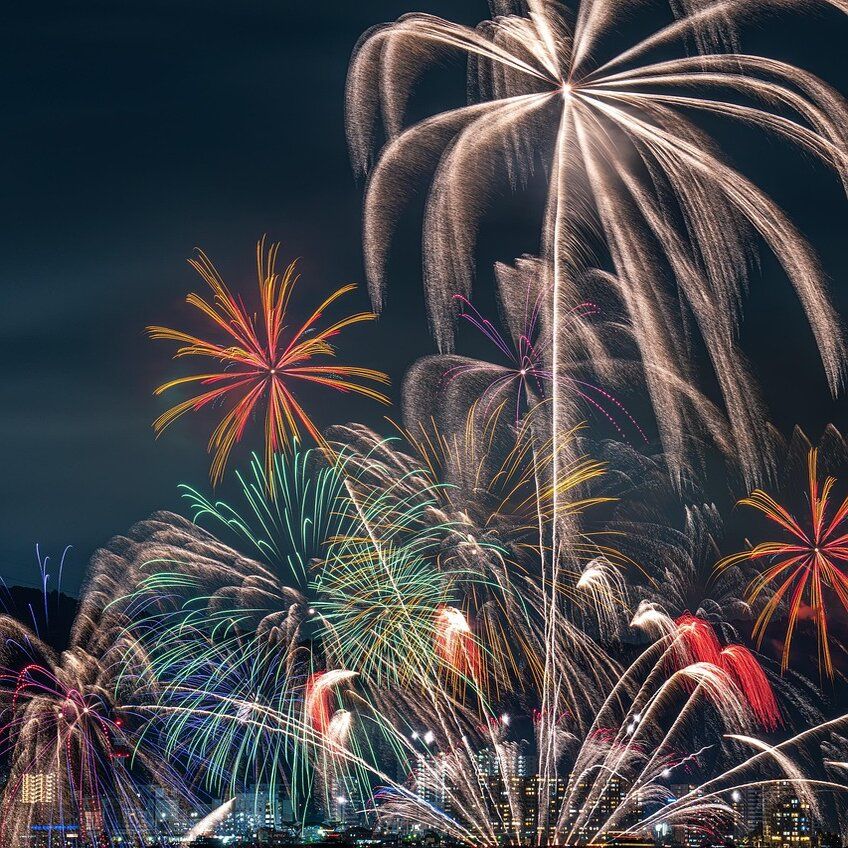} &
    \revimg{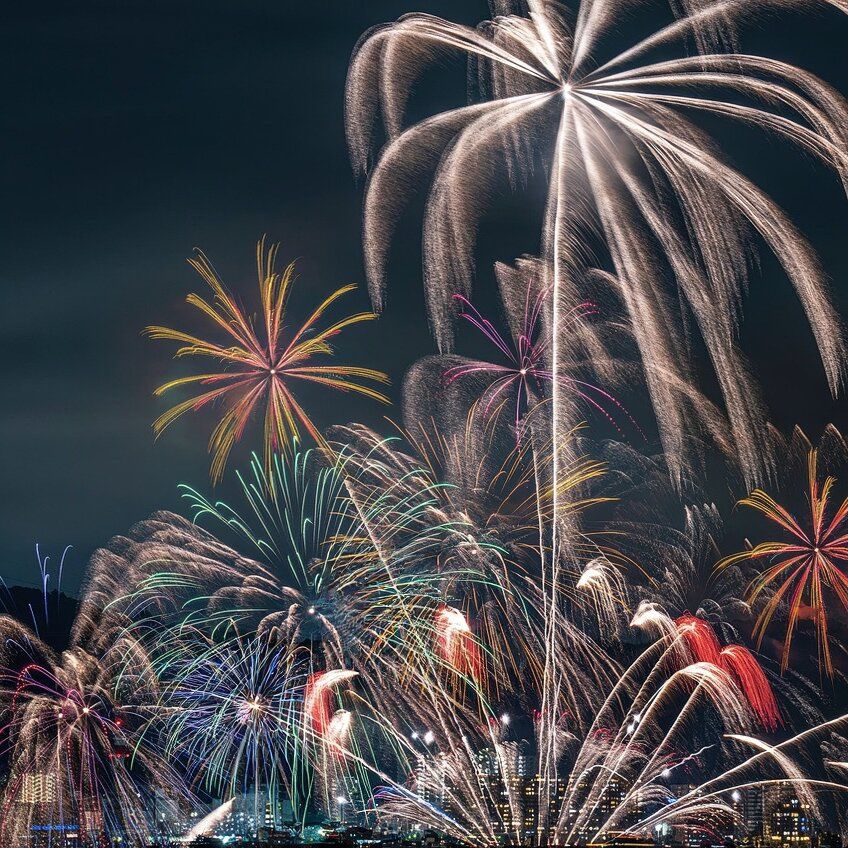} &
    \revimg{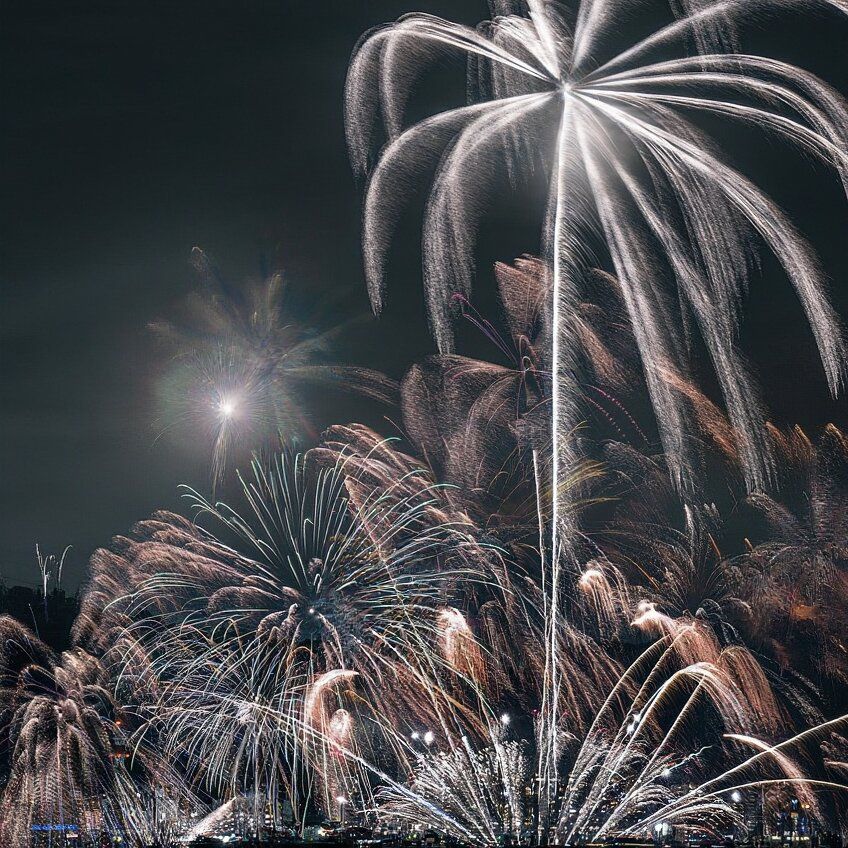} \\
    \multicolumn{5}{c}{Colorful $\rightarrow$ Monochrome (negative: more vivid color)} \\[4pt]
    \revimg{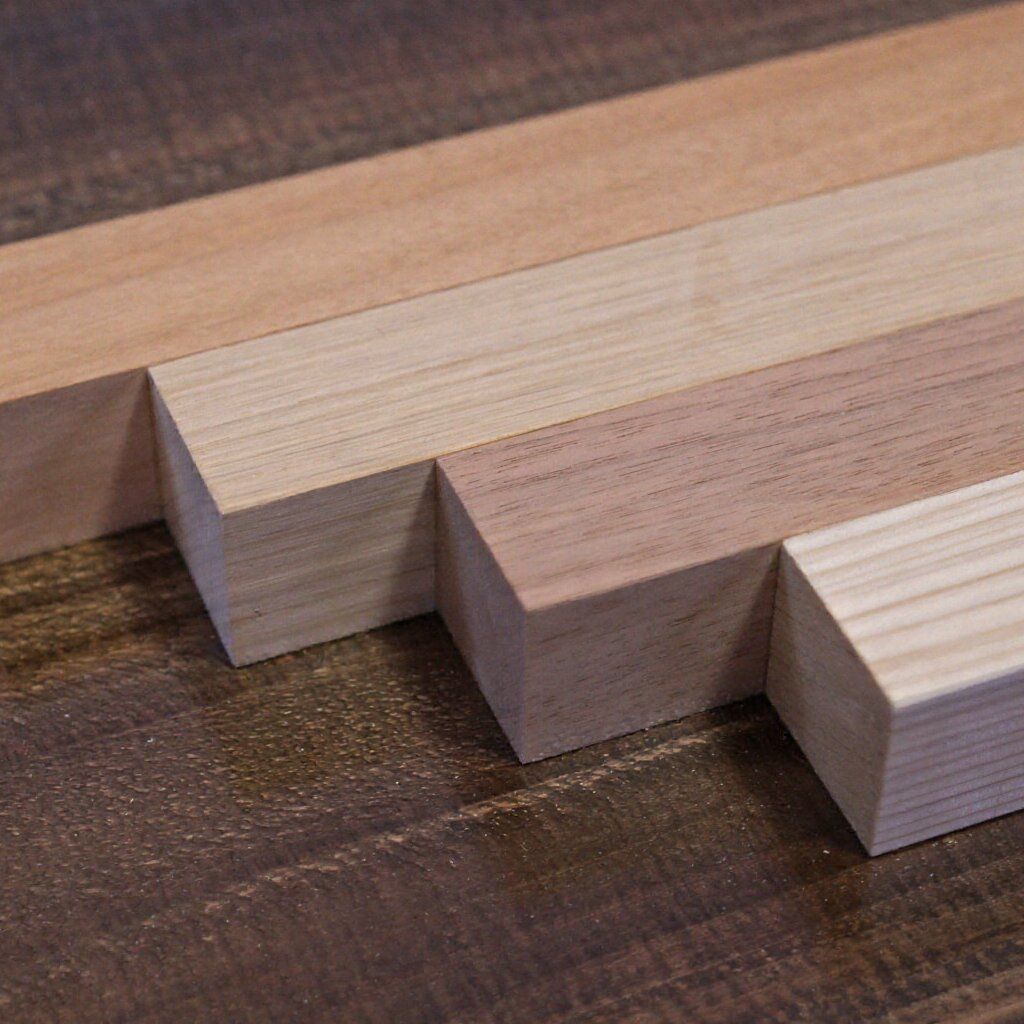} &
    \revimg{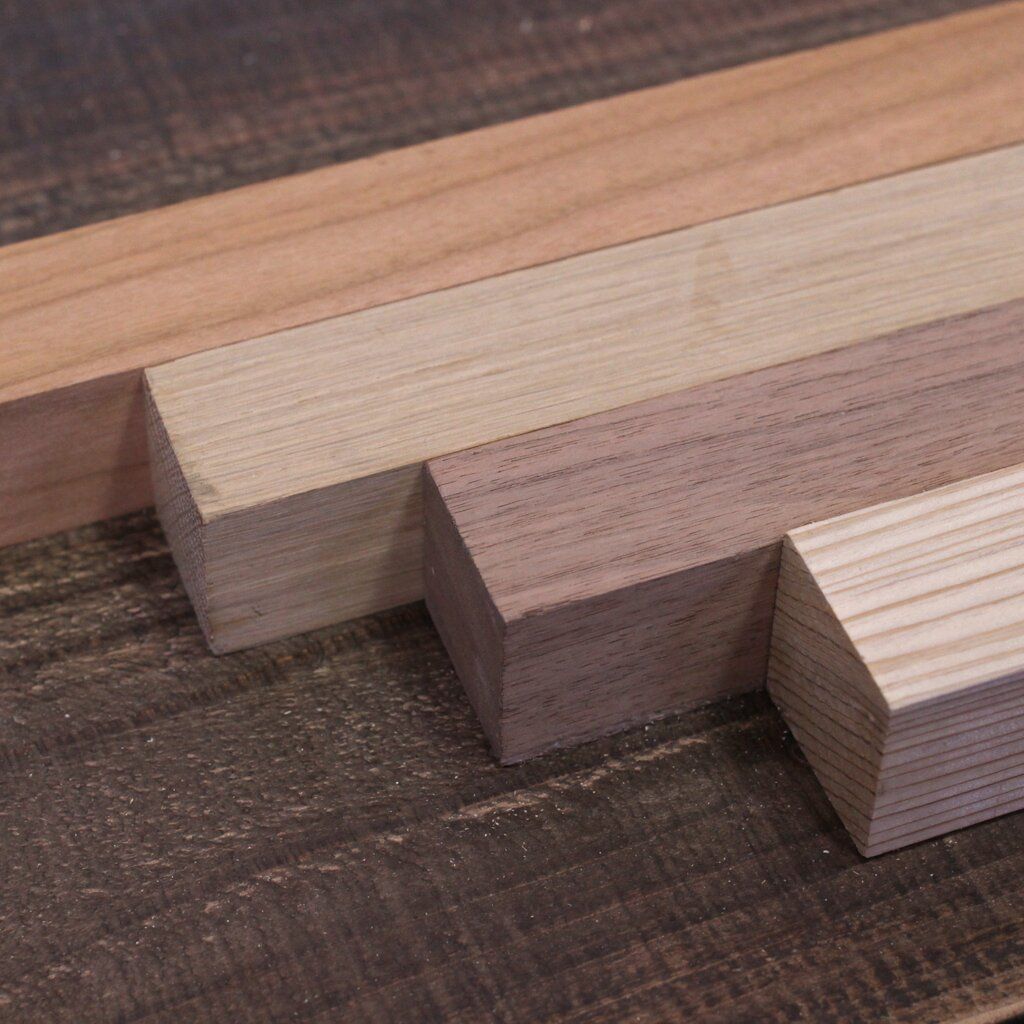} &
    \revimg{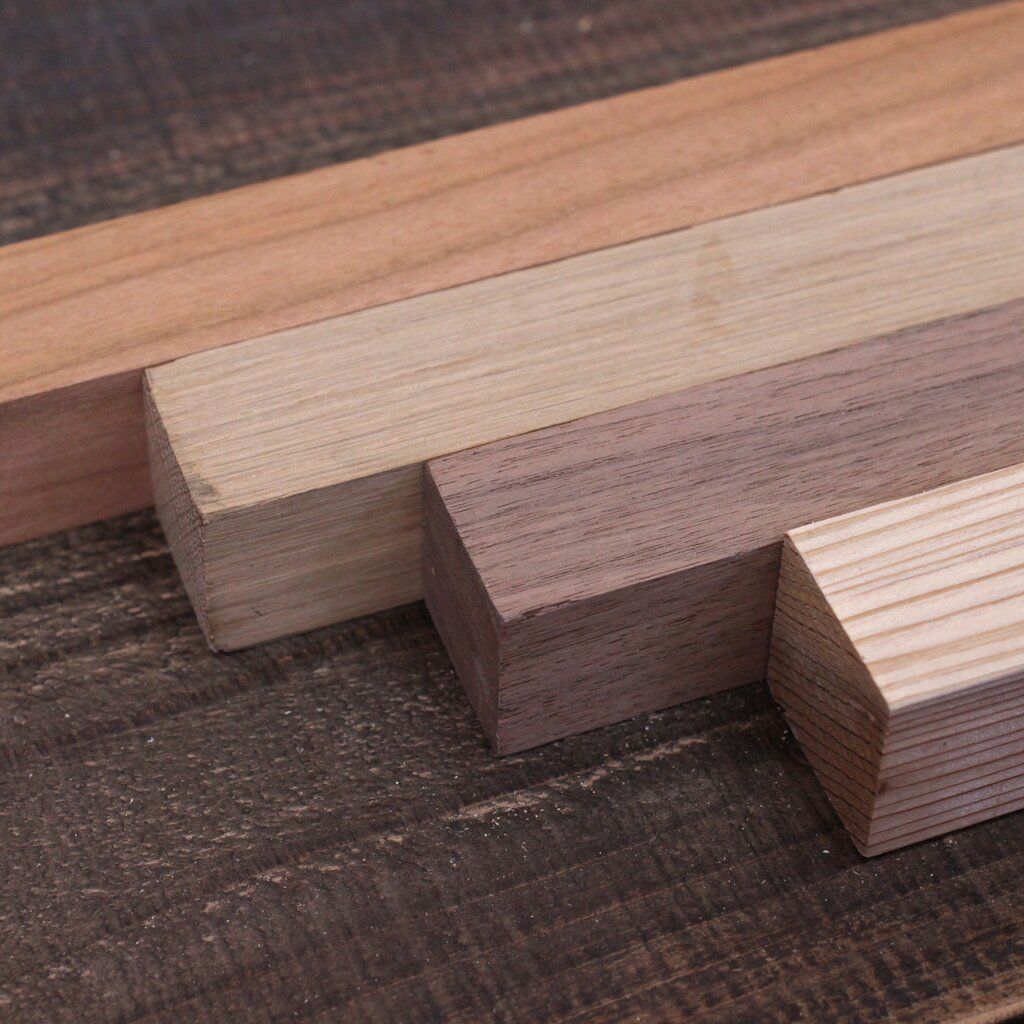} &
    \revimg{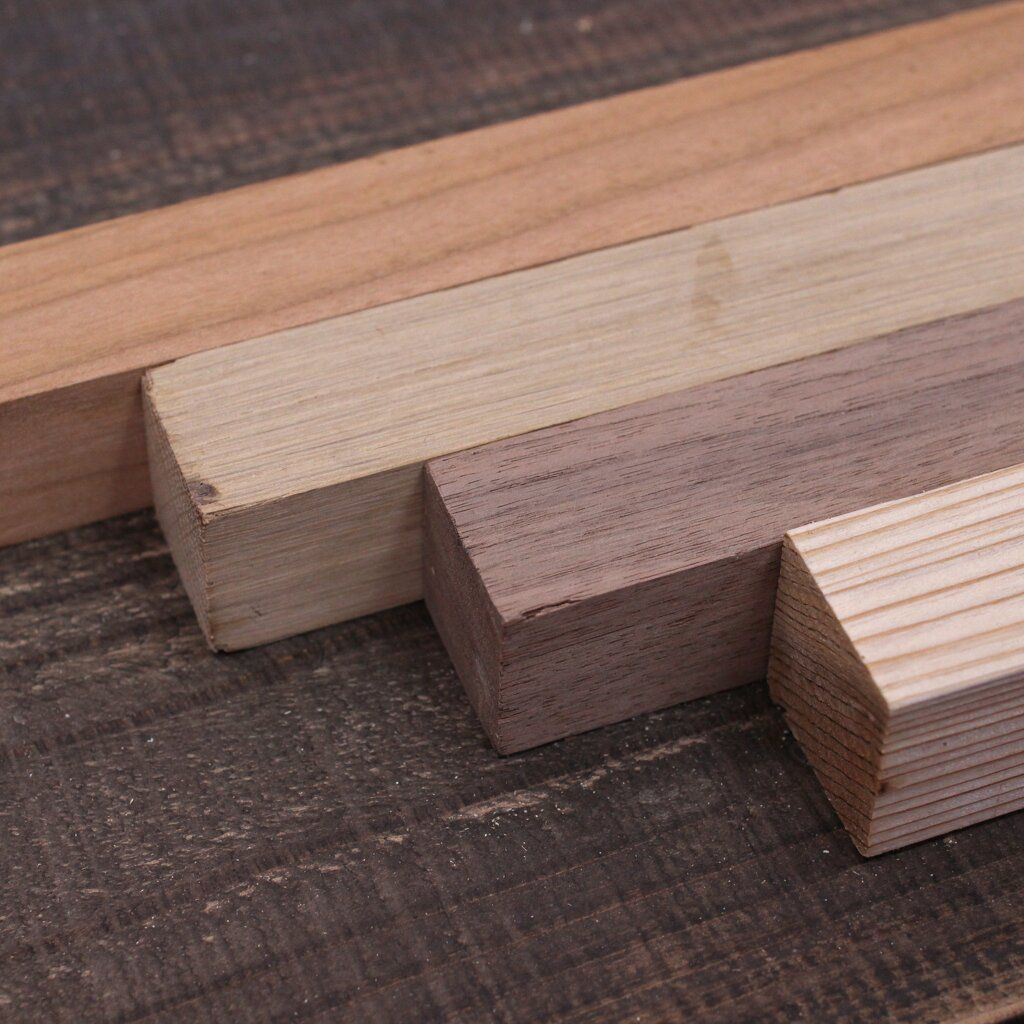} &
    \revimg{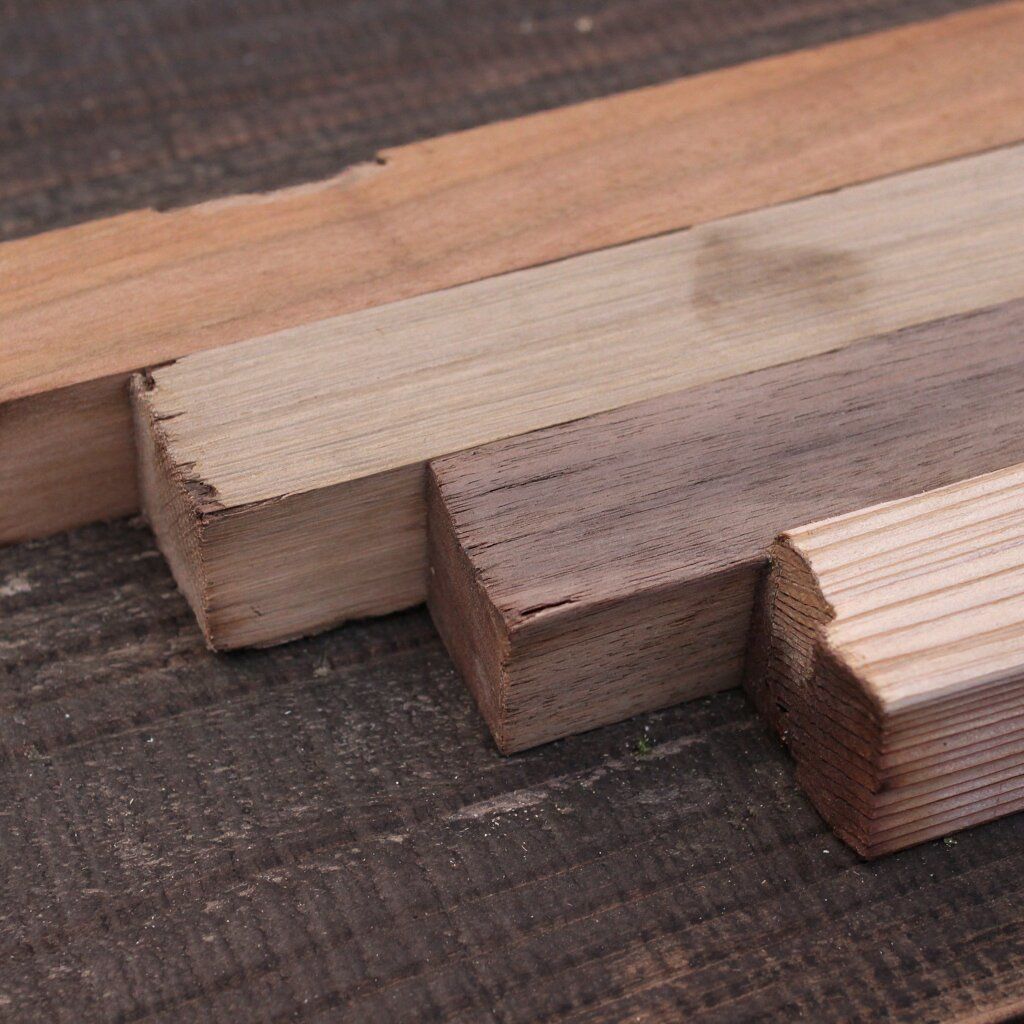} \\
    \multicolumn{5}{c}{Clean $\rightarrow$ Decayed (negative: newer/cleaner appearance)} \\[4pt]
    \revimg{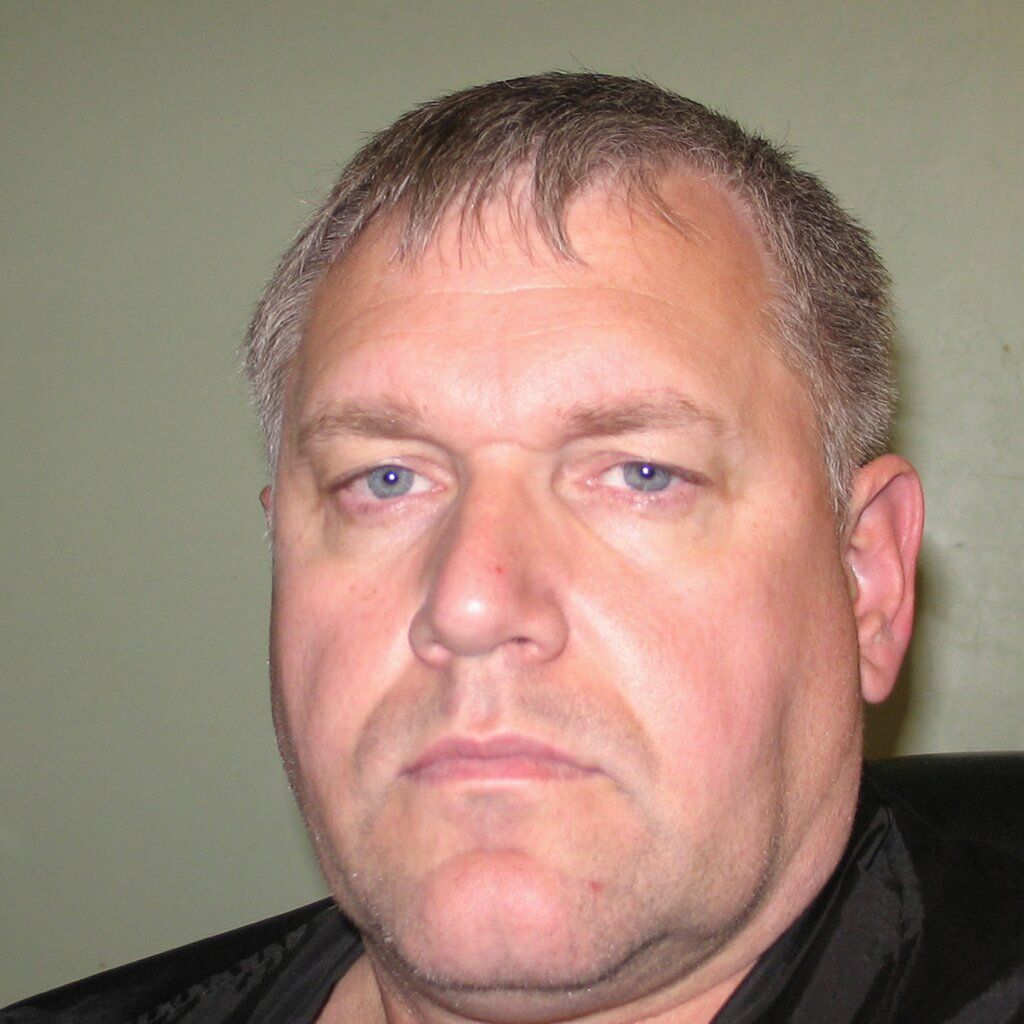} &
    \revimg{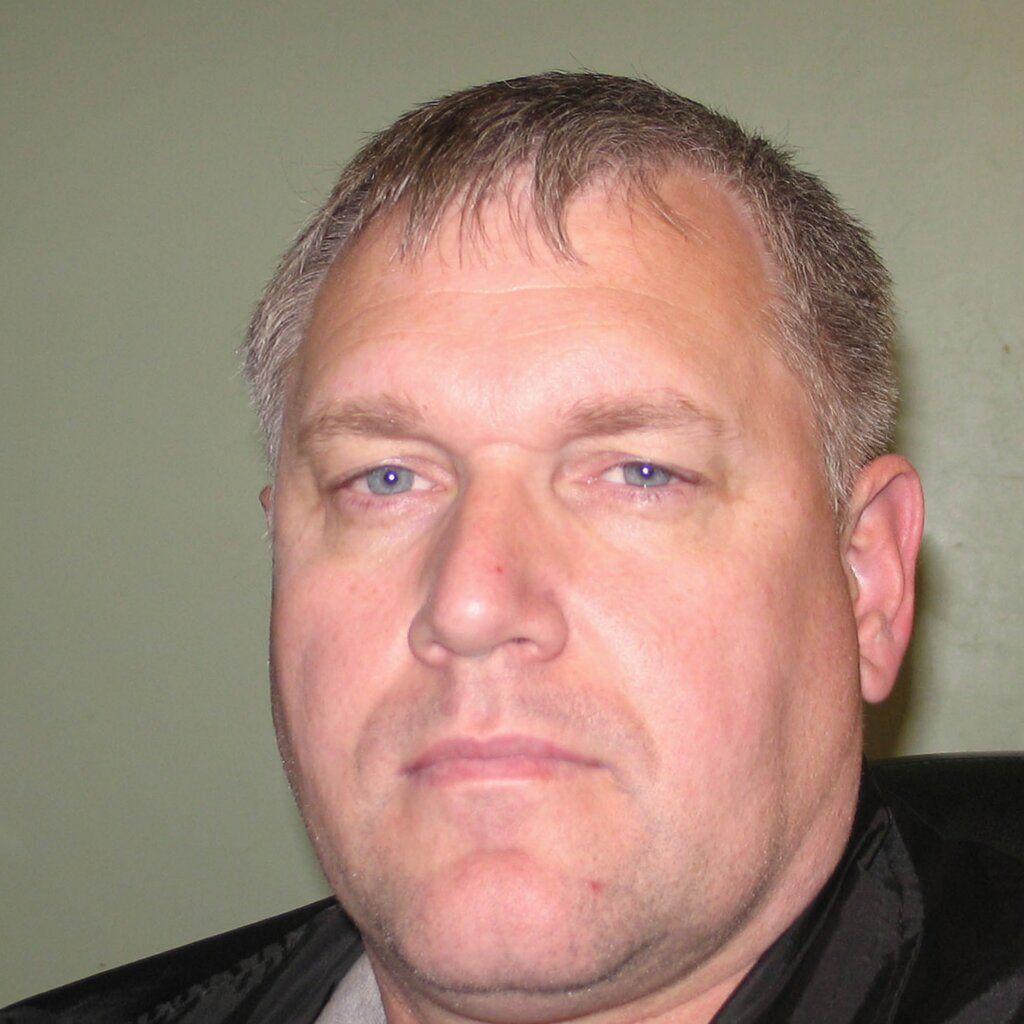} &
    \revimg{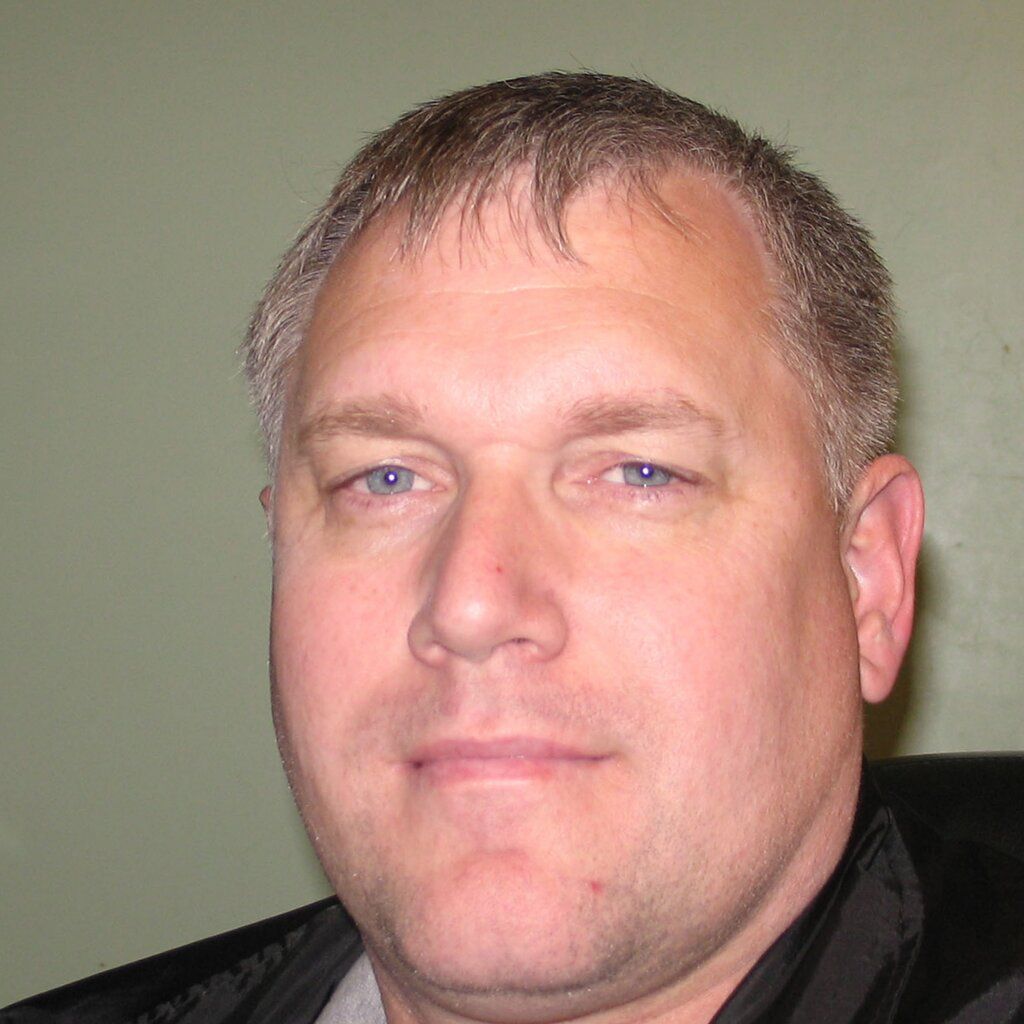} &
    \revimg{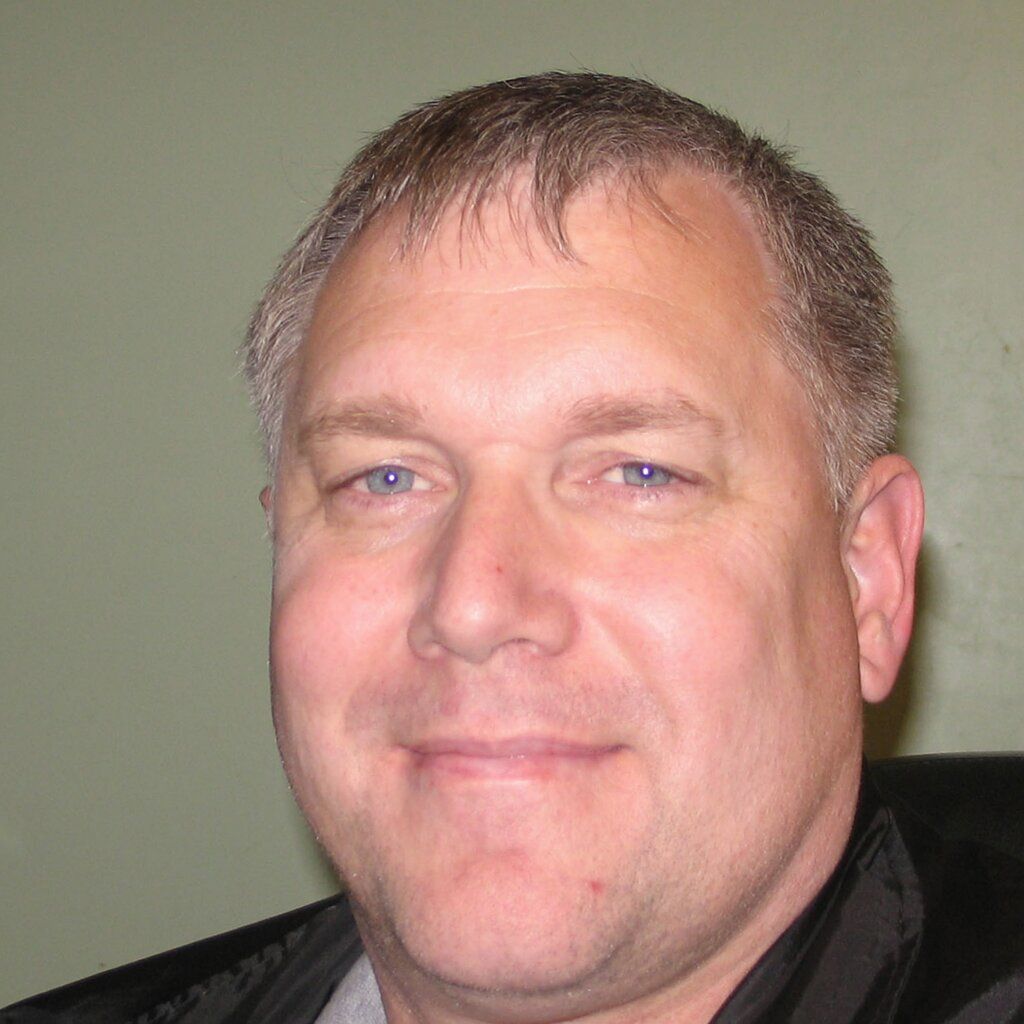} &
    \revimg{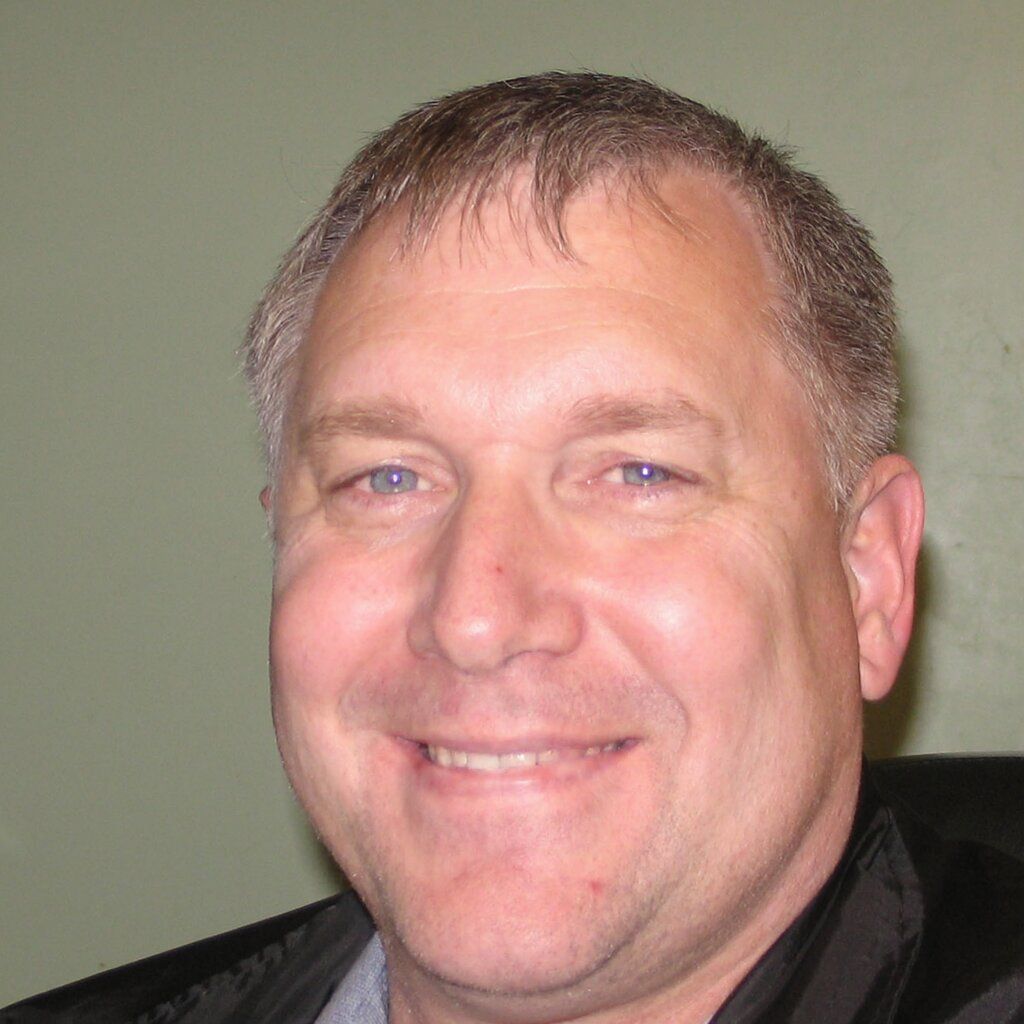} \\
    \multicolumn{5}{c}{Neutral $\rightarrow$ Smiling (negative: more serious)} \\
  \end{tabular}
  \vspace{-2mm}
  \caption{\textbf{Bidirectional editing with negative strength.}
  Setting $s < 0$ reverses the edit direction, enabling bidirectional control from a single prompt pair without retraining.
  Each row shows the same source image edited with negative (left) and positive (right) strengths.}
  \label{fig:reverse_editing}
\end{figure*}

%% file: main.bib
@String(PAMI  = {IEEE Trans. Pattern Anal. Mach. Intell.})

@String(CVPR  = {IEEE Conf. Comput. Vis. Pattern Recog.})

@String(ICCV  = {Int. Conf. Comput. Vis.})

@String(ECCV  = {Eur. Conf. Comput. Vis.})

@String(NeurIPS = {Adv. Neural Inform. Process. Syst.})

@String(ICML  = {Int. Conf. Mach. Learn.})

@String(ICLR  = {Int. Conf. Learn. Represent.})

@String(TOG   = {ACM Trans. Graph.})

@String(PAMI  = {IEEE TPAMI})

@String(CVPR  = {CVPR})

@String(ICCV  = {ICCV})

@String(ECCV  = {ECCV})

@String(NeurIPS = {NeurIPS})

@String(ICML  = {ICML})

@String(ICLR  = {ICLR})

@String(TOG   = {ACM TOG})

@InProceedings{Kulikov_2025_ICCV,
  author = {Kulikov, Vladimir and Kleiner, Matan and Huberman-Spiegelglas, Inbar and Michaeli, Tomer},
  title = {FlowEdit: Inversion-Free Text-Based Editing Using Pre-Trained Flow Models},
  booktitle = ICCV,
  year = {2025},
  pages = {19721-19730}
}

@InProceedings{Brooks_2023_CVPR,
  author = {Brooks, Tim and Holynski, Aleksander and Efros, Alexei A.},
  title = {InstructPix2Pix: Learning To Follow Image Editing Instructions},
  booktitle = CVPR,
  year = {2023},
  pages = {18392--18402}
}

@InProceedings{Kawar_2023_CVPR,
  author = {Kawar, Bahjat and Zada, Shiran and Lang, Oran and Tov, Omer and Chang, Huiwen and Dekel, Tali and Mosseri, Inbar and Irani, Michal},
  title = {Imagic: Text-Based Real Image Editing With Diffusion Models},
  booktitle = CVPR,
  year = {2023},
  pages = {6007--6017}
}

@inproceedings{DBLP:conf/iclr/HertzMTAPC23,
  author       = {Amir Hertz and
                  Ron Mokady and
                  Jay Tenenbaum and
                  Kfir Aberman and
                  Yael Pritch and
                  Daniel Cohen{-}Or},
  title        = {Prompt-to-Prompt Image Editing with Cross-Attention Control},
  booktitle    = ICLR,
  year         = {2023},
}

@inproceedings{shen2020interpreting,
  title     = {Interpreting the Latent Space of GANs for Semantic Face Editing},
  author    = {Shen, Yujun and Gu, Jinjin and Tang, Xiaoou and Zhou, Bolei},
  booktitle = CVPR,
  year      = {2020},
  pages = {9240--9249}
}

@inproceedings{harkonen2020ganspace,
  title     = {GANSpace: Discovering Interpretable GAN Controls},
  author    = {H{\"a}rk{\"o}nen, Erik and Hertzmann, Aaron and Lehtinen, Jaakko and Paris, Sylvain},
  booktitle = NeurIPS,
  year      = {2020},
  pages = {9841--9850},
}

@article{abdal2021styleflow,
  author = {Abdal, Rameen and Zhu, Peihao and Mitra, Niloy J. and Wonka, Peter},
  title = {StyleFlow: Attribute-Conditioned Exploration of StyleGAN-Generated Images Using Conditional Continuous Normalizing Flows},
  journal = TOG,
  volume={40},
  number={3},
  pages={1--21},
  year={2021},
}

@inproceedings{DBLP:conf/eccv/GandikotaMZTB24,
  author       = {Rohit Gandikota and
                  Joanna Materzynska and
                  Tingrui Zhou and
                  Antonio Torralba and
                  David Bau},
  title        = {Concept Sliders: LoRA Adaptors for Precise Control in Diffusion Models},
  booktitle    = ECCV,
  year         = {2024},
}

@article{liu2022rectifiedflow,
  title   = {Flow Straight and Fast: Learning to Generate and Transfer Data with Rectified Flow},
  author  = {Xingchao Liu and Chengyue Gong and Qiang Liu},
  journal = ICLR,
  year    = {2023},
}

@inproceedings{meng2022sdedit,
  title     = {{SDE}dit: Guided Image Synthesis and Editing with Stochastic Differential Equations},
  author    = {Chenlin Meng and Yutong He and Yang Song and Jiaming Song and Jiajun Wu and Jun-Yan Zhu and Stefano Ermon},
  booktitle = ICLR,
  year      = {2022}
}

@inproceedings{mokady2022nulltextinversion,
  title   = {Null-text Inversion for Editing Real Images using Guided Diffusion Models},
  author  = {Mokady, Ron and Hertz, Amir and Aberman, Kfir and Pritch, Yael and Cohen-Or, Daniel},
  booktitle = CVPR,
  year = {2023},
  pages = {6038--6047}
}

@inproceedings{radford2021clip,
  title   = {Learning Transferable Visual Models From Natural Language Supervision},
  author  = {Radford, Alec and Kim, Jong Wook and Hallacy, Chris and Ramesh, Aditya and Goh, Gabriel and Agarwal, Sandhini and Sastry, Girish and Askell, Amanda and Mishkin, Pamela and Clark, Jack and Krueger, Gretchen and Sutskever, Ilya},
  year    = {2021},
  booktitle = ICML,
}

@inproceedings{zhang2018lpips,
  title     = {The Unreasonable Effectiveness of Deep Features as a Perceptual Metric},
  author    = {Zhang, Richard and Isola, Phillip and Efros, Alexei A. and Shechtman, Eli and Wang, Oliver},
  booktitle = CVPR,
  year      = {2018},
  pages = {586--595}
}

@inproceedings{fu2023dreamsim,
  author    = {Fu, Stephanie and Tamir, Netanel and Sundaram, Shobhita and Chai, Lucy and Zhang, Richard and Dekel, Tali and Isola, Phillip},
  title     = {DreamSim: Learning New Dimensions of Human Visual Similarity using Synthetic Data},
  booktitle = NeurIPS,
  year      = {2023}
}

@inproceedings{ju2023direct,
  title   = {PnP Inversion: Boosting Diffusion-based Editing with 3 Lines of Code},
  author  = {Ju, Xuan and Zeng, Ailing and Bian, Yuxuan and Liu, Shaoteng and Xu, Qiang},
  year    = {2024},
  booktitle = ICLR,
}

@inproceedings{esser2024scalingrectifiedflowtransformers,
  title   = {Scaling Rectified Flow Transformers for High-Resolution Image Synthesis},
  author  = {Esser, Patrick and Kulal, Sumith and Blattmann, Andreas and Entezari, Rahim and M{\"u}ller, Jonas and Saini, Harry and Levi, Yam and Lorenz, Dominik and Sauer, Axel and Boesel, Frederic and Podell, Dustin and Dockhorn, Tim and English, Zion and Lacey, Kyle and Goodwin, Alex and Marek, Yannik and Rombach, Robin},
  year    = {2024},
  booktitle = ICML,
  pages = {12606--12633},
}

@misc{bfl2024flux1dev,
  title        = {{FLUX}.1 [dev] Model Card},
  author       = {{Black Forest Labs}},
  year         = {2024},
  howpublished = {Hugging Face},
  url          = {https://huggingface.co/black-forest-labs/FLUX.1-dev},
  note         = {Model card. Accessed: 2026-02}
}

@inproceedings{Goodfellow2014GAN,
  title     = {Generative Adversarial Nets},
  author    = {Goodfellow, Ian and Pouget-Abadie, Jean and Mirza, Mehdi and Xu, Bing and Warde-Farley, David and Ozair, Sherjil and Courville, Aaron and Bengio, Yoshua},
  booktitle = {NIPS},
  year      = {2014}
}

@inproceedings{Ho2022CFG,
  title     = {Classifier-Free Diffusion Guidance},
  author    = {Ho, Jonathan and Salimans, Tim},
  booktitle = NeurIPS,
  year      = {2021}
}

@inproceedings{Saharia2022Imagen,
  title   = {Photorealistic Text-to-Image Diffusion Models with Deep Language Understanding},
  author  = {Saharia, Chitwan and Chan, William and Saxena, Saurabh and Li, Lala and Whang, Jay and Denton, Emily and Ghasemipour, Seyed Kamyar and Ayan, Burcu Karagol and Mahdavi, S Sara and Lopes, Rapha{\"e}l Gontijo and Salimans, Tim and Ho, Jonathan and Fleet, David J. and Norouzi, Mohammad},
  year    = {2022},
  booktitle = NeurIPS,
  
}

@inproceedings{Rombach2022LDM,
  title     = {High-Resolution Image Synthesis with Latent Diffusion Models},
  author    = {Rombach, Robin and Blattmann, Andreas and Lorenz, Dominik and Esser, Patrick and Ommer, Bj{\"o}rn},
  booktitle = CVPR,
  year      = {2022},
}

@misc{Ramesh2022DALLE2,
  title   = {Hierarchical Text-Conditional Image Generation with {CLIP} Latents},
  author  = {Ramesh, Aditya and Dhariwal, Prafulla and Nichol, Alex and Chu, Casey and Chen, Mark},
  year    = {2022},
  howpublished = {arXiv preprint},
  eprint  = {2204.06125},
  archivePrefix = {arXiv},
  primaryClass  = {cs.CV}
}

@misc{ZImage2025,
  title   = {Z-Image: An Efficient Image Generation Foundation Model with Single-Stream Diffusion Transformer},
  author  = {{Z-Image Team}},
  year    = {2025},
  howpublished = {arXiv preprint},
  eprint  = {2511.22699},
  archivePrefix = {arXiv},
  primaryClass  = {cs.CV}
}

@inproceedings{Lipman2023FlowMatching,
  title     = {Flow Matching for Generative Modeling},
  author    = {Lipman, Yaron and Chen, Ricky T. Q. and Ben-Hamu, Heli and Nickel, Maximilian and Le, Matthew},
  booktitle = ICLR,
  year      = {2023},
}

@inproceedings{Brown2020GPT3,
  title     = {Language Models are Few-Shot Learners},
  author    = {Brown, Tom B. and Mann, Benjamin and Ryder, Nick and Subbiah, Melanie and Kaplan, Jared and Dhariwal, Prafulla and Neelakantan, Arvind and Shyam, Pranav and Sastry, Girish and Askell, Amanda and Agarwal, Sandhini and Herbert-Voss, Ariel and Krueger, Gretchen and Henighan, Tom and Child, Rewon and Ramesh, Aditya and Ziegler, Daniel M. and Wu, Jeffrey and Winter, Clemens and Hesse, Chris and Chen, Mark and Sigler, Eric and Litwin, Mateusz and Gray, Scott and Chess, Benjamin and Clark, Jack and Berner, Christopher and McCandlish, Sam and Radford, Alec and Sutskever, Ilya and Amodei, Dario},
  booktitle = NeurIPS,
  year      = {2020}
}

@inproceedings{Peebles2023DiT,
  title     = {Scalable Diffusion Models with Transformers},
  author    = {Peebles, William and Xie, Saining},
  booktitle = ICCV,
  year      = {2023}
}

@misc{Bai2023QwenVL,
  title   = {Qwen-VL: A Versatile Vision-Language Model for Understanding, Localization, Text Reading, and Beyond},
  author  = {Bai, Jinze and Bai, Shuai and Yang, Shusheng and Wang, Shuailei and Tan, Shijie and Wang, Pei and Lin, Junyang and Zhou, Chang and Zhou, Jingren},
  howpublished = {arXiv preprint},
  year    = {2023},
  eprint  = {2308.12966},
  archivePrefix = {arXiv},
  primaryClass  = {cs.CV}
}

@inproceedings{dreambooth2023,
  author={Ruiz, Nataniel and Li, Yuanzhen and Jampani, Varun and Pritch, Yael and Rubinstein, Michael and Aberman, Kfir},
  booktitle= CVPR, 
  title={DreamBooth: Fine Tuning Text-to-Image Diffusion Models for Subject-Driven Generation}, 
  year={2023},
  pages={22500-22510},
  }

@inproceedings{Cao_2025_ICCV,
  author    = {Cao, Yukang and Si, Chenyang and Wang, Jinghao and Liu, Ziwei},
  title     = {FreeMorph: Tuning-Free Generalized Image Morphing with Diffusion Model},
  booktitle = ICCV,
  year      = {2025}
}

@inproceedings{Hu2022LoRA,
  title     = {LoRA: Low-Rank Adaptation of Large Language Models},
  author    = {Hu, Edward J. and Shen, Yelong and Wallis, Phillip and Allen-Zhu, Zeyuan and Li, Yuanzhi and Wang, Shean and Wang, Lu and Chen, Weizhu},
  booktitle = ICLR,
  year      = {2022}
}

@misc{bfl2025fluxkontext,
  title        = {FLUX.1 Kontext: Flow Matching for In-Context Image Generation and Editing in Latent Space},
  author       = {{Black Forest Labs}},
  year         = {2025},
  howpublished = {arXiv preprint},
  eprint       = {2506.15742},
  archivePrefix= {arXiv},
  primaryClass = {cs.CV}
}

@misc{wu2025qwenimagetechnicalreport,
  title   = {Qwen-Image Technical Report},
  author  = {{Qwen Team}},
  year    = {2025},
  howpublished = {arXiv preprint},
  eprint  = {2508.02324},
  archivePrefix = {arXiv},
  primaryClass  = {cs.CV}
}

@inproceedings{ddpm,
  author = {Ho, Jonathan and Jain, Ajay and Abbeel, Pieter},
  booktitle = NeurIPS,
  title = {Denoising Diffusion Probabilistic Models},
  year = {2020}
}

@inproceedings{song2021denoising,
  title={Denoising Diffusion Implicit Models},
  author={Song, Jiaming and Meng, Chenlin and Ermon, Stefano},
  booktitle=ICLR,
  year={2021}
}

@inproceedings{sohl2015deep,
  title={Deep Unsupervised Learning Using Nonequilibrium Thermodynamics},
  author={Sohl-Dickstein, Jascha and Weiss, Eric and Maheswaranathan, Niru and Ganguli, Surya},
  booktitle=ICML,
  year={2015}
}

@inproceedings{parihar2025kontinuouskontext,
  title   = {Kontinuous Kontext: Continuous Strength Control for Instruction-based Image Editing},
  author  = {Parihar, Rishubh and Patashnik, Or and Ostashev, Daniil and Babu, R. Venkatesh and Cohen-Or, Daniel and Wang, Kuan-Chieh},
  year    = {2026},
  booktitle = CVPR,
}

@inproceedings{wang2023pix2pixzero,
  title   = {Zero-shot Image-to-Image Translation},
  author  = {Parmar, Gaurav and Singh, Krishna Kumar and Zhang, Richard and Li, Yijun and Lu, Jingwan and Zhu, Jun-Yan},
  booktitle = {SIGGRAPH Conference Proceedings},
  year    = {2023}
}

@inproceedings{tumanyan2023pnpdiffusion,
  title     = {Plug-and-Play Diffusion Features for Text-Driven Image-to-Image Translation},
  author    = {Tumanyan, Narek and Geyer, Michal and Bagon, Shai and Dekel, Tali},
  booktitle = CVPR,
  year      = {2023},
  pages     = {1921--1930}
}

@inproceedings{cao2023masactrl,
  title     = {MasaCtrl: Tuning-Free Mutual Self-Attention Control for Consistent Image Synthesis and Editing},
  author    = {Cao, Meng and Wang, Xintao and Qi, Jiakai and Shan, Ying and Qie, Xiaopeng and Zheng, Yinqiang},
  booktitle = ICCV,
  year      = {2023},
  pages     = {22503--22513}
}

@inproceedings{couairon2023diffedit,
  title     = {DiffEdit: Diffusion-based Semantic Image Editing with Mask Guidance},
  author    = {Couairon, Guillaume and Verbeek, Jakob and Schwenk, Holger and Cord, Matthieu},
  booktitle = ICLR,
  year      = {2023}
}

@inproceedings{rout2025rfinversion,
  title     = {Semantic image inversion and editing using rectified stochastic differential equations},
  author    = {Rout, Litu and Chen, Yujia and Ruiz, Nataniel and Caramanis, Constantine and Shakkottai, Sanjay and Chu, Wen-Sheng},
  booktitle = ICLR,
  year      = {2025}
}

@inproceedings{wang2025rfediting,
  title     = {Taming Rectified Flow for Inversion and Editing},
  author    = {Wang, Jiangshan and Pu, Junfu and Qi, Zhongang and Guo, Jiayi and Ma, Yue and Huang, Nisha and Chen, Yuxin and Li, Xiu and Shan, Ying},
  booktitle = ICML,
  year      = {2025}
}

@article{shen2020interfacegan,
  title   = {InterFaceGAN: Interpreting the Disentangled Face Representation Learned by GANs},
  author  = {Shen, Yujun and Yang, Ceyuan and Tang, Xiaoou and Zhou, Bolei},
  journal = PAMI,
  volume  = {44},
  number  = {4},
  pages   = {2004--2018},
  year    = {2022},
}

@inproceedings{patashnik2021styleclip,
  title     = {StyleCLIP: Text-Driven Manipulation of StyleGAN Imagery},
  author    = {Patashnik, Or and Wu, Zongze and Shechtman, Eli and Cohen-Or, Daniel and Lischinski, Dani},
  booktitle = ICCV,
  year      = {2021},
  pages     = {2085--2094}
}

@article{gal2021stylegannada,
  title   = {StyleGAN-NADA: CLIP-Guided Domain Adaptation of Image Generators},
  author  = {Gal, Rinon and Patashnik, Or and Maron, Haggai and Bermano, Amit H. and Chechik, Gal and Cohen-Or, Daniel},
  journal = TOG,
  volume  = {41},
  number  = {4},
  pages   = {1--13},
  year    = {2022}
}

@inproceedings{nichol2021improvedddpm,
  title     = {Improved Denoising Diffusion Probabilistic Models},
  author    = {Nichol, Alexander Quinn and Dhariwal, Prafulla},
  booktitle = ICML,
  year      = {2021},
  pages     = {8162--8171}
}

@inproceedings{dhariwal2021diffusionbeatgans,
  title     = {Diffusion Models Beat GANs on Image Synthesis},
  author    = {Dhariwal, Prafulla and Nichol, Alexander Quinn},
  booktitle = NeurIPS,
  year      = {2021},
  pages     = {8780--8794}
}

@inproceedings{lu2022dpmsolver,
  title     = {DPM-Solver: A Fast ODE Solver for Diffusion Probabilistic Model Sampling in Around 10 Steps},
  author    = {Lu, Cheng and Zhou, Yuhao and Bao, Fan and Chen, Jianfei and Li, Chongxuan and Zhu, Jun},
  booktitle = NeurIPS,
  year      = {2022}
}

@misc{pixabay,
  title  = {Pixabay},
  author = {{Pixabay GmbH}},
  year   = {2026},
  note   = {Accessed: 2026-02},
  url    = {https://pixabay.com/}
}

@inproceedings{zarei2025slideredit,
  title   = {SliderEdit: Continuous Image Editing with Fine-Grained Instruction Control},
  author  = {Arman Zarei and Samyadeep Basu and Mobina Pournemat and Sayan Nag and Ryan Rossi and Soheil Feizi},
  year    = {2026},
  booktitle = CVPR,
}
